\documentclass[10pt,twocolumn,letterpaper]{article}
\usepackage{wacv}
\usepackage{times}
\usepackage{makecell}
\usepackage{multirow}
\usepackage{arydshln, booktabs}
\usepackage{siunitx}
\usepackage{subcaption}
\usepackage{enumitem}
\usepackage{rotating}
\usepackage[breaklinks=true,colorlinks,colorlinks,bookmarks=false]{hyperref}

\setlist{nolistsep}

\wacvfinalcopy 

\begin{document}

\title{MobileStereoNet: Towards Lightweight Deep Networks for Stereo Matching}

\author{
	Faranak Shamsafar\thanks{Equal contribution.}\hspace{0.1cm}, Samuel Woerz\textsuperscript{*}\hspace{-0.1cm}, Rafia Rahim, Andreas Zell\\
	{\tt\small \{faranak.shamsafar@, samuel.woerz@student., rafia.rahim@, andreas.zell@\}uni-tuebingen.de}\\
	University of Tuebingen\\
	WSI Institute for Computer Science, Tuebingen, Germany \\
}

\maketitle
\begin{abstract}
Recent methods in stereo matching have continuously improved the accuracy using deep models. This gain, however, is attained with a high increase in computation cost, such that the network may not fit even on a moderate GPU. This issue raises problems when the model needs to be deployed on resource-limited devices. For this, we propose two light models for stereo vision with reduced complexity and without sacrificing accuracy. Depending on the dimension of cost volume, we design a 2D and a 3D model with encoder-decoders built from 2D and 3D convolutions, respectively. To this end, we leverage 2D MobileNet blocks and extend them to 3D for stereo vision application. Besides, a new cost volume is proposed to boost the accuracy of the 2D model, making it performing close to 3D networks. Experiments show that the proposed 2D/3D networks effectively reduce the computational expense (27\%/95\% and 72\%/38\% fewer parameters/operations in 2D and 3D models, respectively) while upholding the accuracy. Our code is available at \url{https://github.com/cogsys-tuebingen/mobilestereonet}.
\end{abstract}
\section{Introduction}
Stereo matching is one of the techniques for depth perception of a scene, which is established based on the displacement of the matching points in a binocular camera setup. Given a pair of rectified left/right images, we can compute depth by redirecting to disparity map estimation. Depth prediction is used in many real-world applications, like self-driving cars \cite{pal2020looking}, robotics \cite{wang2019normalized}, and object detection \cite{shi2020point}. Compared to other techniques for depth perception, like LiDAR and Time-of-Flight sensors, passive stereo vision is more desirable in real-world scenarios because of inherent problems in other techniques, such as sparsity of depth data, incompetency in sunlight or reflective/absorbing surfaces, and limited operating depth range.	

A stereo matching algorithm has three main components: feature extraction, regularization, and disparity selection. Before deep learning, numerous algorithms proposed different schemes for each step, like local descriptors (Sum of Absolute Difference or Census Transform \cite{zabih1994non}) for feature extraction, Semi-global Matching (SGM) \cite{hirschmuller2005accurate} for regularization, and winner-take-all (WTA) for final disparity selection. Nowadays, similar to other computer vision tasks, stereo matching has also benefited from deep networks. Primarily, the research incorporated deep learning in individual components of the pipeline, like \cite{zbontar2016stereo,park2016look} in matching costs and \cite{park2015leveraging,seki2017sgm} in regularization. Later, the research stirred towards end-to-end frameworks, taking all the fundamental components into one network \cite{mayer2016large,kendall2017end,chang2018pyramid,guo2019group}.
\begin{figure}[t]
	\begin{center}
		\includegraphics[width=1\linewidth]{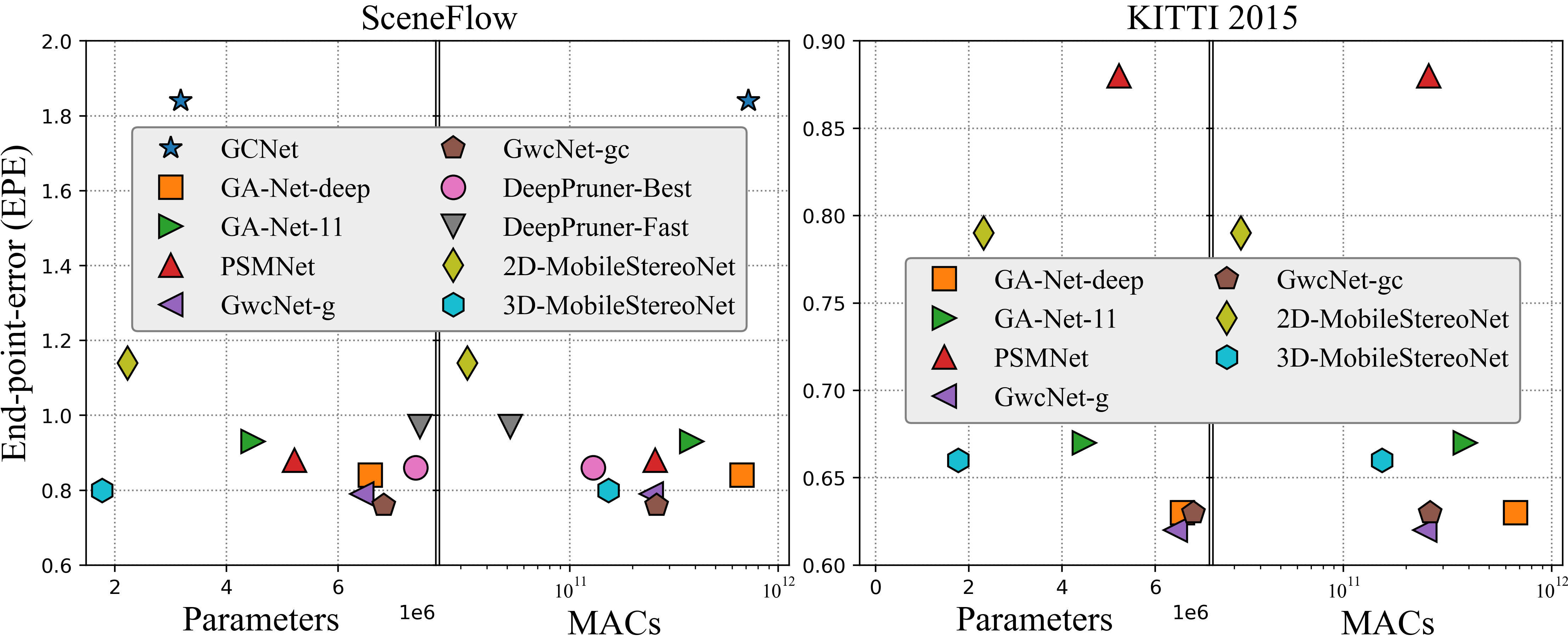}
	\end{center}			
	\vspace*{-0.6cm}
	\caption{Performance \vs computation cost on SceneFlow test set (\emph{Left}) and KITTI 2015 validation set (\emph{Right}): Metrics are EPE, number of parameters ($\times10^6$), and number of operations (MACs in log scale). For all, the lower is better. By using a new parameterized cost volume, 2D-MobileStereoNet shows closer performance to 3D models with the least MACs. 3D-MobileStereoNet obtains competitive accuracy with the least number of parameters.}
	\label{fig:ourResult}
	\vspace*{-0.55cm}	
\end{figure}

Accordingly, in end-to-end pipelines, depending on the dimension of the cost volume built on top of the unary features, the subsequent convolutional layers in the regularization part (encoder-decoder) can be either 2D convolutions (for 3D cost volume) \cite{mayer2016large} or 3D convolutions (for 4D cost volume) \cite{kendall2017end}. We dub the former and the latter group as ``2D'' and ``3D'' models, respectively. While the 3D end-to-end networks introduce highly boosted accuracy in disparity estimation, they are computationally costly due to more complex networks. As such, some of these networks cannot fit even on a moderate GPU, raising an out-of-memory (OOM) state. On the other hand, there are many embedded platforms with memory constraints on which the networks should fit and execute efficiently.

In order to reduce the complexity of 3D models, some works reconsider the configuration of the networks. For instance, DeepPruner \cite{duggal2019deeppruner} develops a PatchMatch module to ignore the cost volume evaluation for most of the disparity range. In \cite{yee2020fast}, authors establish an aggregation module on top of a cost volume computed by traditional local descriptors. Thus, skipping the convolutional feature extraction, the network benefits from a lighter learning paradigm. Also, it creates a 3D cost to avoid the curse of 3D convolutions.

In this work, we develop two end-to-end stereo matching networks, which exploit MobileNet-V1 \cite{howard2017mobilenets} and MobileNet-V2 \cite{sandler2018mobilenetv2} blocks to mitigate the computational burden in favor of real embedded platforms, like FPGAs or mobile devices. Depending on the cost volume dimension, \ie 3D or 4D, we propose a 2D and a 3D network. Moreover, for the 3D cost volume, a new learning construction module is devised based on interlacing the features from two viewpoints. Although reducing the computation cost is usually accompanied by a degradation in performance, we show that the proposed architectures are competitive with their state-of-the-art counterparts (Fig. \ref{fig:ourResult}). 

Overall, our main contributions are as follows: $i)$ Two lightweight models (2D/3D) are designed and proposed for stereo matching using MobileNet blocks without sacrificing accuracy. $ii)$ We raise MobileNet blocks from the originally proposed 2D convolutions to 3D for the application of stereo matching. Also, by analyzing their costs, we prove their merit in reducing the computational load when processing 4D data. $iii)$ We introduce a learnable cost volume module for the 2D model to keep the accuracy comparable with over-parameterized 3D models. $iv)$ Extensive experiments for analyzing the accuracy/complexity trade-off in different design choices are conducted. Our findings in the design choice can be applicable to similar 2D/3D networks to reduce their complexity.
\section{Related work}
Stereo vision is one of the popular techniques for estimating the depth from images. In the last decade, machine learning and deep learning approaches have well-progressed in computer vision tasks, including stereo matching. Deep learning-based methods can be categorized into two groups: methods that focus on transferring only one or some of the general pipeline components into a deep learning framework \cite{zbontar2016stereo,seki2017sgm,batsos2018cbmv}, and approaches that formulate the whole process in an end-to-end scheme \cite{mayer2016large, kendall2017end,chang2018pyramid,guo2019group,zhang2019ga}. Most of the end-to-end methods are developed based on 3D convolutional layers. Although these architectures achieve a substantial increase in accuracy, they require a high amount of memory usage, making them impractical for mobile and real-time applications, such as robotics and autonomous vehicles. 

\textbf{Lighter networks.} Lightweight architectures have become an active research domain, ringing the bell that it is getting impractical to slide through complex networks without considering the load of computations. Generally, convolutions entail the most considerable computational load.
Some deep networks for stereo reconstruction have been developed to achieve less complexity while being competitive in terms of accuracy with heavy 3D architectures. In \cite{yee2020fast}, an initial matching cost is constructed based on the traditional cost computation. After reducing the channel dimension through $1\times1$ convolutions, the data is fed into a U-Net to regress the disparity map. In another work, DeepPruner \cite{duggal2019deeppruner} mitigates the computational complexity of 3D convolutional layers by calculating the matching cost for a subset of possible disparity values. Recently, \cite{rahim2021} proposed to use feature-wise and feature-disparity-wise separable convolutions for optimizing the 3D stereo models. On the whole, according to the results in \cite{yee2020fast}, the 2D architectures can make a better trade-off between accuracy and speed. 

Other works mainly focus on reducing the complexity of 3D convolutional layers for other 3D vision tasks. Qiu \etal \cite{qiu2017learning} developed pseudo-convolutions that decouple a 3D convolution into 3D convolutions equivalent to a 2D convolution and a light 3D convolution, which aggregates information only across the third domain. The network is demonstrated on video classification. In \cite{ye20193d}, authors made use of 3D depth-wise convolutions for 3D reconstruction. Recently, \cite{feichtenhofer2020x3d} proposed progressively forward expanding a 2D tiny network along multiple axes to get fewer parameters for video classification and detection.
 
\textbf{Cost volume computation.} In a stereo model, measuring the similarity of left/right features is just as important as feature extraction and regularization. Traditional algorithms utilized simple calculations, like absolute difference, Hamming distance, or correlation. Similar solutions carried on to deep learning-based networks. Specifically, correlation is used on top of unary features to compute a 3D cost volume \cite{dosovitskiy2015flownet,luo2016efficient,mayer2016large,zbontar2016stereo,ilg2018occlusions}. Later, Kendall \etal \cite{kendall2017end} proposed to concatenate unary features to make a 4D cost, requiring 3D convolutions 
in the following. After that, this approach was mainly adapted for 3D models with some modifications to enhance the accuracy, \eg variance-based \cite{rao_he_dai_zhu_li_he_2020}, group-wise correlation \cite{guo2019group} and pyramid \cite{wu2019semantic} cost volume.
\section{Methodology}
As a first step, we reformulate two common light blocks \cite{howard2017mobilenets,sandler2018mobilenetv2} to raise them from 2D convolutions to 3D for the application of stereo matching. We also analyze their computation cost \wrt the standard 2D/3D convolution counterparts. The computation cost of a deep network is measured by the number of operations in MACs (Multiply-Accumulate) and the number of parameters. While the number of parameters is fixed for a model, MACs depend on the input size.  
\subsection{Light blocks replacing 2D/3D convolutions}
\begin{figure}[t]
	\begin{center}
		\includegraphics[width=0.185\linewidth]{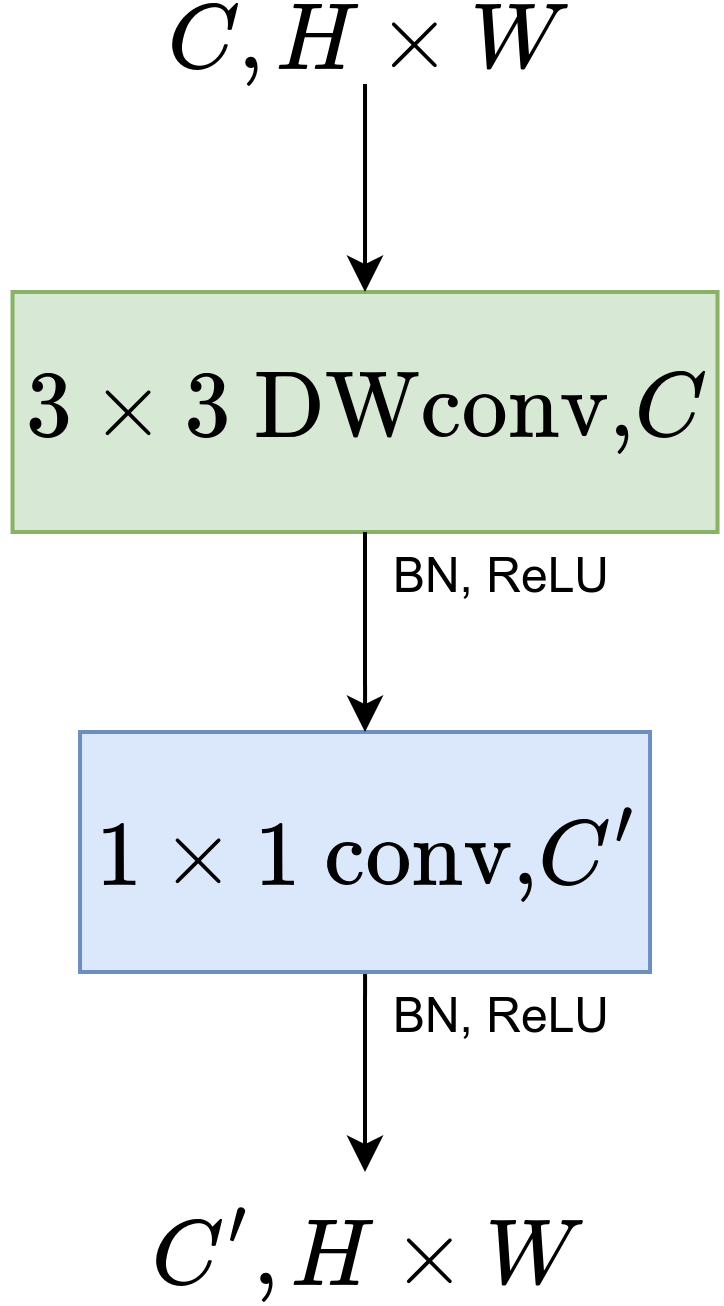}	
		\includegraphics[width=0.229\linewidth]{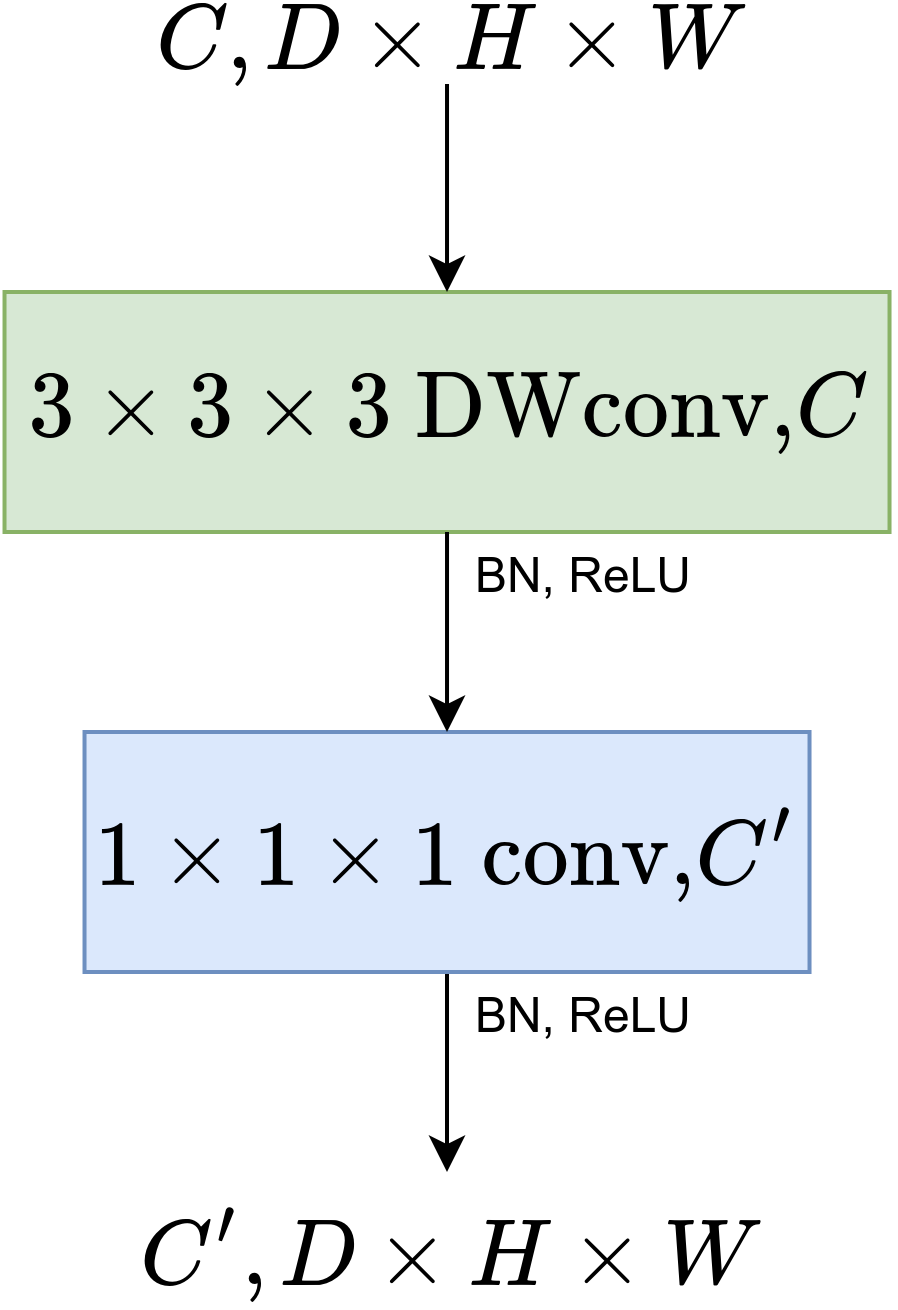} \hspace{0.27cm}
		\includegraphics[width=0.242\linewidth]{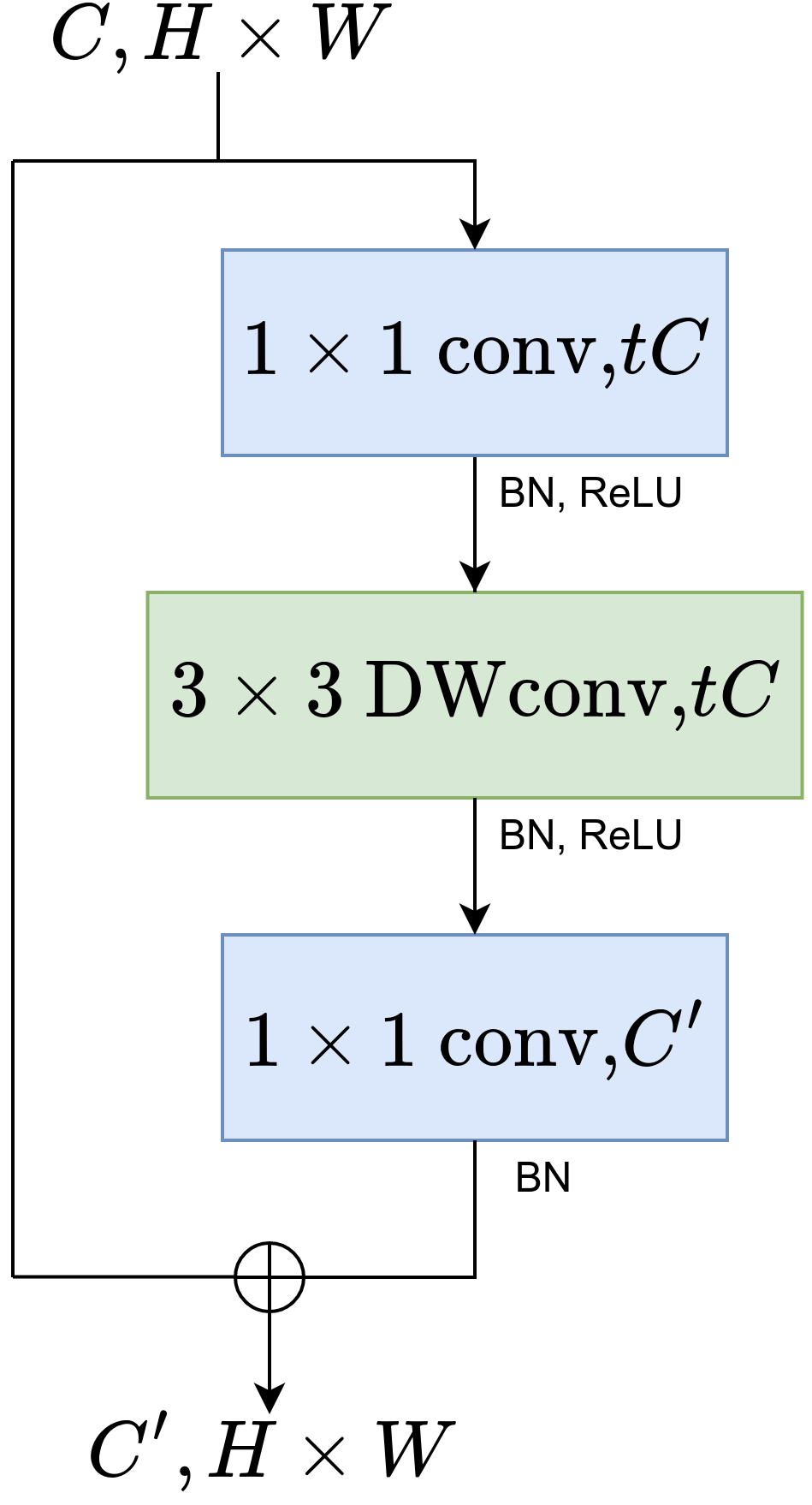} \hspace{-0.15cm}
		\includegraphics[width=0.274\linewidth]{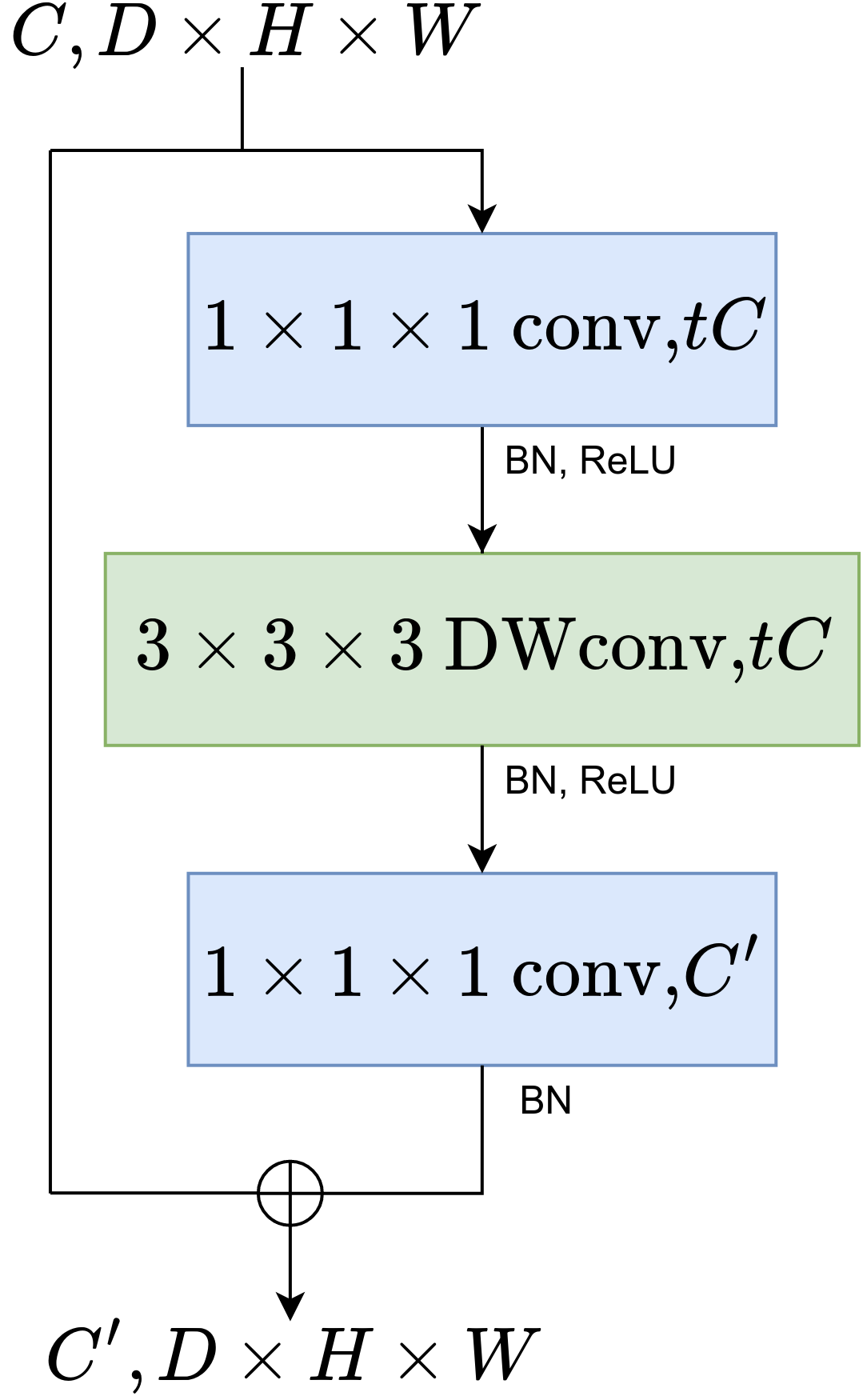}
	\end{center}
	\vspace*{-0.5cm}
	\caption{\emph{Left}: MobileNet-V1 block and its extension to 3D, \emph{Right}: MobileNet-V2 block and its extension to 3D.}
	\label{fig:mobilenets}
	\vspace*{-0.5cm}	
\end{figure}

As a pioneer work, MobileNet-V1 \cite{howard2017mobilenets} employs depth-wise and point-wise convolutions to produce an output with the same size as the output of a standard convolution, but with fewer computations. Later, MobileNet-V2 \cite{sandler2018mobilenetv2} was introduced, which formulates its block with point-wise, depth-wise, and once again, point-wise layers. The number of input and output channels are specific for each layer, such that the channel dimension is expanded with an expansion factor ($t$) within the block. In a non-downsampling layer, a skip connection is included as well, making it a so-called \emph{Inverted Residual} block. In Fig. \ref{fig:mobilenets}, MobileNet-V1 and MobileNet-V2 blocks (shortened to $v1$ and $v2$ hereon, for simplicity) are shown.

\textbf{Raising MobileNet blocks to 3D for stereo network.} Originally, $v1$ and $v2$ blocks are designed to replace 2D convolutions and are proved mainly for sparse prediction tasks, like image classification. Still, many other computer vision problems require 3D convolutions to operate on 4D data, \eg tasks with temporal input besides the spatial data. Likewise, dense disparity estimation is such a topic, which explores the space for the 3rd dimension of the scene. This outlook of using 3D convolutions for stereo matching has emerged recently, where they are utilized for processing 4D cost data. Hence, we take $v1$ and $v2$ blocks to their 3D counterparts for stereo vision application and show their necessity for light stereo vision models. 

For this, just as in 2D convolutions, we commit the depth-wise and point-wise convolutions in the channel (feature) dimension in 3D convolutions. To be more precise, to raise the convolutions from 2D to 3D, the input data is extended from $C\times H\times W$ to $C\times D\times H\times W$, where $H$ and $W$ indicate the input height and width, respectively; $D$ is the new third dimension, and $C$ is the number of channels. In our formulation for stereo matching, the 4D data of cost volume \cite{kendall2017end,wu2019semantic,guo2019group} is likewise of size $C\times D\times H\times W$, where $D$ is the predefined disparity range for building the cost volume. Accordingly, we raise kernels from 2D to 3D, considering the same size for the added dimension, \ie if the 2D kernel is $3\times3$, the 3D kernel is $3\times3\times3$. Therefore, taking $v1$ and $v2$ blocks to 3D is straightforward by applying depth-wise and point-wise convolutions in the channel dimension. Figure \ref{fig:mobilenets} displays the new $v1$ and $v2$ blocks raised to 3D.

In Tab. \ref{tab:ConvMAC}, we compute the cost of the standard 2D/3D convolutions and 2D/3D $v1$ and $v2$ blocks. Comparing to standard convolution counterparts, there is a reduction factor in computation cost that depends on the kernel size $k$, channels ($C_{in}$ and $C_{out}$) and expansion factor ($t$) in $v2$ blocks. The example in the last column shows that $v1$ blocks are lighter than $v2$ in both 2D and 3D versions, as expected. Moreover, exploiting $v1$ and $v2$ blocks in 3D type is more desirable as they show the capability for further reduction in operations compared to standard counterparts.

\begin{figure}
	\begin{center}
		\includegraphics[width=0.75\linewidth]{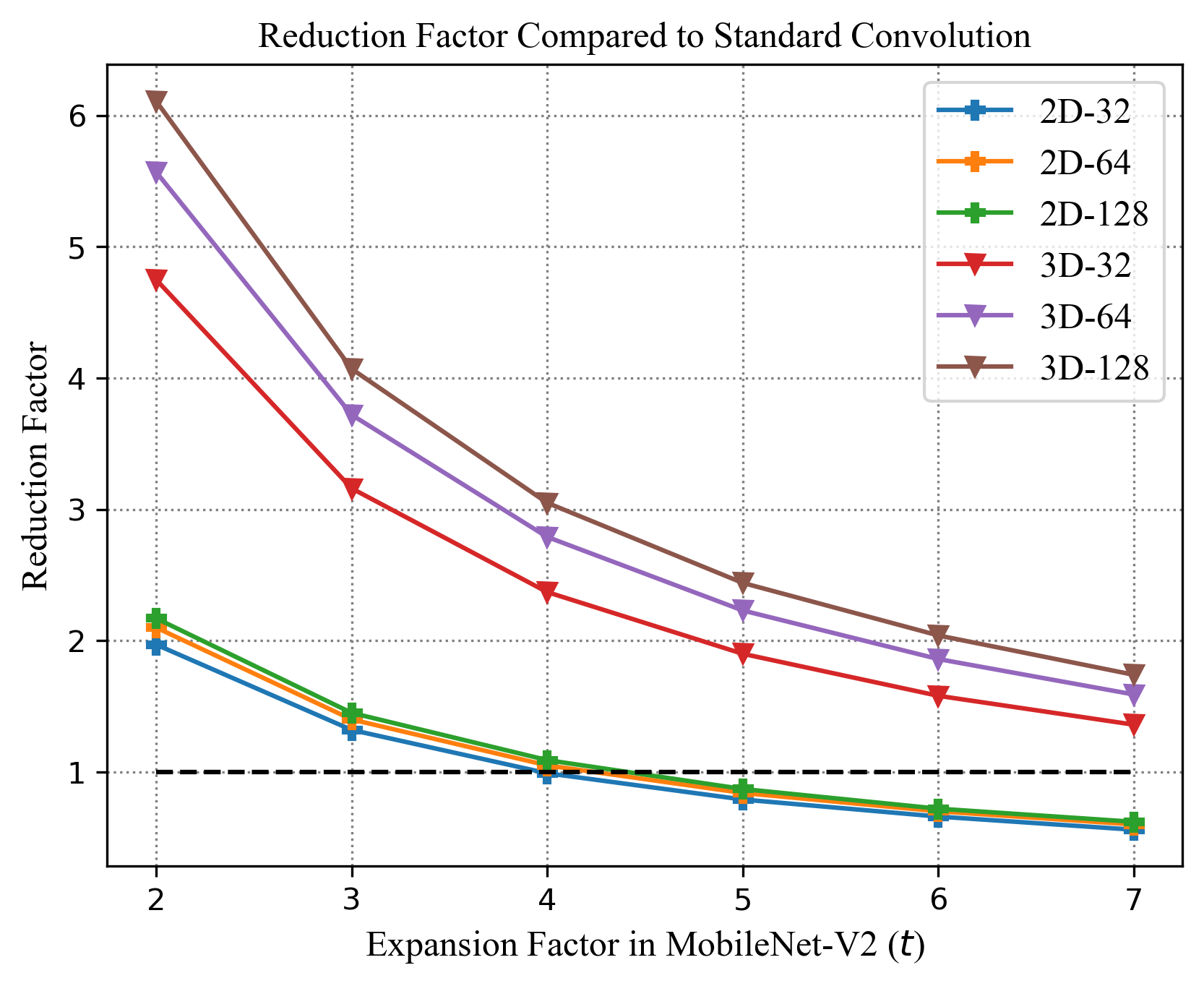}	
	\end{center}
	\vspace*{-0.65cm}
	\caption{Reduction factor of 2D/3D $v2$ blocks with varying expansion factor ($t$) \wrt standard convolution counterparts. The right hand numbers indicate the channel numbers ($C_{in}=C_{out}$ and $k=3$).}
	\label{fig:expansion}
	\vspace*{-0.3cm}	
\end{figure}

Additionally, we examine the impact of the expansion factor of $v_2$ blocks in Fig. \ref{fig:expansion}. We consider the common cases in an hourglass module (\eg in \cite{chang2018pyramid,guo2019group}) for evaluation, \ie $C_{in}=C_{out}=\{32, 64, 128\}$ with $k=3$. We see that by increasing $t$, the reduction factor decreases until a point where the block becomes heavier than a standard convolution counterpart. Also, the 2D block is more sensitive to $t$, such that the cost is increased beyond the convolution after $t=5$. Here, once again, we can verify the merit of MobileNet architectures in our reformulation for 3D convolutional layers of 3D networks.
\begin{table}[]
	\begin{center}
		\footnotesize	
		\begin{tabular}{c|l|c|c}
			\multicolumn{2}{c|}{Operator} & \makecell{Operations \textendash\ MACs \\($ (D\times) H \times W \times )$} & \makecell{Example for \\ Reduction Factor} \\ \hline \hline
			& Std. Conv. & $k\times k \times C_{in} \times C_{out}$ &  {-}\\ \cline{2-3}
			2D & $v_1$ Block & \makecell{$(k\times k \times C_{in}) +$ \\ $( C_{in} \times C_{out})$} & 7.9x \\ \cline{2-3}
			& $v_2$ Block & \makecell{$(C_{in} \times t \times C_{in})$ \\ $(k\times k \times t \times C_{in}) +$ \\ $( t \times C_{in} \times C_{out}) $} &  2.7x \\ \hline
			& Std. Conv. & $k\times k \times k \times C_{in} \times C_{out}$ &  {-}\\ \cline{2-3}
			3D & $v_1$ Block & \makecell{$(k\times k\times k \times C_{in}) +$ \\ $( C_{in} \times C_{out})$} & 18.9x \\ \cline{2-3}
			& $v_2$ Block & \makecell{$(C_{in} \times t \times C_{in})$ \\ $(k\times k\times k \times t \times C_{in}) +$ \\ $( t \times C_{in} \times C_{out}) $} & 7.0x\\ \hline 
		\end{tabular}
	\end{center}
	\vspace*{-0.65cm}
	\caption{Computation cost of the standard convolutions \vs MobileNet-V1 ($v_1$) and MobileNet-V2 ($v_2$) blocks in 2D and 3D counterparts. In MACs, $(D \times)$ belongs to 3D types. We have ignored the computation cost of batch normalization, ReLU and residual connection of $v_2$. Example for reduction factor (\wrt the standard convolution counterparts) is computed for $k=3$, $C_{in}=32$, $C_{out}=64$, and $t=2$.}
	\label{tab:ConvMAC}
	\vspace*{-0.55cm}
\end{table}
\subsection{Proposed models}
Here, we describe two end-to-end baselines (2D and 3D) and design MobileStereoNets in reliance on these two models and 2D/3D $v1$ and $v2$. Following the common pipeline in recent work, we first feed the rectified left/right images to the feature extraction backbone and obtain the unary features. The backbone is shared for the left and right images. The results are passed into a cost volume construction module to merge the data from two viewpoints. Finally, an encoder-decoder (hourglass) is applied on top of the cost volume to estimate the disparity map. 

\begin{figure}[t]
	\begin{center}
		\includegraphics[width=1\linewidth]{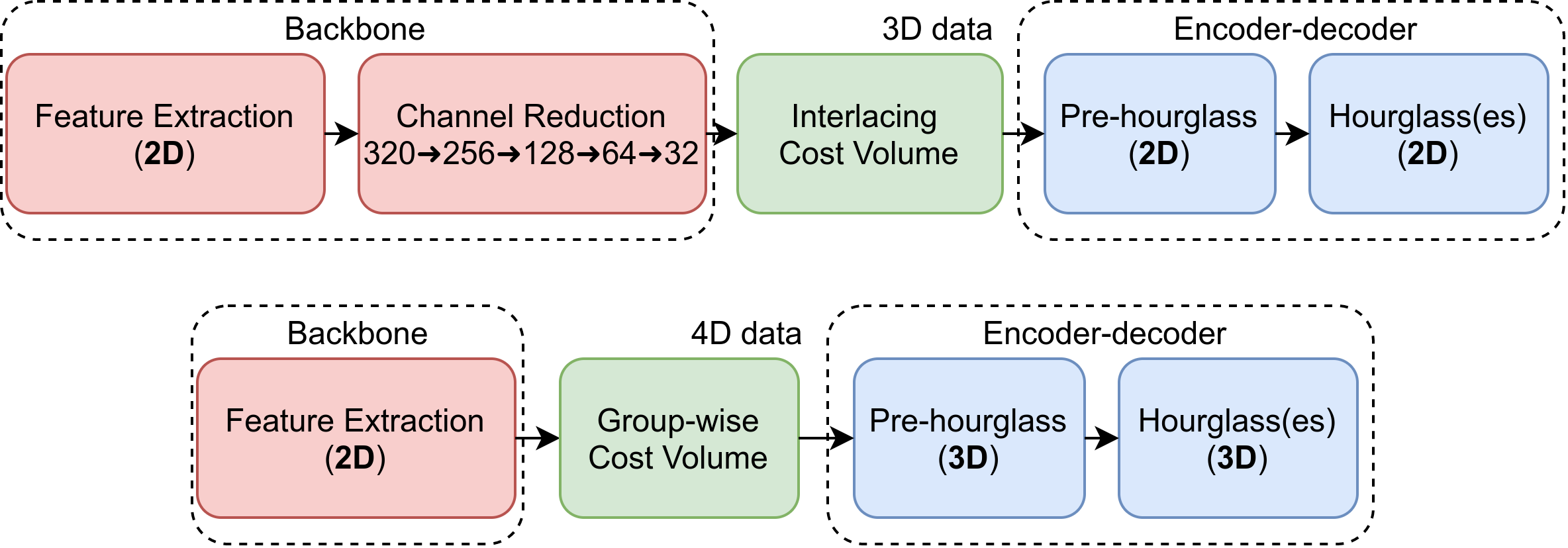}	
	\end{center}
	\vspace*{-0.5cm}
	\caption{\emph{Top} (2D Baseline): After feature extraction, data is reduced to $32$ channels. A 3D cost volume, {\small$(d_{max}/4)\times (H/4)\times (W/4)$}, is generated using the proposed interlacing cost volume construction and it is processed by 2D convolutions. \emph{Bottom} (3D Baseline): A 4D cost volume, {\small$40\times (d_{max}/4)\times (H/4)\times (W/4)$}, is computed using Gwc40 \cite{guo2019group} and it is processed with 3D convolutions.}
	\label{fig:BaseModels}
	\vspace*{-0.5cm}	
\end{figure}

\textbf{3D baseline.} For this case, we adopt GwcNet-g \cite{guo2019group} with only one hourglass (Fig. \ref{fig:BaseModels}) as it is performing superior to other similar designs, \eg GCNet \cite{kendall2017end}, PSMNet \cite{chang2018pyramid}, and GA-Net \cite{zhang2019ga}. This baseline utilizes a ResNet-like backbone and an encoder-decoder with 3D convolutions. Namely, a 4D cost volume is constructed by group-wise correlation of unary features, requiring 3D convolutions afterward. The encoder-decoder consists of an hourglass \cite{chang2018pyramid,guo2019group} outputting a downsampled disparity map, which after upsampling is compared against the ground-truth with a smooth-$L1$ loss function. For the detailed architecture, we refer the reader to the appendix.

\textbf{2D baseline.} In order to develop a much lighter stereo network, we modify the 3D baseline such that it uses an encoder-decoder with 2D convolutions (Fig. \ref{fig:BaseModels}). This approach contrasts with the recent trend, where 3D convolutions are deployed to add a feature dimension for disparity via a 4D cost volume. With a ResNet-like backbone similar to the 3D baseline, an input image with resolution $H\times W$ is turned into a feature of size $320\times(H/4)\times(W/4)$. We add further processing by four successive point-wise convolutions to reduce the number of channels and attain a size of $32\times(H/4)\times(W/4)$. In order to aggregate two unary features to form a cost volume, which indicates a similarity measurement in the left/right images across the disparity dimension, we propose a new \emph{Interlacing Cost Volume}. Note that we need a 3D cost volume to retain the encoder-decoder with 2D convolutions. Ignoring the feature dimension for disparity, 3D cost volume is of size $(d_{max}/4)\times(H/4)\times(W/4)$, where $d_{max}$ is the maximum disparity level. Finally, the 3D cost volume is taken into the encoder-decoder module after passing through two convolutions as a pre-hourglass module.

\textbf{Interlacing cost volume construction.} Traditionally, a cost volume for stereo matching is computed for comparing the descriptors of binocular images across the disparity dimension, mainly as \emph{3D data} as following:
\vspace{-0.2cm}
\begin{equation}
	C_{3D}(d,x,y) = G(f_L(x,y),f_R(x-d,y)),
\end{equation}
where $(x,y)$ and $d$ are the spatial location and the disparity value within a range of $(0,d_{max})$, respectively. $f_R(x-d,y)$ is the traversed right feature for a specific disparity level. $G(.,.)$ indicates a similarity measurement function that was conventionally chosen as correlation or Hamming distance. With the advent of deep learning in stereo vision, as the unary features raised into 3D data, \ie $f(x,y)\to f(.,x,y)$, DispNetC \cite{mayer2016large} proposed to use a correlation layer (dot product) to merge these data by $f_L(.,x,y)\odot f_R(.,x-d,y)$. Later, \cite{kendall2017end} introduced a cost volume as \emph{4D data}, $C_{4D}(d,.,x,y)$, by concatenating the unary features.

These 3D or 4D cost volumes are obtained in an unparameterized module of the network. Thus, we propose a subnetwork, named \emph{Interlacing Cost Volume Construction} with the motivations as following: $i)$ In order to achieve a better aggregation of the two unary features, we propose a parameterized subnetwork to learn the aggregation. $ii)$ By interlacing the left/right unary features, the corresponding features maps are distilled by a kernel. $iii)$ we aim to retain the encoder-decoder with only 2D convolutions (and not 3D, which is the case in recent work), as this can contribute to significant reduction of operations.
\begin{figure*}[t]
	\begin{center}
		\includegraphics[width=1\linewidth]{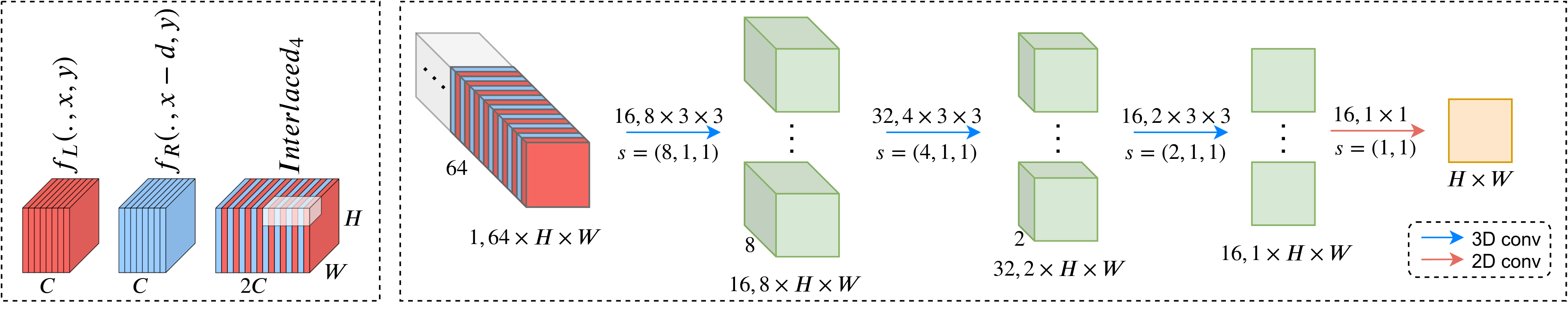}
	\end{center}
	\vspace*{-0.8cm}
	\caption{\emph{Left}: Interlacing cost volume construction at a particular disparity from the left (red) and right (blue) unary features. The kernels of the first layer take a group of non-overlapping interlaced features (here, four channels from each unary feature, $Interlaced_4$). \emph{Right}: Processing the data with more convolutional layers to yield the aggregated feature for that disparity level. The numbers above and below the arrows indicate the kernel size and the stride, respectively.}
	\label{fig:interlaced}
	\vspace*{-0.4cm}	
\end{figure*}

Namely, given the left and traversed right unary features, each of size $C\times H\times W$, we first interlace them across the channel dimension to form a data with double-sized channels, $2C\times H\times W$ (Fig. \ref{fig:interlaced}). After unsqueezing the output and raising it to 4D data ($1\times2C\times H\times W$), a 3D convolutional layer is applied such that the 3D kernels cover a specific number of left/right feature pairs. That is , a kernel of size $2i\times3\times3$ covers $i$ channel from each of the unary features. By increasing the $i$, the kernel covers more features and thus, integrates more information. The two following layers (Fig. \ref{fig:interlaced}) are also 3D convolution with double and same number of kernels' of the first layer. In general, the kernels are of size $m\times3\times3$ with strides as $m\times1\times1$, showing that they are covering non-overlapping channels. Finally, we convert the data to a single channel and pass through one 2D convolutional layer. Note that similar to other methods for cost volume, the spatial resolution is unchanged. We can write the general formula for a certain disparity level as Eg. \ref{Eq}. In Sec. \ref{experimental}, we show that the inclusion of learnable weights in stereo network contributes to better aggregation of the left/right features.
\vspace{-0.2cm}
\begin{equation}
	\label{Eq}
	C_{3D}(d,x,y) = \textrm{Interlace}\{f_L(.,x,y),f_R(.,x-d,y)\}
\end{equation}

\textbf{MobileStereoNets.}  In our baselines, the feature extraction and channel reduction are essentially processed with 2D convolutions. On the other hand, the hourglass employs 2D or 3D convolutions depending on the constructed cost volume. In order to obtain lighter networks, we replace these components with 2D and 3D counterparts of $v1$ and $v2$ blocks. There are different design choices when exploiting these blocks in individual modules of the networks. Thus, extensive experiments are conducted to answer the following questions: $i)$ Can we replace different modules in the 2D/3D baselines with 2D/3D $v1$ and $v2$ to achieve lighter stereo networks and keep the error rates low? $ii)$ If so, which modules should be replaced with them for a better compromise? $iii)$ Which block type performs better in terms of accuracy and computational load?  

Our experiments (\emph{cf.} Sec. \ref{experimental}) lead to our MobileStereoNets by modifying the baselines as follows:
\begin{itemize}[leftmargin=*, noitemsep]
	\item First Convolutions: Each of the three initial $3\times3$ convolutions are replaced with one $v2$. Using an expansion factor of $t=3$ provides a favorable trade-off between the performance and computational complexity. 
	\item Feature Extraction: We retained the original layer architecture and block structure, consisting of two $3\times3$ convolutions and a residual connection between each block. Substituting these convolutions with $v1$ keeps performance competitive while reducing the computational complexity significantly. 
	\item Channel Reduction: In the 2D baseline, we keep this module, \ie four $1\times1$ convolutions, unchanged with standard convolutions as replacing them with lighter blocks deteriorates the performance.
	\item Pre-hourglass: In the 2D baseline, we replace the two $3\times3$ convolutions in both of the blocks with $v2$. For 3D-MobileStereoNet, we use our extension of $v2$ to 3D instead of 3D convolutions. In both models, we choose expansion factor as $t=3$. 
	\item Hourglass: We employ a stack of three hourglasses for both 2D and 3D models. While the 3D network uses the same channel dimension as GwcNet \cite{guo2019group} (32), the hourglass width is increased to 48 in the 2D model. 2D-MobileStereoNet uses $v2$ instead of 2D convolutions. In 3D-MobileStereoNet, we once again swap the 3D convolutions for our extension of $v2$ to 3D. For both models, the expansion factor is $t=2$.  
\end{itemize}
\section{Experimental results and discussion}
\label{experimental}
Here, we first evaluate the performance of the proposed interlaced cost volume. Then, extensive experiments for taking the 2D/3D $v1$ and $v2$ blocks into the baselines are elaborated to show the path taken to reach the final architectures. Finally, we compare our developed networks with other methods. Note that $d_{max}$ is 192 in all cases.

In order to analyze different design choices, we use the SceneFlow ``final pass'' dataset \cite{mayer2016large}, consisting of 35,454/4,370 training/test samples with $540\times960$ resolution. This dataset can also help to pretrain the networks for limited real datasets. The quantitative evaluation for SceneFlow images is mainly reported with End-Point-Error (EPE), the mean average disparity error in $px$. Two more errors are also reported, \ie px-3 and D1, which are percentages of the outliers with disparity errors larger than $3px$ and $\max(3px, 0.05\times\text{ground-truth})$, respectively.
\subsection{Cost volume construction}
To verify the performance of the interlacing cost volume, we replace the corresponding module in 2D baseline (Fig. \ref{fig:BaseModels}) with correlation, which is adopted by 2D networks \cite{dosovitskiy2015flownet,mayer2016large,ilg2018occlusions}. Table \ref{fig:interlaced} shows the evaluation of this baseline against the model embedded with our interlacing cost volume. $Interlaced_i$ indicates that $i$ channels are taken from each unary feature, and the kernel of the subsequent layer is of size $2i\times3\times3$. For instance, when $i=4$, four channels from each feature data are combined by $8\times3\times3$ kernels. Therefore, cases of $i>1$ can be interpreted as \emph{group-wise} interlacing. We observe that better similarity measurements between the left/right features are achieved by introducing this learnable cost volume construction, resulting in lower EPE. According to the table, the best case is achieved when $i=4$ (also depicted in Fig. \ref{fig:interlaced}), and hence, we consider this case for further experiments in the 2D baseline.

We also investigate the effect of interlacing against direct concatenation of left/right unary features. According to the table, the error has increased in this case, showing that direct concatenation is not efficiently distilling the corresponding left/right features, which is essential for stereo matching.
\begin{table}[t]
	\begin{center}
		\footnotesize	
		\begin{tabular}{l|c|c|c}
			\hline
			{Method} & {EPE($px$) $\downarrow$} & {D1(\%) $\downarrow$} & {px-3(\%) $\downarrow$} \\
			\hline
			$Concatenation$ & 1.86 & 7.46 & 8.48\\
			$Correlation$ & 1.71 & 6.80 & 7.84 \\ \hdashline
			$Interlaced_1$ & 1.70 & 6.20 & \textbf{7.06} \\
			$Interlaced_2$ & 1.61 & 6.39 & 7.31 \\
			$Interlaced_4$ & \textbf{1.55} & \textbf{6.15} & \textbf{7.06} \\
			$Interlaced_8$ & 1.64 & 6.41 & 7.35 \\
			$Interlaced_{16}$ & 1.73 & 6.65 & 7.58\\ \hline
		\end{tabular}
	\end{center}
	\vspace*{-0.65cm}
	\caption{Performance evaluation on SceneFlow test set for the proposed 3D cost volume: The costs created by $Interlaced_i$ contribute to lower error rates.}
	\label{tab:interlacing}
	\vspace*{-0.55cm}	
\end{table}
\subsection{Effect of incorporating MobileNet blocks}
In this section, we incorporate $v_1$ and $v_2$ blocks (either 2D or 3D, depending on the type of convolutional module) in various components of the 2D and 3D baselines. In addition to error metric, we monitor the reduction of computational complexity to help us choose lighter models. We analyze replacing the fundamental modules of the network, \ie feature extraction and hourglass with lighter blocks. The results of substituting these modules with $v_1$ and $v_2$ ($t=2$) for 2D and 3D models are tabulated in Tables \ref{tab:2DNet_comb} and \ref{tab:3DNet_comb}. The best model is selected according to the least EPE obtained in 20 epochs. Also, the input resolution for computing MACs is $256\times512$. In these cases, the first convolutions of feature extraction are kept in standard type.
\begin{table}[tbp]
	\captionsetup[subtable]{labelformat=empty}
	\centering
	\begin{minipage}[c]{0.01\linewidth}
		\begin{turn}{90}\scriptsize{``FE'' and ``HG'' stand for feature extraction and hourglass.}\end{turn}			
	\end{minipage}
	\begin{minipage}[c]{0.97\linewidth}
		\begin{center} 
			\begin{subtable}{0.97\linewidth}
				\footnotesize
				\centering		
				\begin{tabular}{c;{1pt/3pt}c|S[table-format=1.2]S[table-format=2.2]S[table-format=1.2]}
					\hline
					{FE$_{2D}$} & {HG$_{2D}$} & {EPE($px$)} & {MACs($G$)} & {Params($M$)} \\ \hline
					conv. & conv. & 1.55 & 74.42 & 4.07 \\ \hline			
					conv. & $v_1$ & 1.62 & 73.97 & 3.52 \\
					$v_1$ & conv. & 1.66 & 30.43 & 1.52 \\
					$v_1$ & $v_1$ & 1.59 & 29.98 & 0.98 \\ \hline			
					conv. & $v_2$ & 1.63 & 74.32 & 3.75 \\
					$v_2$ & conv. & 1.57 & 35.54 & 1.81 \\
					$v_2$ & $v_2$ & 1.53 & 35.44 & 1.49 \\ \hline
					$v_1$ & $v_2$ & 1.50 & 30.33 & 1.21 \\
					$v_2$ & $v_1$ & 1.60 & 35.10 & 1.26 \\ \hline
				\end{tabular}
				\vspace*{-0.2cm}
				\caption{\footnotesize{(a) 2D models: 3D cost volume using $Interlaced_4$ method}}
				\label{tab:2DNet_comb}
				\footnotesize
				\centering
				\begin{tabular}{c;{1pt/3pt}c|S[table-format=1.2]S[table-format=2.2]S[table-format=1.2]}			
					\hline
					{FE$_{2D}$} & {HG$_{3D}$}& {EPE($px$)} & {MACs($G$)} & {Params($M$)} \\ \hline
					conv.	& conv. & 0.97 & 155.2 & 4.21 \\ \hline
					conv.	& $v_1$ & 0.99 & 143.66 & 3.42 \\
					$v_1$	& conv. & 0.98 & 111.2 & 1.66 \\
					$v_1$	& $v_1$ & 1.03 & 99.67 & 0.87  \\ \hline
					conv.	& $v_2$ & 0.97 & 149.01 & 3.53 \\
					$v_2$	& conv. & 0.96 & 116.32 & 1.94 \\			
					$v_2$	& $v_2$ & 0.97 & 110.13 & 1.27 \\ \hline
					$v_1$	& $v_2$ & 0.99 & 105.01 & 0.98 \\
					$v_2$	& $v_1$ & 1.02 & 104.78 & 1.15 \\ \hline 
				\end{tabular}	
				\vspace*{-0.2cm}
				\caption{\footnotesize{(b) 3D models: 4D cost volume using Gwc40 method \cite{guo2019group}}}
				\label{tab:3DNet_comb}	
			\end{subtable}
		\end{center}
	\end{minipage}
	\vspace*{-0.3cm}
	\caption{Performance evaluation on SceneFlow test set for variants of (a) 2D and (b) 3D baselines with $1\times HG$.}
	\vspace*{-0.5cm}
\end{table}

We can conclude that: $i)$ In both 2D and 3D baselines, feature extraction is responsible for much of the computational load. $ii)$ Substituting feature extraction with $v_1$ and hourglass with $v_2$ yields a better compromise between accuracy and computational complexity. For the 2D baseline, this combination results in the least EPE. We consider this combination for both 2D and 3D models.

We also examine replacing other modules with lighter blocks to make the network even lighter. Namely, the first convolutional layers in feature extraction and pre-hourglass modules are replaced with $v2$. The reason for choosing $v2$ instead of $v1$ is the higher accuracy $v2$ can maintain after substituting for standard convolutions. We observed that in this case, for the 2D baseline, both the complexity and the error are reduced (tables are available in the appendix). We also tried replacing other modules with MobileNet blocks, \ie the channel reduction module and the convolutions in interlacing cost volume construction in the 2D baseline. However, since these replacements deteriorate the learning capability of the network, they are kept unchanged with standard convolutions.

\subsection{Quantitative and qualitative results}
The discussed experiments support our design choice for the final 2D and 3D models, \ie 2D-MobileStereoNet and 3D-MobileStereoNet. To increase the accuracy, we utilize a stack of three hourglasses. Also, we found out that higher $t$ values make the network less accurate and heavier.
\begin{figure}[tbp]
	\captionsetup[subfigure]{labelformat=empty}
	\centering
	\begin{subfigure}[c]{.33\linewidth}
		\includegraphics[width=1\linewidth]{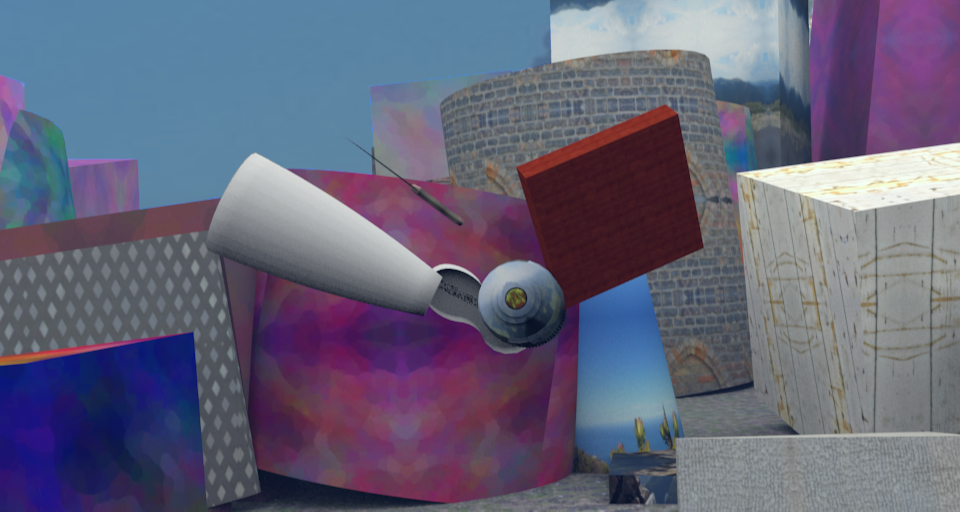}
		\vspace*{-4mm}
	\end{subfigure}
	\hspace{-1.4\baselineskip}
	\hfill
	\begin{subfigure}[c]{.33\linewidth}
		\includegraphics[width=1\linewidth]{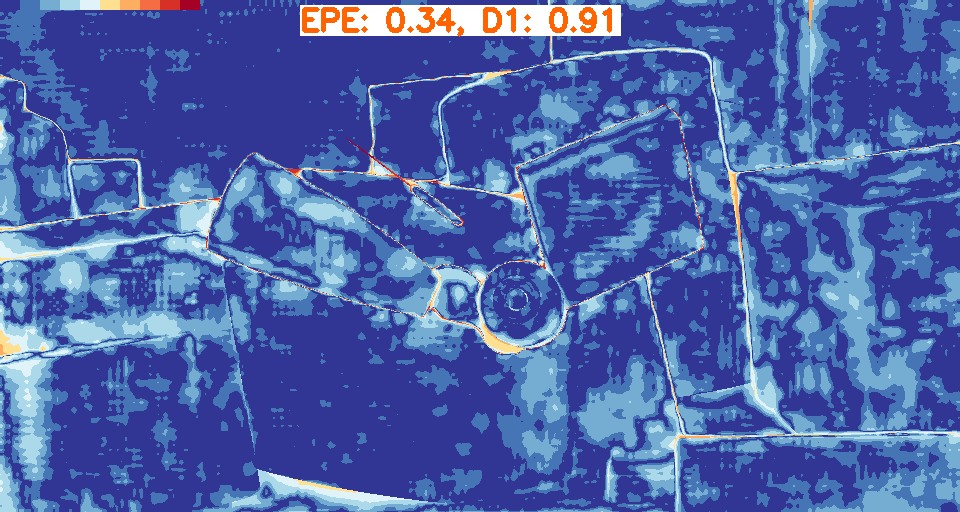}
		\vspace*{-4mm}
	\end{subfigure}
	\hspace{-1.4\baselineskip}
	\hfill
	\begin{subfigure}[c]{.33\linewidth}
		\includegraphics[width=1\linewidth]{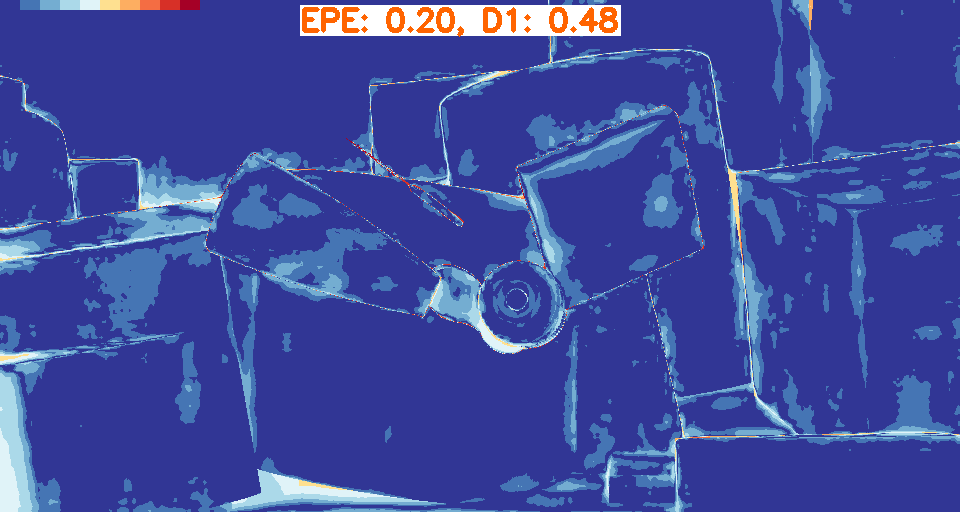}
		\vspace*{-4mm}
	\end{subfigure}
	\hspace{-1.4\baselineskip}
	\hfill
	\begin{subfigure}[c]{.33\linewidth}
		\includegraphics[width=1\linewidth]{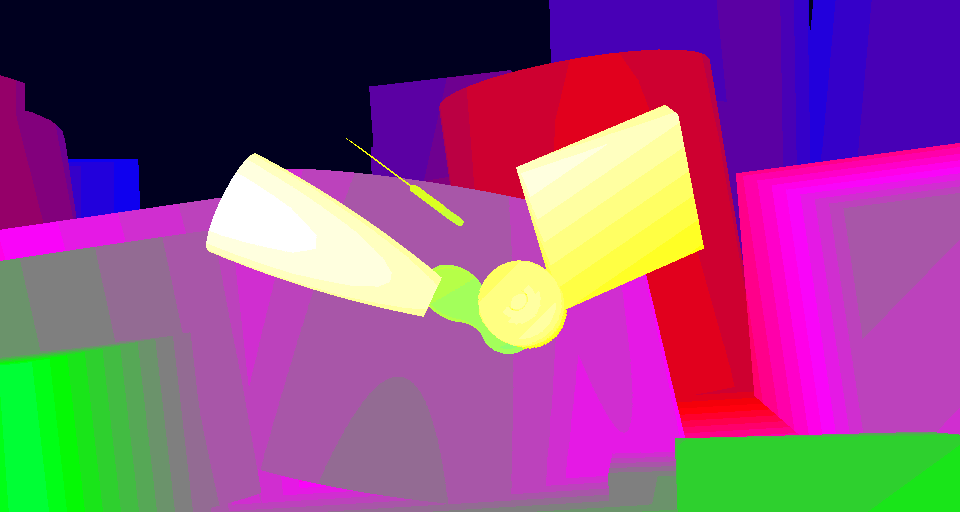}
		\vspace*{-4mm}
	\end{subfigure}
	\hspace{-1.4\baselineskip}
	\hfill
	\begin{subfigure}[c]{.33\linewidth}
		\includegraphics[width=1\linewidth]{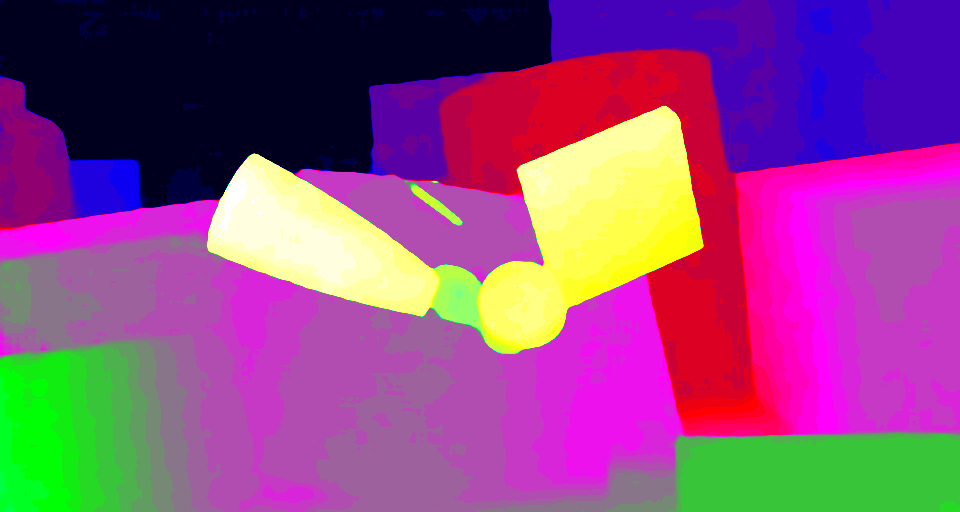}
		\vspace*{-4mm}
	\end{subfigure}
	\hspace{-1.4\baselineskip}
	\hfill
	\begin{subfigure}[c]{.33\linewidth}
		\includegraphics[width=1\linewidth]{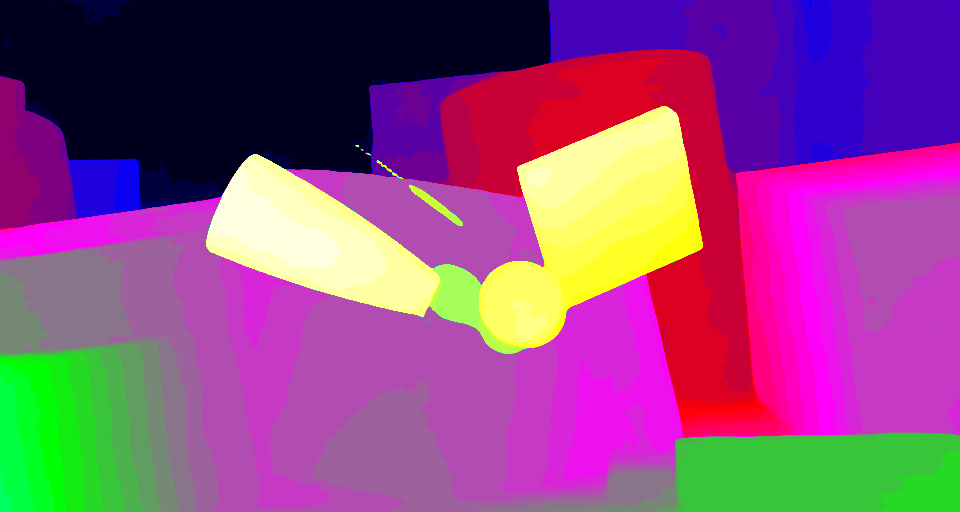}
		\vspace*{-4mm}
	\end{subfigure}
	\hspace{-1.4\baselineskip}
	\hfill
	\begin{subfigure}[c]{.33\linewidth}
		\includegraphics[width=1\linewidth]{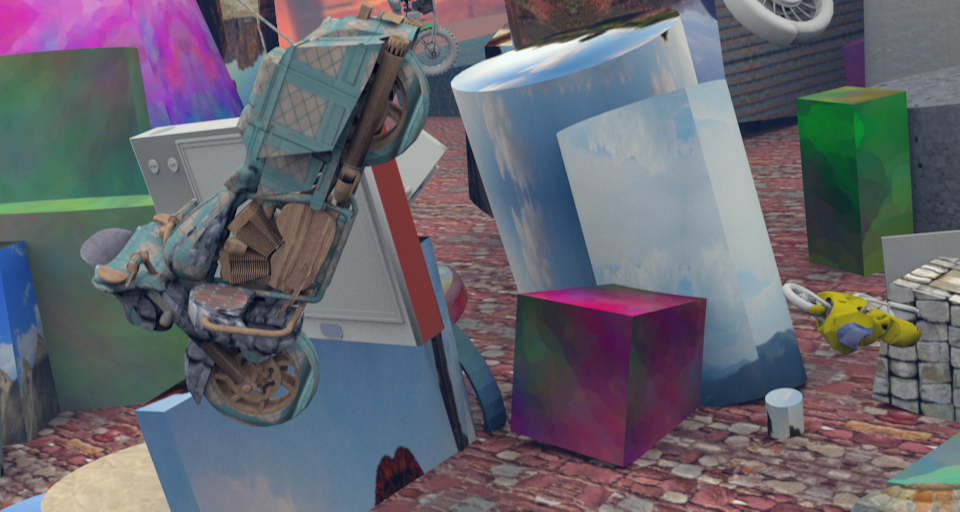}
		\vspace*{-4mm}
	\end{subfigure}
	\hspace{-1.4\baselineskip}
	\hfill
	\begin{subfigure}[c]{.33\linewidth}
		\includegraphics[width=1\linewidth]{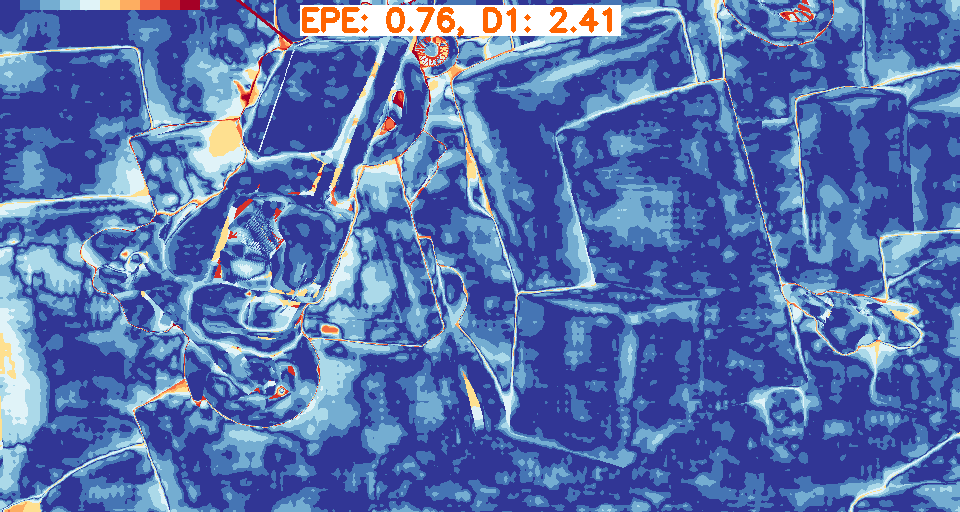}
		\vspace*{-4mm}
	\end{subfigure}
	\hspace{-1.4\baselineskip}
	\hfill
	\begin{subfigure}[c]{.33\linewidth}
		\includegraphics[width=1\linewidth]{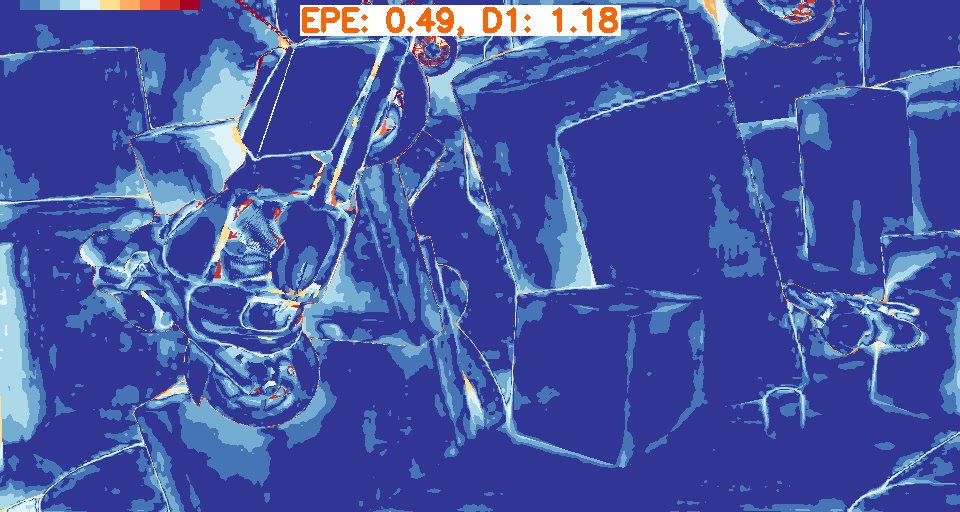}
		\vspace*{-4mm}
	\end{subfigure}
	\hspace{-1.4\baselineskip}
	\hfill
	\begin{subfigure}[c]{.33\linewidth}
		\includegraphics[width=1\linewidth]{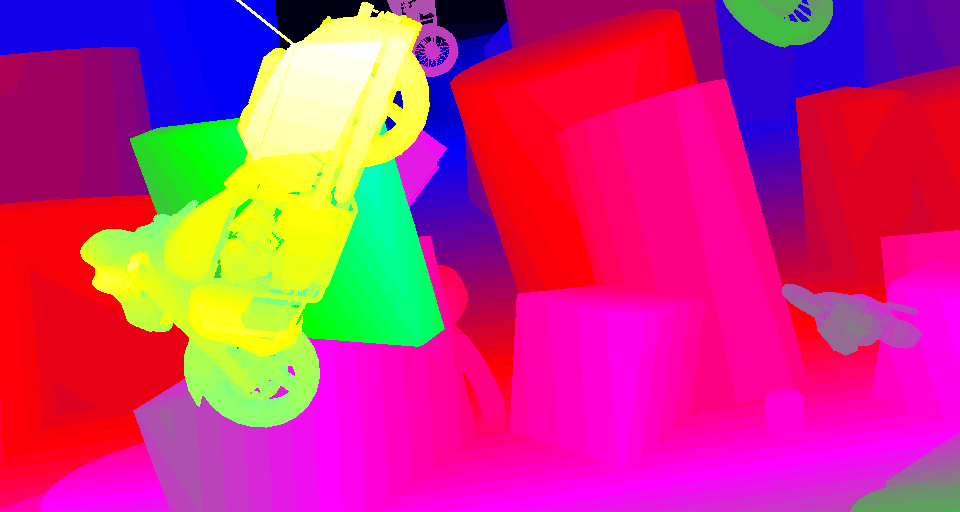}
		\vspace*{-0.6cm}
		\caption{\footnotesize{Left image/Disparity}}
	\end{subfigure}
	\hspace{-1.4\baselineskip}
	\hfill
	\begin{subfigure}[c]{.33\linewidth}
		\includegraphics[width=1\linewidth]{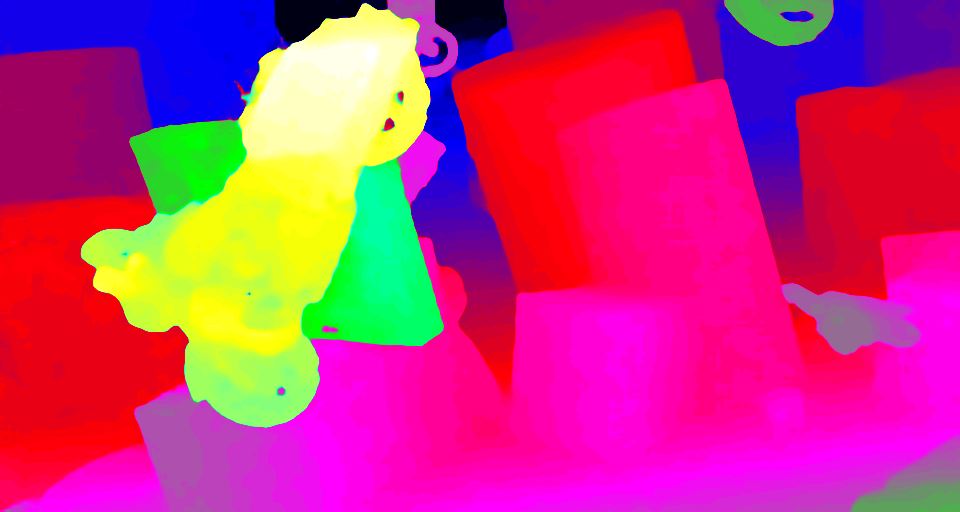}
		\vspace*{-0.6cm}
		\caption{\footnotesize{2D-MobileStereoNet}}
	\end{subfigure}
	\hspace{-1.4\baselineskip}
	\hfill
	\begin{subfigure}[c]{.33\linewidth}
		\includegraphics[width=1\linewidth]{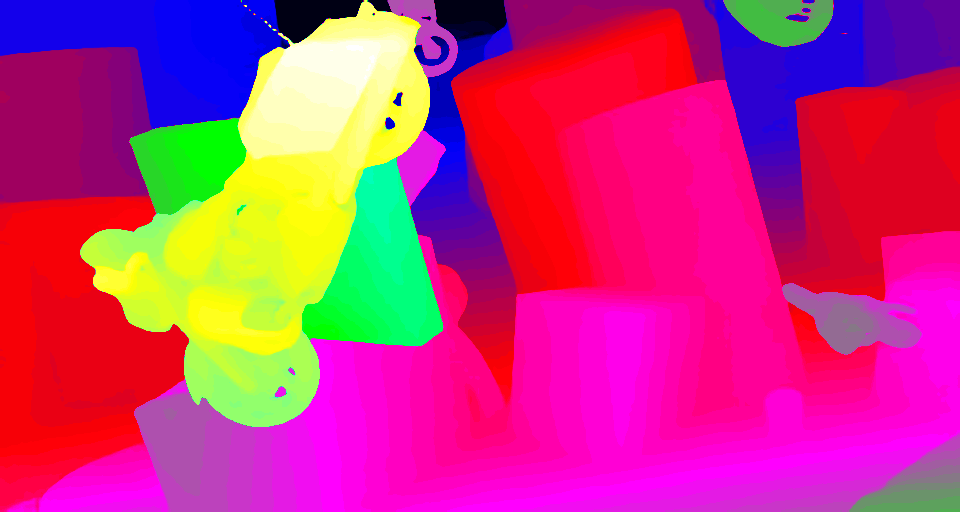}
		\vspace*{-0.6cm}
		\caption{\footnotesize{3D-MobileStereoNet}}
	\end{subfigure}
	\\
	\vspace{-0.3cm}
	\caption{Qualitative performance on SceneFlow: Every two rows correspond to a test sample. In the left-most column, the samples and the ground-truth disparity maps are illustrated. The following two columns show the disparity and error maps (embedded with error values) estimated by 2D-MobileStereoNet and 3D-MobileStereoNet.}
	\label{fig:SceneFlow}
	\vspace*{-0.3cm}
\end{figure}
\begin{table}[tbp]
	\captionsetup[subtable]{labelformat=empty}
	\centering
	\begin{center} 
		\begin{subtable}{1\linewidth}
			\footnotesize
			\centering		
			\begin{tabular}{@{\hskip2pt}l@{\hskip2pt}@{\hskip2pt}|@{\hskip2pt}S[table-format=1.2]@{\hskip2pt}S[table-format=2.2]@{\hskip2pt}|@{\hskip2pt}c@{\hskip2pt}}	
				\hline
				{Method} & {EPE($px$)} & {Params($M$)} & {Red. Params} \\
				\hline
				DispNet-C\cite{ilg2018occlusions}         & 1.67   & 38    & 17.0x \\				
				CRL\cite{pang2017cascade}	   	          & 1.60   & 78.77 & 35.3x \\				
				AutoDispNet-C\cite{saikia2019autodispnet}	  & 1.51    & 37    & 16.6x \\
				iResNet\cite{liang2018learning}	          & 1.40    & 43.11 & 19.3x \\\hline
				2D-MobileStereoNet                        & {\textbf{1.14}}  & {\textbf{2.23}}  & {-}\\\hline		
			\end{tabular}
			\vspace*{-0.2cm}
			\caption{\footnotesize{(a) 2D models}}
			\label{tab:sceneflow2d}
			\footnotesize
			\centering
			\begin{tabular}{@{\hskip2pt}l@{\hskip2pt}@{\hskip2pt}|@{\hskip2pt}c@{\hskip2pt}c@{\hskip2pt}c@{\hskip2pt}|@{\hskip2pt}c@{\hskip2pt}c@{\hskip2pt}}		
				\hline
				{Method} & {EPE($px$)} & {MACs($G$)} & {Params($M$)} & \makecell{Red. \\ MACs}
				& \makecell{Red. \\ Params} \\
				\hline
				GCNet\cite{kendall2017end}      	   			& 1.84 & 718.01 & 3.18    & 4.7x & 1.8x\\
				PSMNet\cite{chang2018pyramid}		  			 & 0.88 & 256.66 & 5.22    & 1.7x & 2.9x\\
				GA-Net-deep\cite{zhang2019ga} 		   			& 0.84 & 670.25 & 6.58    & 4.4x & 3.7x\\
				GA-Net-11\cite{zhang2019ga}			  		 & 0.93 & 383.42 & 4.48    & 2.5x & 2.5x\\
				Gwc40-Cat24-Base\cite{guo2019group}    		& 1.12 & 169.42 & 4.60    & 1.1x & 2.6x\\
				GwcNet-gc\cite{guo2019group} 	 	   		& \textbf{0.76} & 260.49 & 6.82    & 1.7x & 3.9x\\  
				GwcNet-g\cite{guo2019group}			   		& 0.79 & 246.27 & 6.43    & 1.6x & 3.6x\\
				DeepPruner-Best\cite{duggal2019deeppruner} & 0.86 & 129.23 & 7.39& 0.8x & 4.2x\\ 
				DeepPruner-Fast\cite{duggal2019deeppruner} & 0.97 & \textbf{51.83} & 7.47& 0.3x & 4.2x\\ \hline
				3D-MobileStereoNet           				& 0.80 & 153.14 & \textbf{1.77}  & - & - \\ \hline 
			\end{tabular}
			\vspace*{-0.2cm}
			\caption{\footnotesize{(b) 3D models}}
			\label{tab:sceneflow3d}	
		\end{subtable}
	\end{center}
	\vspace*{-0.7cm}
	\caption{Comparison on SceneFlow test set for (a) 2D and (b) 3D models. ``Red.'' indicates the reduction factor of our models compared to other methods.}
	\vspace*{-0.5cm}
\end{table}

\textbf{SceneFlow dataset.} The evaluation of the proposed models on SceneFlow are presented in Tables \ref{tab:sceneflow2d} and \ref{tab:sceneflow3d}. Note that 2D-MobileStereoNet has more parameters than 3D-MobileStereoNet due to $i)$ channel reduction module, $ii)$ parameterized cost volume construction, and $iii)$ wider hourglass. Nevertheless, with the least operations, it can be practical on systems with limited computation capacities. Compared to other 2D models, a lower error is achieved by 2D-MobileStereoNet with 16.6x fewer parameters. Moreover, we observe that in 3D-MobileStereoNet, significantly fewer parameters are achieved, while the performance is competitive with or better than other methods. Compared to GwcNet-gc with the best EPE metric in the table, 3D-MobileStereoNet uses 1.7x fewer parameters (in millions) and 3.9x fewer GigaMACs. Note that although DeepPruner-Fast \cite{duggal2019deeppruner} obtains the least number of operations, it is still over-parametrized and this causes issues when finetuning on smaller datasets like KITTI (\emph{cf.} Table \ref{tab:3DKITTI2015}). Figure \ref{fig:SceneFlow} shows the disparity estimation results.

\textbf{KITTI 2015 dataset.} This dataset consists of images of real-world driving scenarios \cite{menze2015object}, with $376\times1236$ resolution. To evaluate our models on this dataset, which has only ground-truth available for 200 for training samples, we use a 160/40 training/validation split. We finetune the networks that are pretrained on SceneFlow. For a fair comparison, we also train and evaluate other methods with the same data split. As shown in Tab. \ref{tab:kitti2015}, 2D-MobileStereoNet attains comparable results to PSMNet, which is a 3D model, with much less computational load (2.3x/8x fewer parameters/operations). Also, 3D-MobileStereoNet is outperforming PSMNet and GA-Net-11, and is competitive with GA-Net-deep. Compared to GwcNet-g, 3D-MobileStereoNet is lighter with 3.6x/1.6x fewer parameters/operations.
\begin{figure*}[htpb]
	\captionsetup[subfigure]{labelformat=empty}
	\centering
	\begin{center}
		\begin{minipage}[c]{0.01\textwidth}
			\begin{turn}{90}\footnotesize{}\end{turn}			
		\end{minipage}
		\begin{minipage}[c]{0.985\textwidth}
			\begin{subfigure}[c]{0.33\linewidth}
				\includegraphics[width=0.9\linewidth]{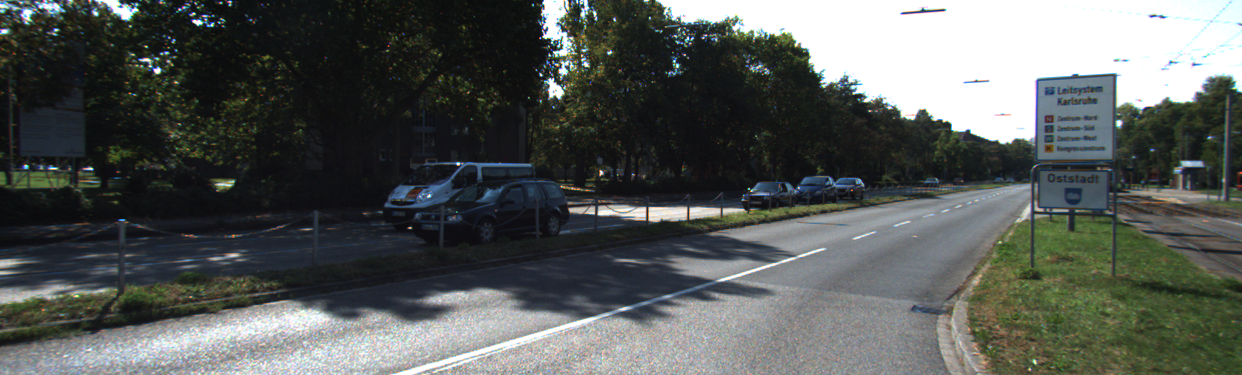}
			\end{subfigure}
			\hspace{-0.5\baselineskip}
			\hfill
			\begin{subfigure}[c]{0.33\linewidth}
				\includegraphics[width=0.9\linewidth]{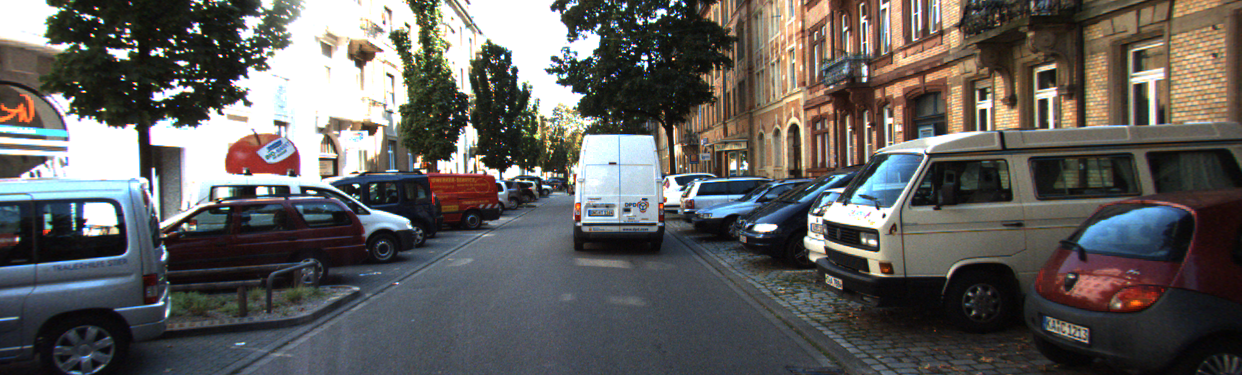}
			\end{subfigure}
			\hspace{-0.5\baselineskip}
			\hfill
			\begin{subfigure}[c]{0.33\linewidth}
				\includegraphics[width=0.9\linewidth]{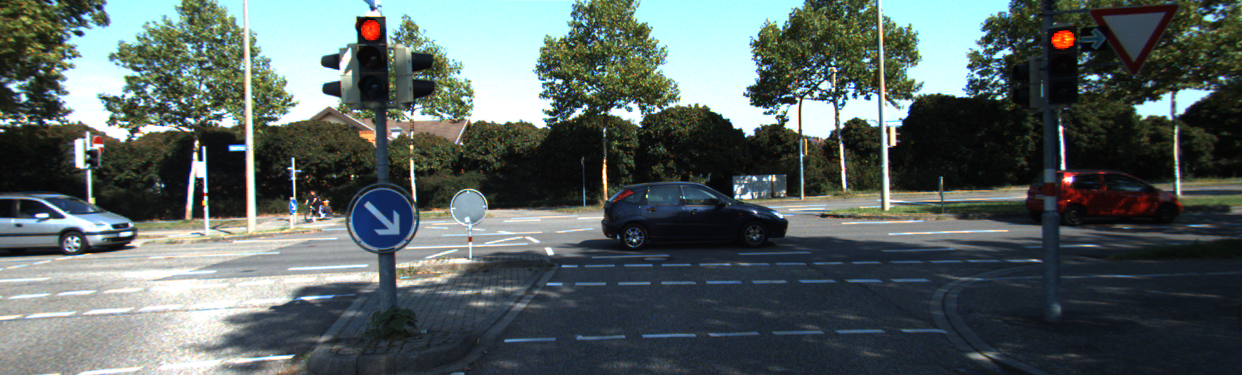}
			\end{subfigure}
		\end{minipage}
		\\
		\begin{minipage}[c]{0.01\textwidth}
			\begin{turn}{90}\footnotesize{GCNet \cite{kendall2017end}}\end{turn}			
		\end{minipage}
		\begin{minipage}[c]{0.985\textwidth}
			\begin{subfigure}[c]{0.33\linewidth}
				\includegraphics[width=0.9\linewidth]{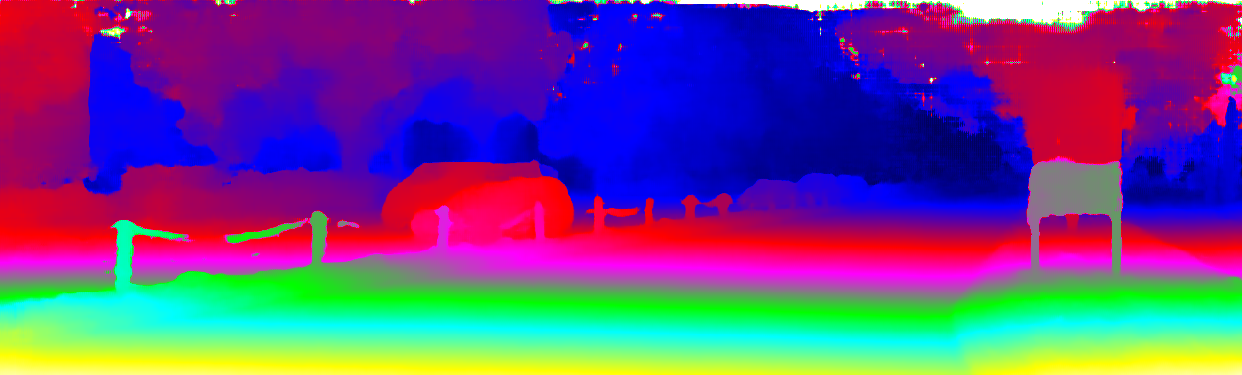}
			\end{subfigure}
			\hspace{-0.5\baselineskip}
			\hfill
			\begin{subfigure}[c]{0.33\linewidth}
				\includegraphics[width=0.9\linewidth]{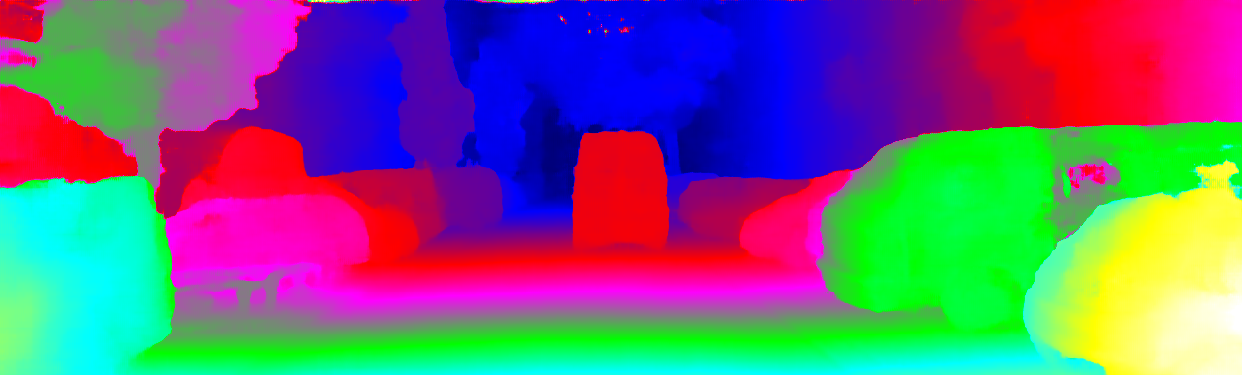}
			\end{subfigure}
			\hspace{-0.5\baselineskip}
			\hfill
			\begin{subfigure}[c]{0.33\linewidth}
				\includegraphics[width=0.9\linewidth]{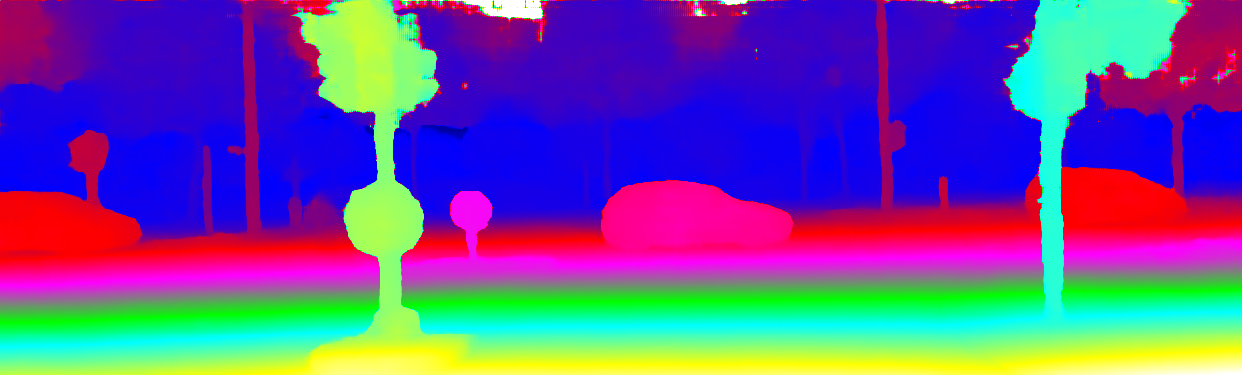}
			\end{subfigure}
			\begin{subfigure}[c]{0.33\linewidth}
				\includegraphics[width=0.9\linewidth]{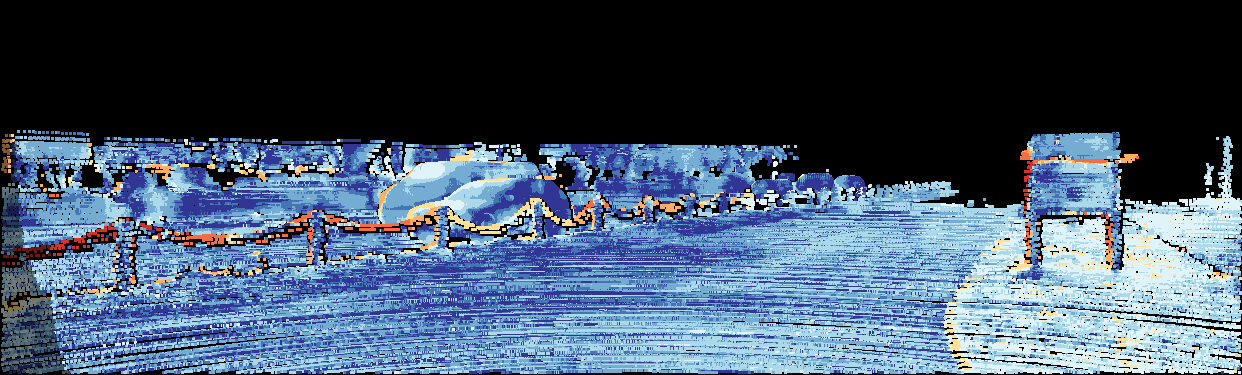}
			\end{subfigure}
			\hspace{-0.5\baselineskip}
			\hfill
			\begin{subfigure}[c]{0.33\linewidth}
				\includegraphics[width=0.9\linewidth]{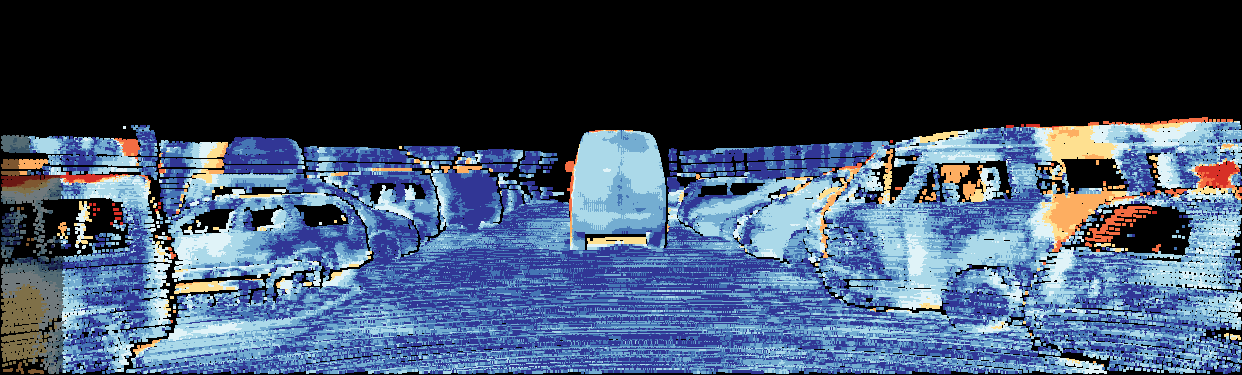}
			\end{subfigure}
			\hspace{-0.5\baselineskip}
			\hfill
			\begin{subfigure}[c]{0.33\linewidth}
				\includegraphics[width=0.9\linewidth]{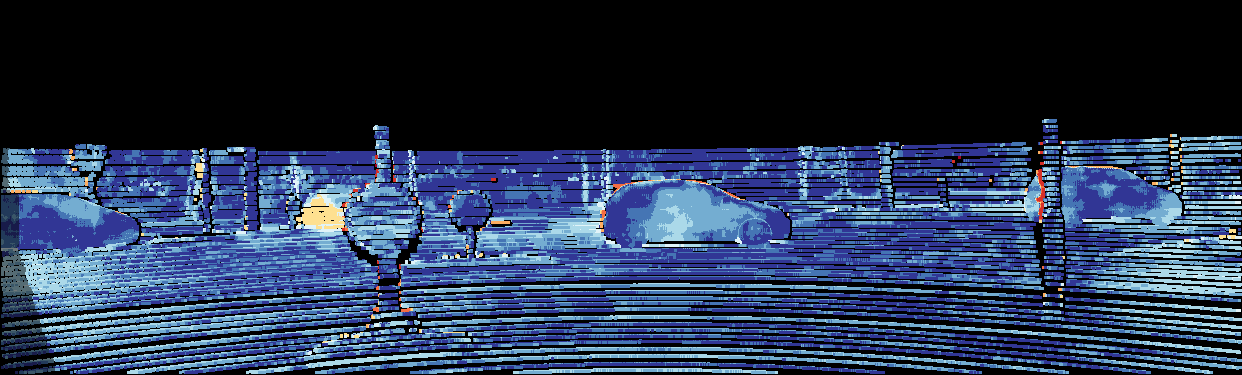}
			\end{subfigure}
		\end{minipage}
		\\
		\begin{minipage}[c]{0.01\textwidth}
			\begin{turn}{90}\footnotesize{PSMNet \cite{chang2018pyramid}}\end{turn}			
		\end{minipage}
		\begin{minipage}[c]{0.985\textwidth}
			\begin{subfigure}[c]{0.33\linewidth}
				\includegraphics[width=0.9\linewidth]{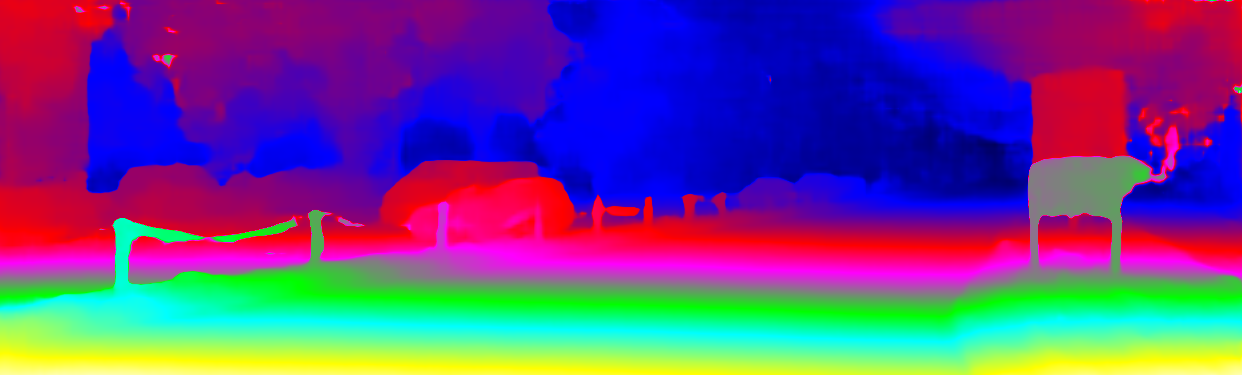}
			\end{subfigure}
			\hspace{-0.5\baselineskip}
			\hfill
			\begin{subfigure}[c]{0.33\linewidth}
				\includegraphics[width=0.9\linewidth]{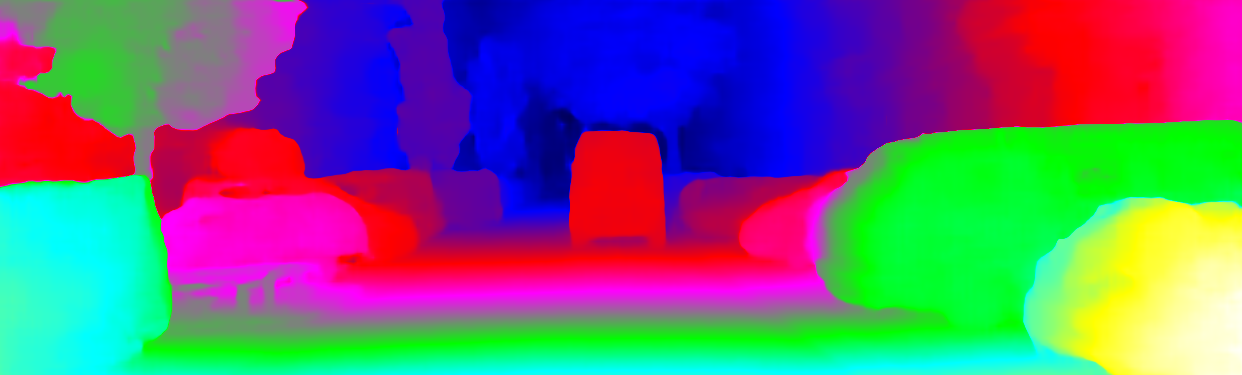}
			\end{subfigure}
			\hspace{-0.5\baselineskip}
			\hfill
			\begin{subfigure}[c]{0.33\linewidth}
				\includegraphics[width=0.9\linewidth]{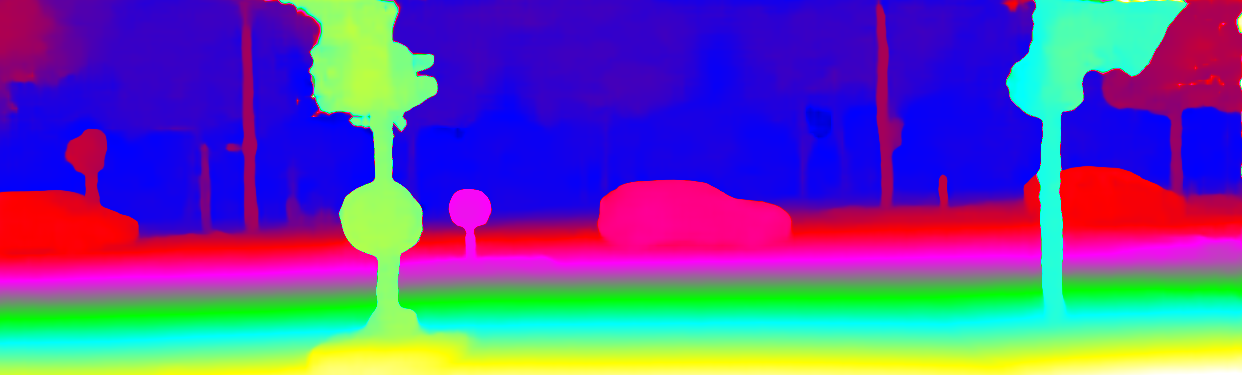}
			\end{subfigure}
			\begin{subfigure}[c]{0.33\linewidth}
				\includegraphics[width=0.9\linewidth]{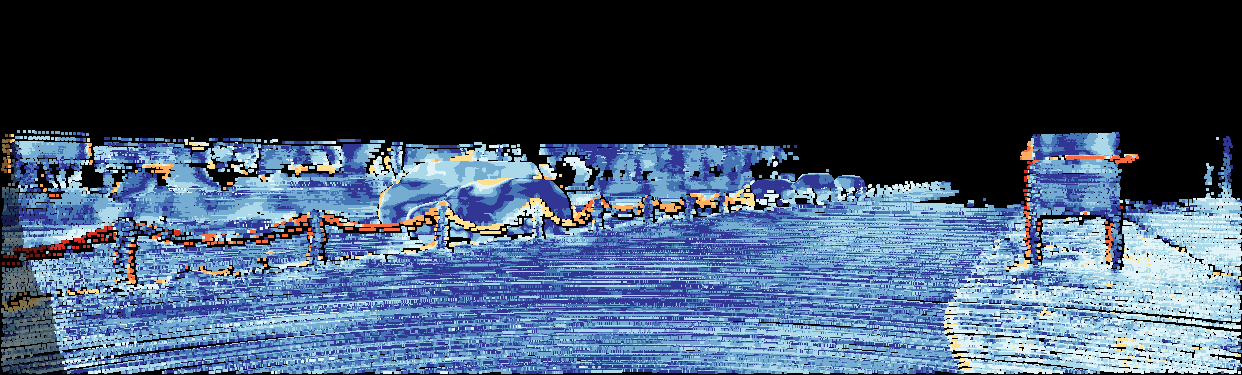}
			\end{subfigure}
			\hspace{-0.5\baselineskip}
			\hfill
			\begin{subfigure}[c]{0.33\linewidth}
				\includegraphics[width=0.9\linewidth]{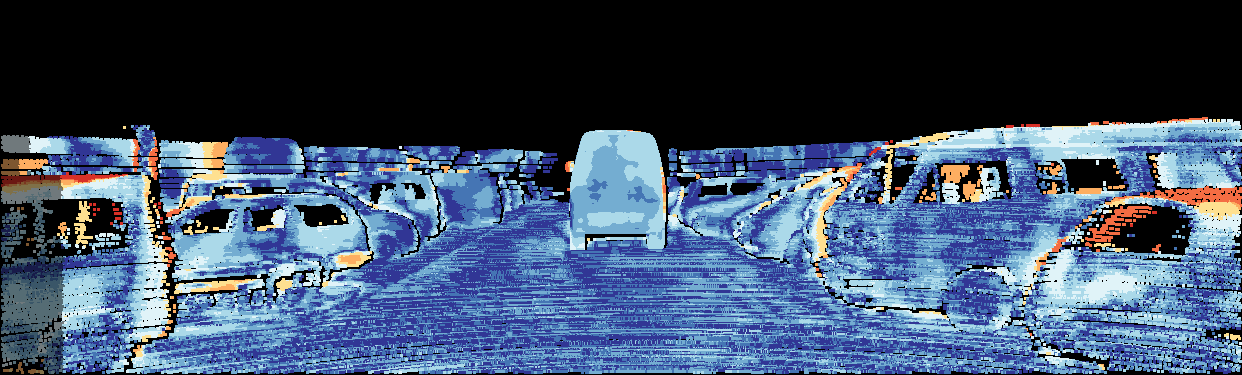}
			\end{subfigure}
			\hspace{-0.5\baselineskip}
			\hfill
			\begin{subfigure}[c]{0.33\linewidth}
				\includegraphics[width=0.9\linewidth]{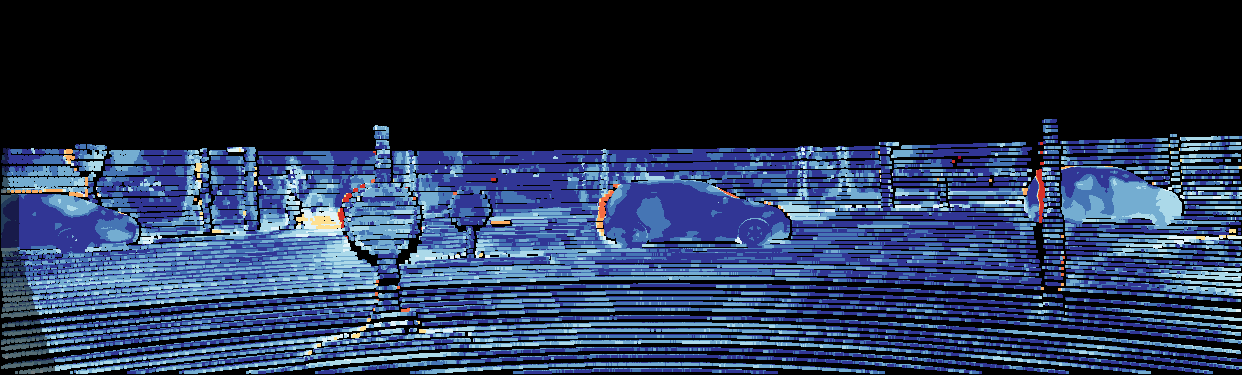}
			\end{subfigure}
		\end{minipage}
		\\
		\begin{minipage}[c]{0.01\textwidth}
			\begin{turn}{90}\footnotesize{GwcNet-g \cite{guo2019group}}\end{turn}			
		\end{minipage}
		\begin{minipage}[c]{0.985\textwidth}
			\begin{subfigure}[c]{0.33\linewidth}
				\includegraphics[width=0.9\linewidth]{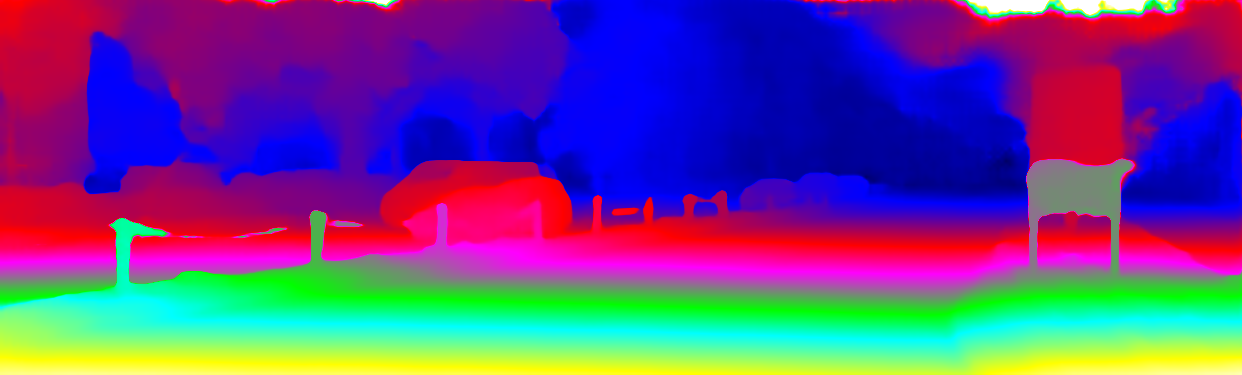}
			\end{subfigure}
			\hspace{-0.5\baselineskip}
			\hfill
			\begin{subfigure}[c]{0.33\linewidth}
				\includegraphics[width=0.9\linewidth]{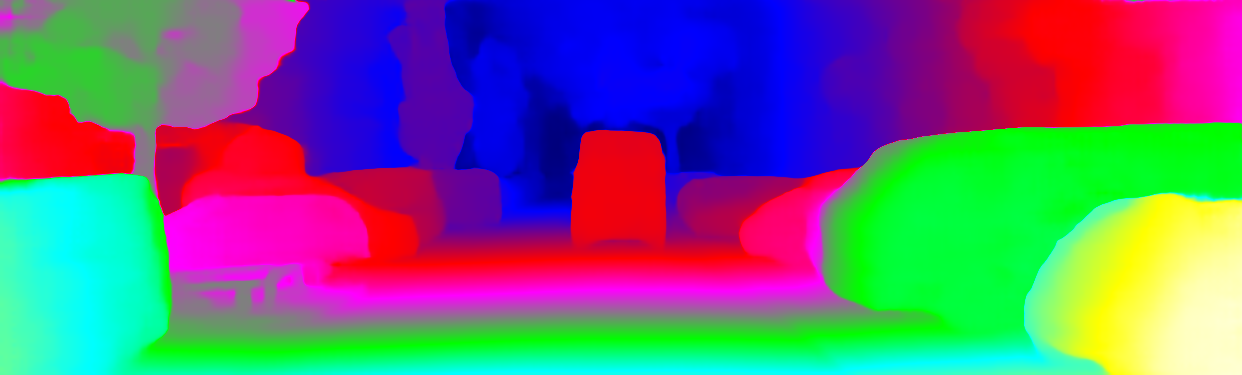}
			\end{subfigure}
			\hspace{-0.5\baselineskip}
			\hfill
			\begin{subfigure}[c]{0.33\linewidth}
				\includegraphics[width=0.9\linewidth]{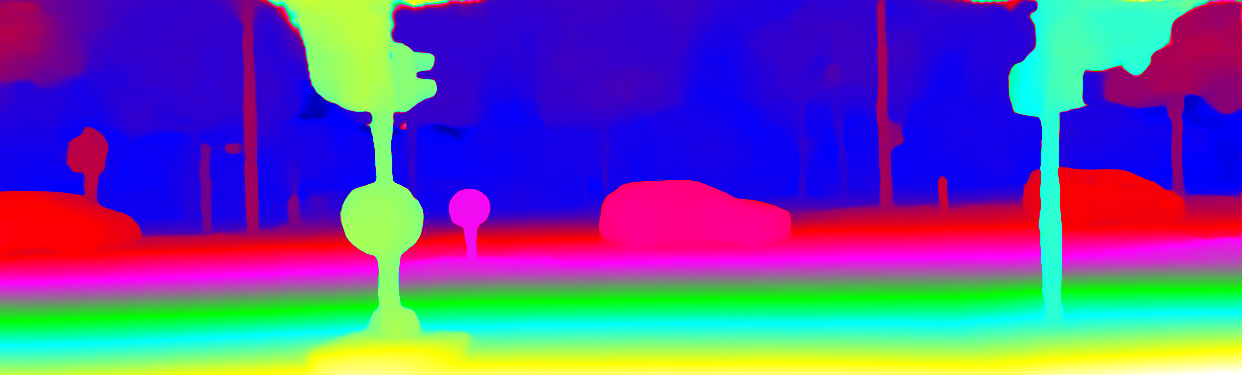}
			\end{subfigure}
			\begin{subfigure}[c]{0.33\linewidth}
				\includegraphics[width=0.9\linewidth]{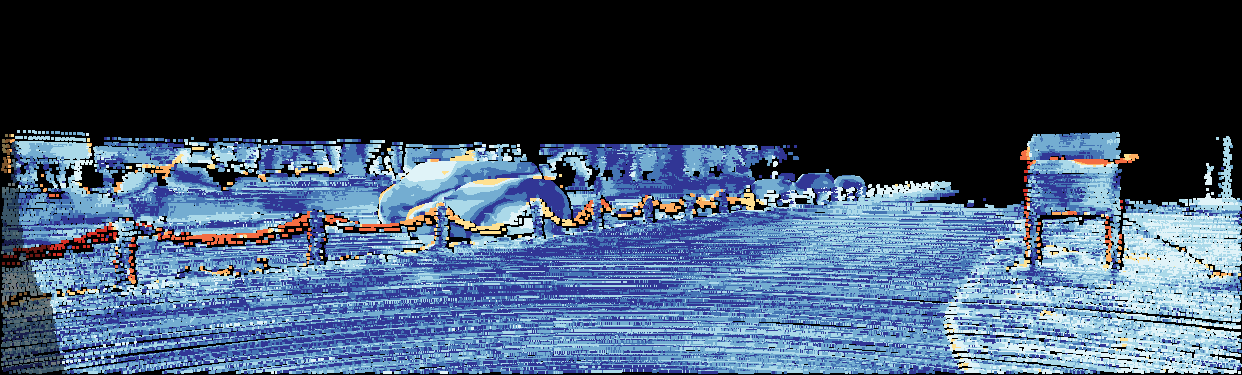}
			\end{subfigure}
			\hspace{-0.5\baselineskip}
			\hfill
			\begin{subfigure}[c]{0.33\linewidth}
				\includegraphics[width=0.9\linewidth]{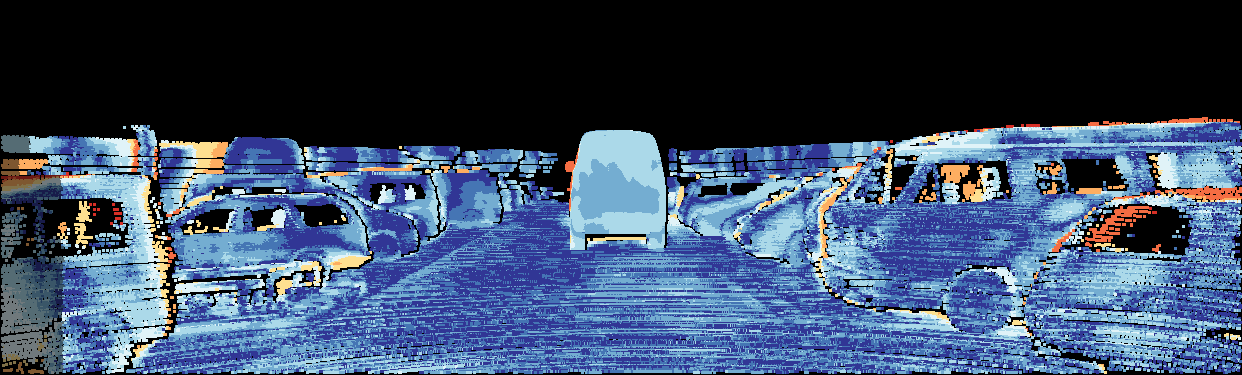}
			\end{subfigure}
			\hspace{-0.5\baselineskip}
			\hfill
			\begin{subfigure}[c]{0.33\linewidth}
				\includegraphics[width=0.9\linewidth]{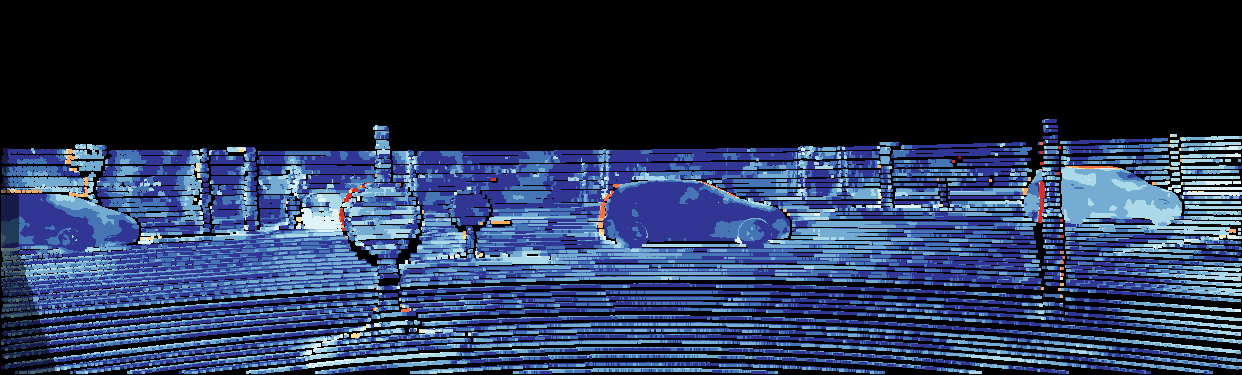}
			\end{subfigure}
		\end{minipage}
		\\
		\begin{minipage}[c]{0.01\textwidth}
			\begin{turn}{90}\footnotesize{2D-MobileStereoNet}\end{turn}			
		\end{minipage}
		\begin{minipage}[c]{0.985\textwidth}
			\begin{subfigure}[c]{0.33\linewidth}
				\includegraphics[width=0.9\linewidth]{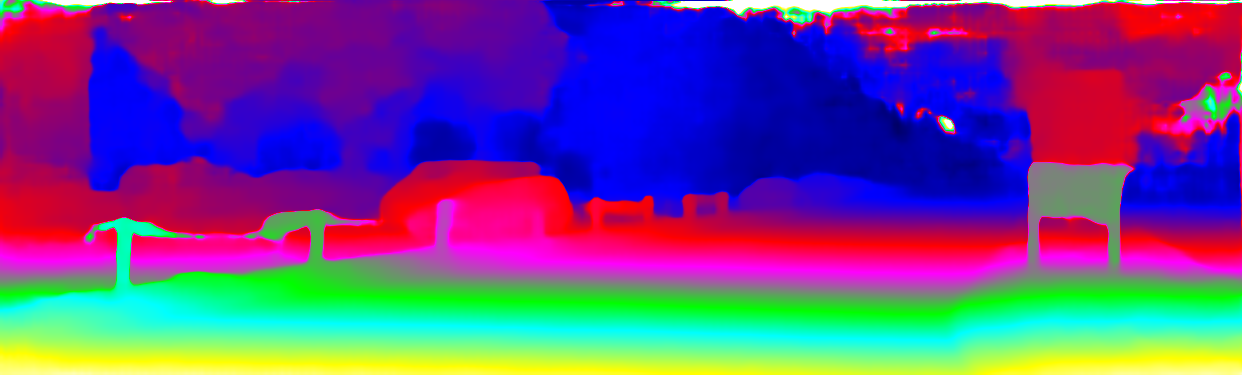}
			\end{subfigure}
			\hspace{-0.5\baselineskip}
			\hfill
			\begin{subfigure}[c]{0.33\linewidth}
				\includegraphics[width=0.9\linewidth]{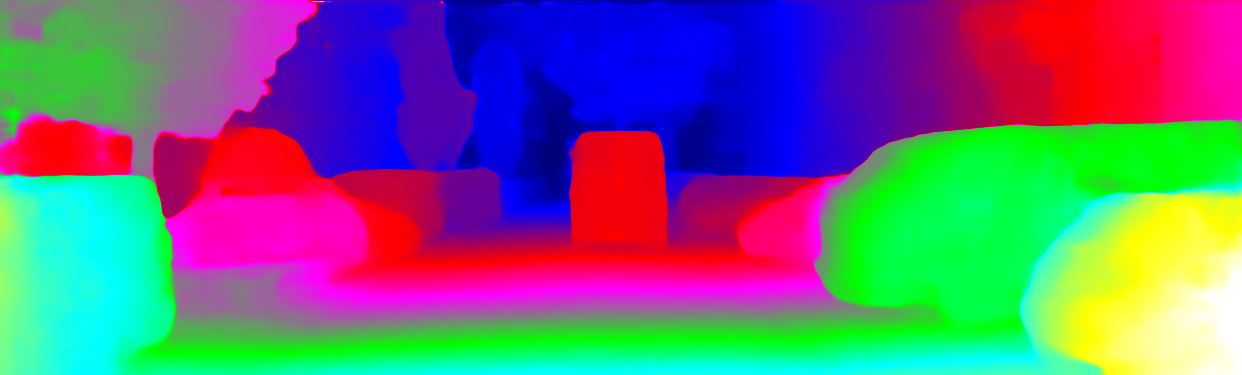}
			\end{subfigure}
			\hspace{-0.5\baselineskip}
			\hfill
			\begin{subfigure}[c]{0.33\linewidth}
				\includegraphics[width=0.9\linewidth]{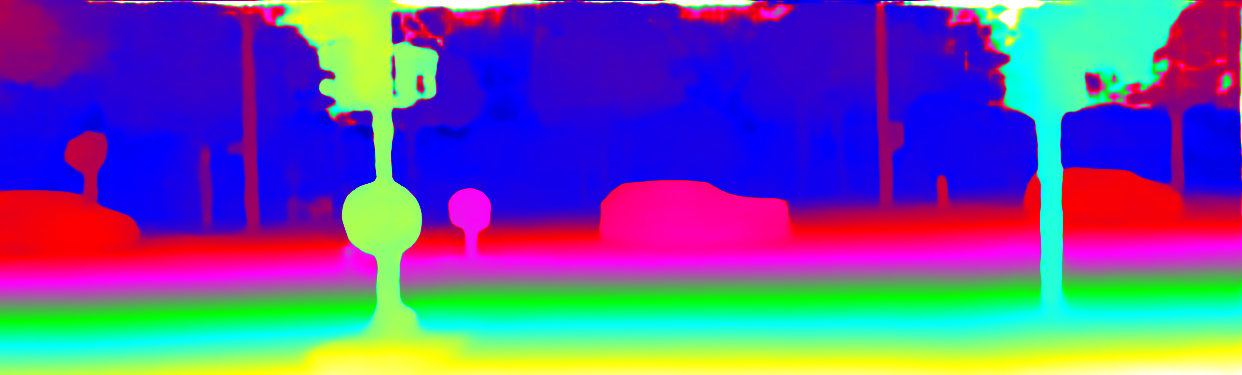}
			\end{subfigure}
			\begin{subfigure}[c]{0.33\linewidth}
				\includegraphics[width=0.9\linewidth]{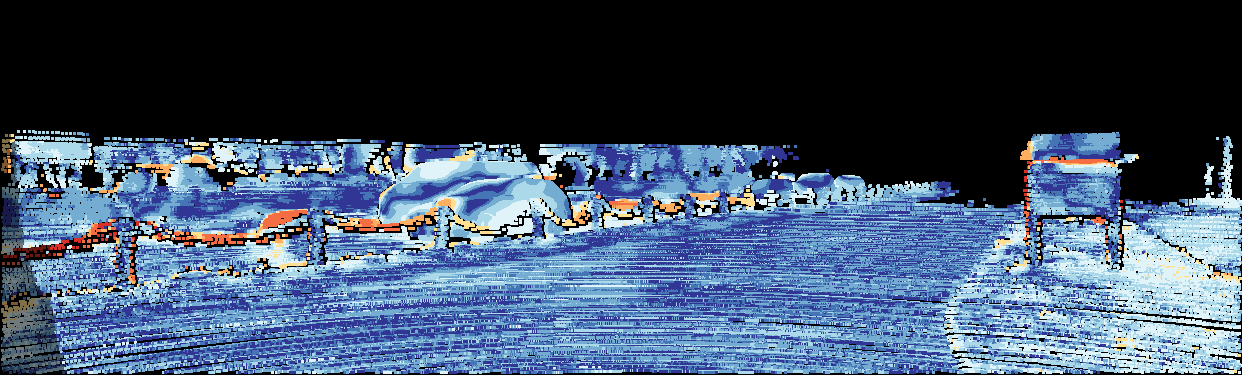}
			\end{subfigure}
			\hspace{-0.5\baselineskip}
			\hfill
			\begin{subfigure}[c]{0.33\linewidth}
				\includegraphics[width=0.9\linewidth]{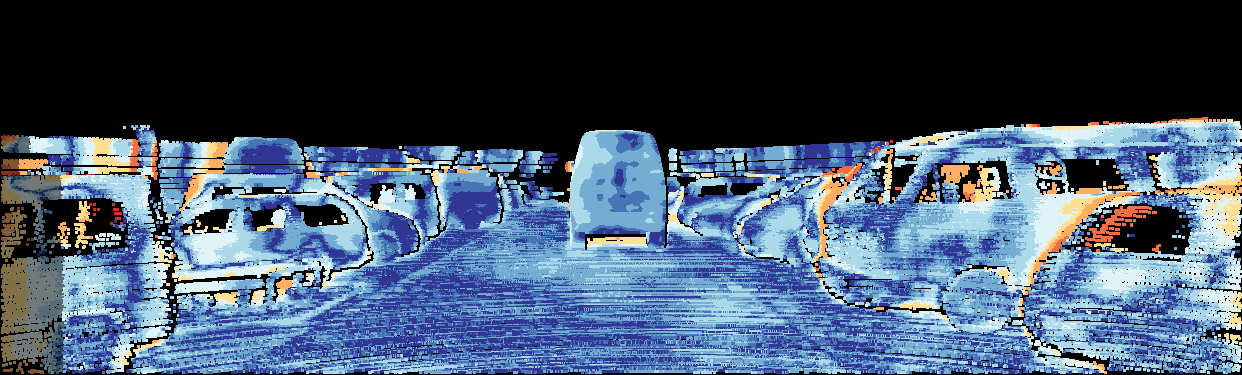}
			\end{subfigure}
			\hspace{-0.5\baselineskip}
			\hfill
			\begin{subfigure}[c]{0.33\linewidth}
				\includegraphics[width=0.9\linewidth]{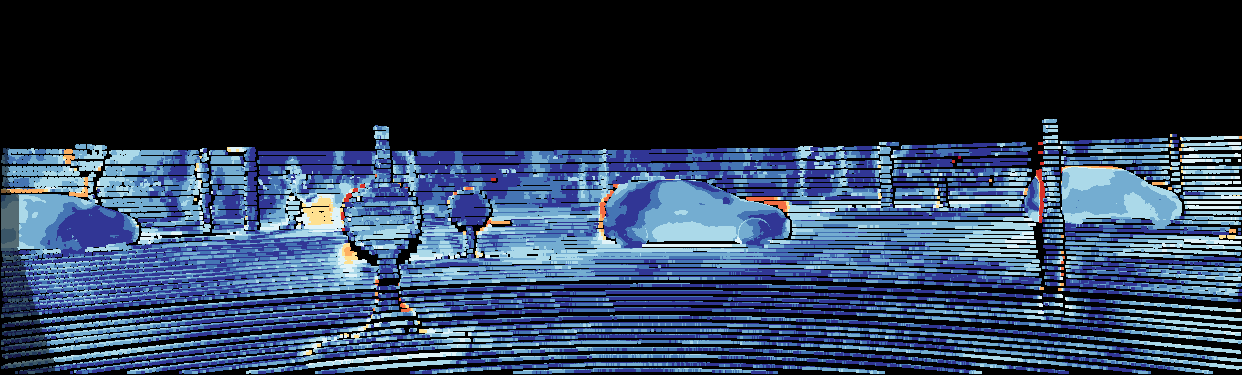}
			\end{subfigure}
		\end{minipage}
		\\
		\begin{minipage}[c]{0.01\textwidth}
			\begin{turn}{90}\footnotesize{3D-MobileStereoNet}\end{turn}			
		\end{minipage}
		\begin{minipage}[c]{0.985\textwidth}
			\begin{subfigure}[c]{0.33\linewidth}
				\includegraphics[width=0.9\linewidth]{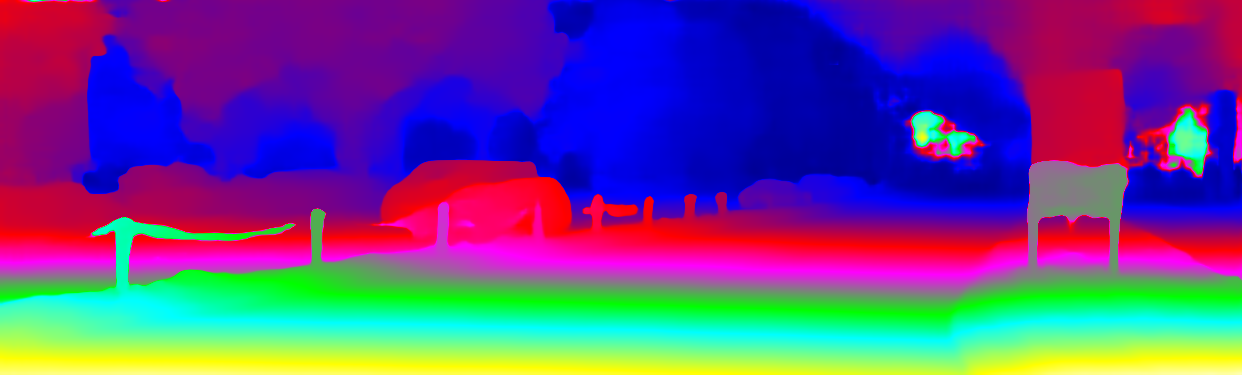}
			\end{subfigure}
			\hspace{-0.5\baselineskip}
			\hfill
			\begin{subfigure}[c]{0.33\linewidth}
				\includegraphics[width=0.9\linewidth]{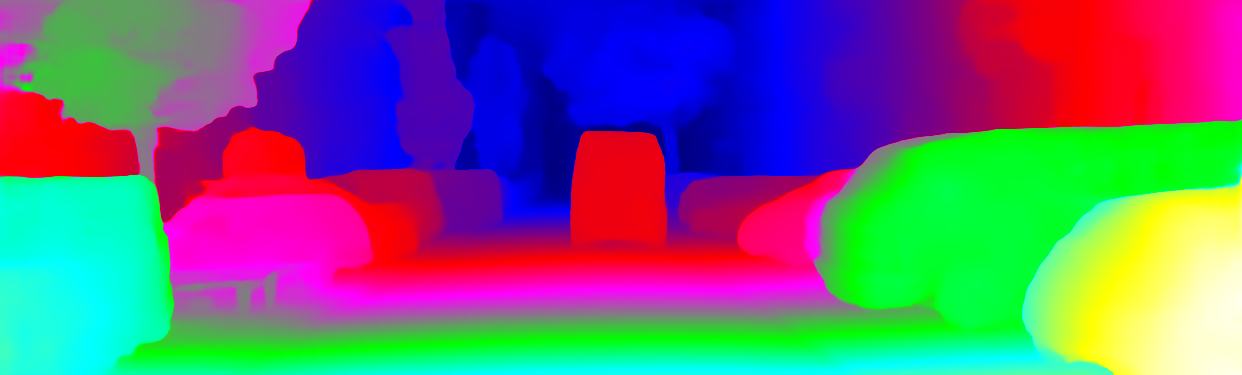}
			\end{subfigure}
			\hspace{-0.5\baselineskip}
			\hfill
			\begin{subfigure}[c]{0.33\linewidth}
				\includegraphics[width=0.9\linewidth]{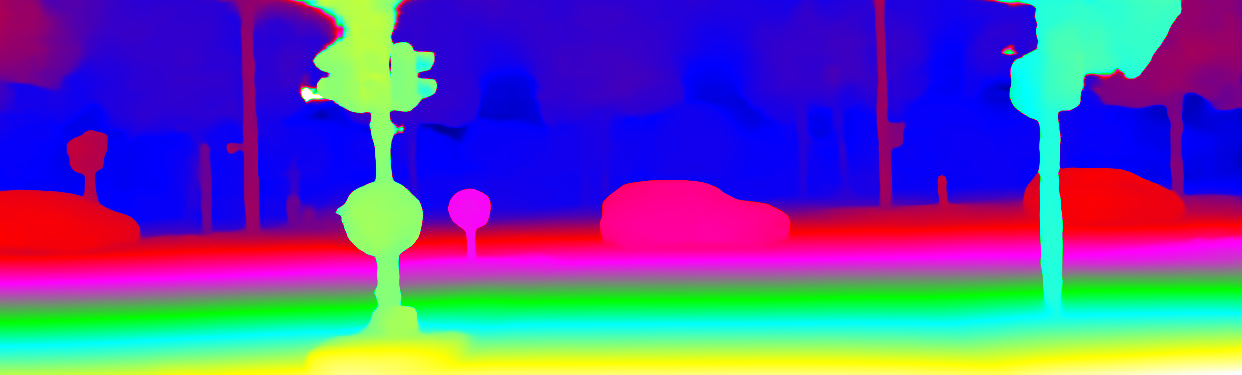}
			\end{subfigure}
			\begin{subfigure}[c]{0.33\linewidth}
				\includegraphics[width=0.9\linewidth]{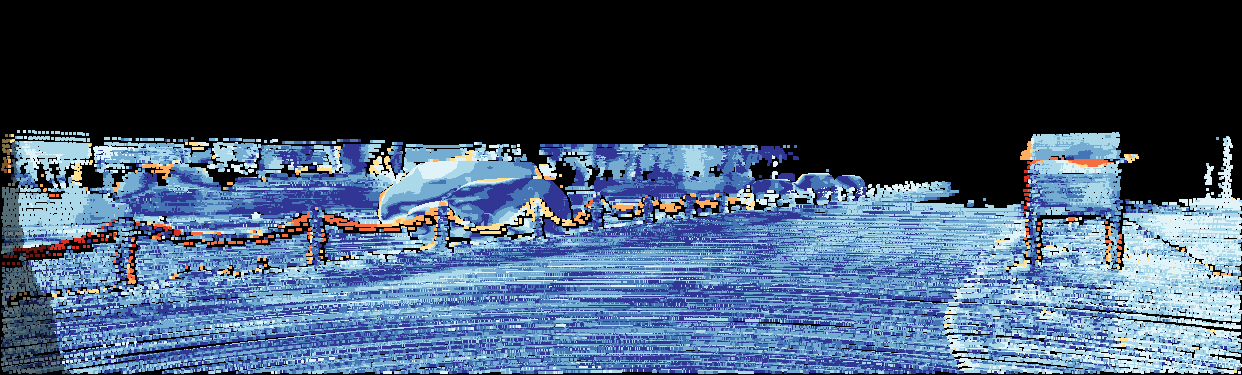}
			\end{subfigure}
			\hspace{-0.5\baselineskip}
			\hfill
			\begin{subfigure}[c]{0.33\linewidth}
				\includegraphics[width=0.9\linewidth]{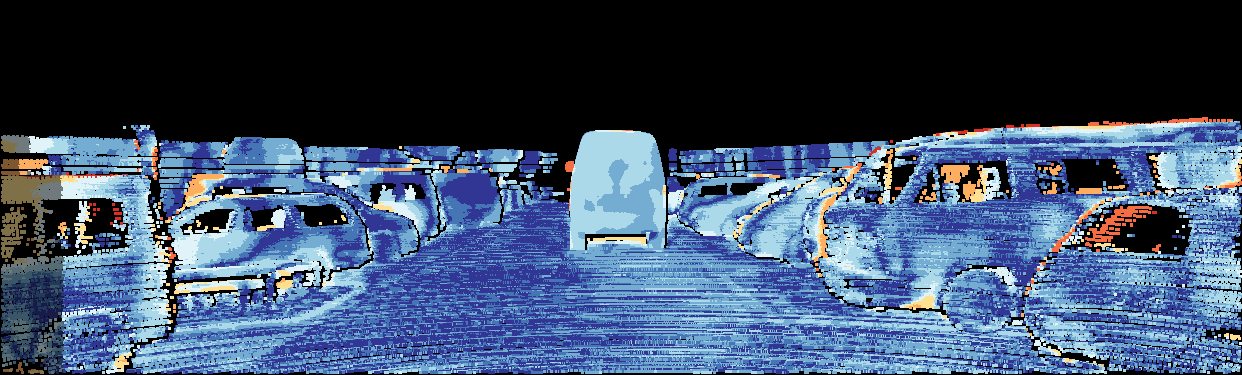}
			\end{subfigure}
			\hspace{-0.5\baselineskip}
			\hfill
			\begin{subfigure}[c]{0.33\linewidth}
				\includegraphics[width=0.9\linewidth]{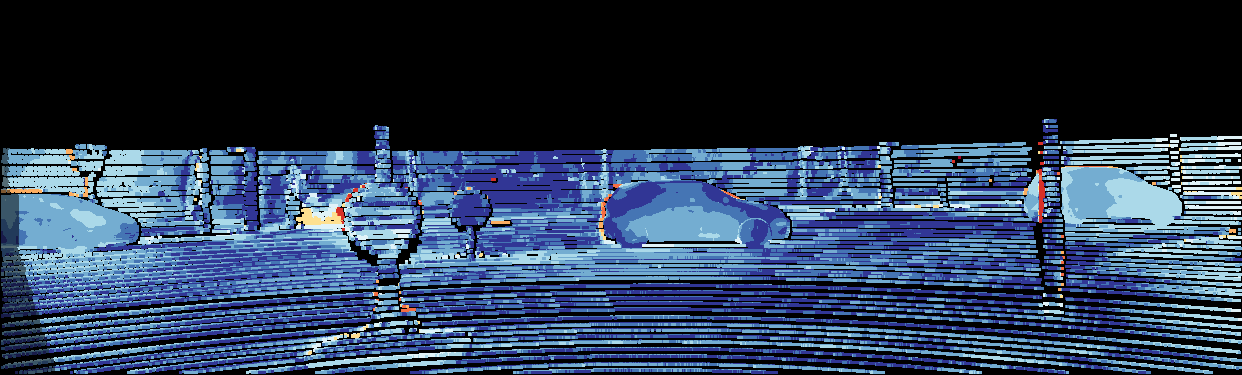}
			\end{subfigure}
		\end{minipage}
		\\
		\vspace{-0.2cm}
		\caption{Qualitative performance (disparity images together with error maps) from KITTI 2015 benchmark. Warmer colors in error maps denote larger values.}
		\vspace{-0.3cm}
		\label{fig:KittiBenchmark}
	\end{center}
\end{figure*}
\begin{table}[tbp]
	\begin{center}
		\footnotesize
		\begin{tabular}{@{\hskip2pt}l@{\hskip1pt}|@{\hskip2pt}c@{\hskip2pt}c@{\hskip2pt}c@{\hskip2pt}|@{\hskip2pt}c@{\hskip2pt}c@{\hskip2pt}}
			\hline				
			{Method} & {EPE($px$)} & {D1(\%)} & {px-3(\%)} & {MACs($G$)} & {Params($M$)} \\				
			\hline				
			PSMNet\cite{chang2018pyramid} & 0.88 & 2.00 & 2.10 & 256.66 & 5.22 \\
			GA-Net-deep\cite{zhang2019ga} & 0.63 & 1.61 & 1.67 & 670.25 & 6.58 \\
			GA-Net-11\cite{zhang2019ga}  & 0.67 & 1.92 & 2.01 & 383.42 & 4.48 \\
			GwcNet-gc\cite{guo2019group} & 0.63 & 1.55 & 1.60 & 260.49 & 6.82 \\
			GwcNet-g\cite{guo2019group}  & \textbf{0.62} & \textbf{1.49} & \textbf{1.53} & 246.27 & 6.43 \\\hdashline
			2D-MobileStereoNet 		 & 0.79 & 2.53 & 2.67 & \textbf{32.2}  & 2.32 \\
			3D-MobileStereoNet      & 0.66 & 1.59 & 1.69 & 153.14 & \textbf{1.77}\\				
			\hline				
		\end{tabular}
	\end{center}
	\vspace*{-0.65cm}
	\caption{Comparison on KITTI 2015 validation set.}
	\label{tab:kitti2015}
	\vspace*{-0.5cm}	
\end{table}
\begin{table}[tbp]
	\begin{center}
		\footnotesize
		\begin{tabular}{@{\hskip5pt}l@{\hskip5pt}|@{\hskip5pt}c@{\hskip5pt}c@{\hskip5pt}c@{\hskip5pt}|@{\hskip5pt}c@{\hskip5pt}c@{\hskip5pt}c@{\hskip5pt}}
			\hline
			\multirow{2}{*}{Methods} & \multicolumn{3}{c|}{All(\%)} & \multicolumn{3}{c}{Noc(\%)}  \\	 \cline{2-7} 
			& {D1$_{bg}$} & {D1$_{fg}$} & {D1$_{all}$} & {D1$_{bg}$} & {D1$_{fg}$} & {D1$_{all}$}  \\ \hline
			MC-CNN\cite{zbontar2016stereo} & 2.89 & 8.88 &3.89 & 2.48 & 7.64& 3.33\\
			Fast DS-CS\cite{yee2020fast} & 2.83 &	4.31	& 3.08 & 2.53&	3.74&	2.73 \\
			GCNet\cite{kendall2017end} & 2.21 & 6.16 & 2.87 & 2.02 & 5.58 & 2.61 \\
			DeepPruner-Fast\cite{duggal2019deeppruner} & 2.32 & 3.91 & 2.59 & 2.13 & 3.43 & 2.35 \\ 
			PSMNet\cite{chang2018pyramid} & 1.86 & 4.62 & 2.32 & 1.71 & 4.31 & 2.14 \\
			AutoDispNet-CSS\cite{saikia2019autodispnet} &1.94	&\textbf{3.37}	&2.18 &1.80	& \textbf{2.98}	&2.00\\
			DeepPruner-Best\cite{duggal2019deeppruner} & 1.87 & 3.56 & 2.15 & 1.71 & 3.18 & 1.95 \\ 
			GwcNet-g\cite{guo2019group} & \textbf{1.74} & 3.93 & 2.11 & \textbf{1.61} & 3.49 & \textbf{1.92} \\\hline
			2D-MobileStereoNet &  2.49 & 4.53 & 2.83 &  2.29 & 3.81 & 2.54  \\
			3D-MobileStereoNet &  1.75 & 3.87 & \textbf{2.10} &  \textbf{1.61}& 3.50 & \textbf{1.92}  \\ \hline
		\end{tabular}
	\end{center}
	\vspace*{-0.65cm}
	\caption{Comparison on KITTI 2015 benchmark. 3D-MobileStereoNet requires 98\% and 72\% fewer parameters compared to AutoDispNet-CSS and GwcNet-g, respectively.}
	\label{tab:3DKITTI2015}
	\vspace*{-0.55cm}	
\end{table}
\begin{table*}[t]
	\begin{center}
		\footnotesize	
		\begin{tabular}{l|c|c|c|c|c|c|c}
			\hline		
			{Method} & GA-Net-deep\cite{zhang2019ga} & 	GA-Net-11\cite{zhang2019ga} & GwcNet-gc\cite{guo2019group}&	 GwcNet-g\cite{guo2019group}& PSMNet\cite{chang2018pyramid} &2D-MobileStereoNet & 3D-MobileStereoNet \\ \hline
			{Memory (MB)} &26.4&18.0& 27.9 & 26.3 & 21.1 & 10.03 & 7.99\\ \hline
		\end{tabular}
	\end{center}
	\vspace*{-0.65cm}
	\caption{Comparison of model size. Our two proposed methods yield more compact models compared to other works.}
	\label{tab:memorytime}
\end{table*}

Additionally, we submitted the results of our finetuned models to the KITTI 2015 benchmark. To this end, we finetuned the epoch with the best cross-domain generalizability from SceneFlow to KITTI 2015. Table \ref{tab:3DKITTI2015} shows that 2D-MobileStereoNet is surpassing GCNet (a 3D model) with 27\%/95\% fewer parameters/operations. Also, we can verify that 3D-MobileStereoNet shows superior performance when compared to GCNet and PSMNet and it is surpassing GwcNet-g (in D1$_{fg}$ and in D1$_{all}$ in all pixels) with 72\%/38\% fewer parameters/operations. 

Figure \ref{fig:KittiBenchmark} visualizes the results from KITTI 2015 benchmark. Comparing 2D-MobileStereoNet and 3D-MobileStereoNet, we observe that 3D-MobileStereoNet obtains crisp edges due to deploying 3D convolutions in the encoder-decoder. In other words, in the upsampling/downsampling process in encoder-decoder, 3D convolutions can better preserve finer details compared to an encoder-decoder with 2D convolutions. Nevertheless, 2D-MobileStereoNet achieves visually similar outputs to 3D models while requiring considerably fewer operations (87\% fewer operations compared to GwcNet-g). Also, 3D-MobileStereoNet visually achieves competitive or better results compared to other methods. 

\textbf{Model size.} We also report the memory requirement of our models in Table \ref{tab:memorytime}. Both of the proposed methods show smaller memory sizes, which is promising for memory-constrained chips and their power consumption.
\vspace{-0.4cm}
\section{Conclusion}
This paper presented lightweight stereo networks to alleviate high memory usage on embedded or mobile devices. Namely, we proposed two models (2D and 3D) with the primary goal of reducing the cost (in terms of parameters, operations, and model size) by using MobileNet blocks. To increase the accuracy of the 2D model, we also designed a new cost volume to \emph{learn} the similarity of unary features. Yielding a favorable accuracy/complexity trade-off, these MobileStereoNets are promising for deploying end-to-end stereo networks on edge devices.
\subsubsection*{Acknowledgment}
This work is supported by the German Federal Ministry of Education and Research (BMBF) for the project DeepStereoVision (FRE: 01I518024B).
{\small
	\bibliographystyle{ieee_fullname}
	\bibliography{egbib}

\begin{thebibliography}{10}\itemsep=-1pt

\bibitem{batsos2018cbmv}
Konstantinos Batsos, Changjiang Cai, and Philippos Mordohai.
\newblock {CBMV}: A coalesced bidirectional matching volume for disparity
  estimation.
\newblock In {\em Proceedings of the IEEE Conference on Computer Vision and
  Pattern Recognition}, pages 2060--2069, 2018.

\bibitem{chang2018pyramid}
Jia-Ren Chang and Yong-Sheng Chen.
\newblock Pyramid stereo matching network.
\newblock In {\em Proceedings of the IEEE Conference on Computer Vision and
  Pattern Recognition}, pages 5410--5418, 2018.

\bibitem{dosovitskiy2015flownet}
Alexey Dosovitskiy, Philipp Fischer, Eddy Ilg, Philip Hausser, Caner Hazirbas,
  Vladimir Golkov, Patrick Van Der~Smagt, Daniel Cremers, and Thomas Brox.
\newblock Flow{N}et: Learning optical flow with convolutional networks.
\newblock In {\em Proceedings of the IEEE International Conference on Computer
  Vision}, pages 2758--2766, 2015.

\bibitem{duggal2019deeppruner}
Shivam Duggal, Shenlong Wang, Wei-Chiu Ma, Rui Hu, and Raquel Urtasun.
\newblock {DeepPruner}: Learning efficient stereo matching via differentiable
  patchmatch.
\newblock In {\em Proceedings of the IEEE International Conference on Computer
  Vision}, pages 4384--4393, 2019.

\bibitem{feichtenhofer2020x3d}
Christoph Feichtenhofer.
\newblock {X3D}: Expanding architectures for efficient video recognition.
\newblock In {\em Proceedings of the IEEE Conference on Computer Vision and
  Pattern Recognition}, pages 203--213, 2020.

\bibitem{guo2019group}
Xiaoyang Guo, Kai Yang, Wukui Yang, Xiaogang Wang, and Hongsheng Li.
\newblock Group-wise correlation stereo network.
\newblock In {\em Proceedings of the IEEE Conference on Computer Vision and
  Pattern Recognition}, pages 3273--3282, 2019.

\bibitem{hirschmuller2005accurate}
Heiko Hirschmuller.
\newblock Accurate and efficient stereo processing by semi-global matching and
  mutual information.
\newblock In {\em Proceedings of the IEEE Conference on Computer Vision and
  Pattern Recognition}, volume~2, pages 807--814, 2005.

\bibitem{howard2017mobilenets}
Andrew~G Howard, Menglong Zhu, Bo Chen, Dmitry Kalenichenko, Weijun Wang,
  Tobias Weyand, Marco Andreetto, and Hartwig Adam.
\newblock {MobileNets}: Efficient convolutional neural networks for mobile
  vision applications.
\newblock {\em arXiv preprint arXiv:1704.04861}, 2017.

\bibitem{ilg2018occlusions}
Eddy Ilg, Tonmoy Saikia, Margret Keuper, and Thomas Brox.
\newblock Occlusions, motion and depth boundaries with a generic network for
  disparity, optical flow or scene flow estimation.
\newblock In {\em Proceedings of the European Conference on Computer Vision},
  pages 614--630, 2018.

\bibitem{kendall2017end}
Alex Kendall, Hayk Martirosyan, Saumitro Dasgupta, Peter Henry, Ryan Kennedy,
  Abraham Bachrach, and Adam Bry.
\newblock End-to-end learning of geometry and context for deep stereo
  regression.
\newblock In {\em Proceedings of the IEEE International Conference on Computer
  Vision}, pages 66--75, 2017.

\bibitem{liang2018learning}
Zhengfa Liang, Yiliu Feng, Yulan Guo, Hengzhu Liu, Wei Chen, Linbo Qiao, Li
  Zhou, and Jianfeng Zhang.
\newblock Learning for disparity estimation through feature constancy.
\newblock In {\em Proceedings of the IEEE Conference on Computer Vision and
  Pattern Recognition}, pages 2811--2820, 2018.

\bibitem{luo2016efficient}
Wenjie Luo, Alexander~G Schwing, and Raquel Urtasun.
\newblock Efficient deep learning for stereo matching.
\newblock In {\em Proceedings of the IEEE Conference on Computer Vision and
  Pattern Recognition}, pages 5695--5703, 2016.

\bibitem{mayer2016large}
Nikolaus Mayer, Eddy Ilg, Philip Hausser, Philipp Fischer, Daniel Cremers,
  Alexey Dosovitskiy, and Thomas Brox.
\newblock A large dataset to train convolutional networks for disparity,
  optical flow, and scene flow estimation.
\newblock In {\em Proceedings of the IEEE Conference on Computer Vision and
  Pattern Recognition}, pages 4040--4048, 2016.

\bibitem{menze2015object}
Moritz Menze and Andreas Geiger.
\newblock Object scene flow for autonomous vehicles.
\newblock In {\em Proceedings of the IEEE Conference on Computer Vision and
  Pattern Recognition}, pages 3061--3070, 2015.

\bibitem{pal2020looking}
Anwesan Pal, Sayan Mondal, and Henrik~I Christensen.
\newblock "looking at the right stuff"-guided semantic-gaze for autonomous
  driving.
\newblock In {\em Proceedings of the IEEE Conference on Computer Vision and
  Pattern Recognition}, pages 11883--11892, 2020.

\bibitem{pang2017cascade}
Jiahao Pang, Wenxiu Sun, Jimmy~SJ Ren, Chengxi Yang, and Qiong Yan.
\newblock Cascade residual learning: A two-stage convolutional neural network
  for stereo matching.
\newblock In {\em Proceedings of the IEEE International Conference on Computer
  Vision Workshops}, pages 887--895, 2017.

\bibitem{park2016look}
Haesol Park and Kyoung~Mu Lee.
\newblock Look wider to match image patches with convolutional neural networks.
\newblock {\em IEEE Signal Processing Letters}, 24(12):1788--1792, 2016.

\bibitem{park2015leveraging}
Min-Gyu Park and Kuk-Jin Yoon.
\newblock Leveraging stereo matching with learning-based confidence measures.
\newblock In {\em Proceedings of the IEEE Conference on Computer Vision and
  Pattern Recognition}, pages 101--109, 2015.

\bibitem{qiu2017learning}
Zhaofan Qiu, Ting Yao, and Tao Mei.
\newblock Learning spatio-temporal representation with pseudo-3d residual
  networks.
\newblock In {\em Proceedings of the IEEE International Conference on Computer
  Vision}, pages 5533--5541, 2017.

\bibitem{rahim2021}
Rafia Rahim, Faranak Shamsafar, and Andreas Zell.
\newblock Separable convolutions for optimizing 3d stereo networks.
\newblock In {\em Proceedings of the IEEE International Conference on Image
  Processing}, 2021.

\bibitem{rao_he_dai_zhu_li_he_2020}
Zhibo Rao, Mingyi He, Yuchao Dai, Zhidong Zhu, Bo Li, and Renjie He.
\newblock {NLCA-Net}: A non-local context attention network for stereo
  matching.
\newblock {\em APSIPA Transactions on Signal and Information Processing},
  9:e18, 2020.

\bibitem{saikia2019autodispnet}
Tonmoy Saikia, Yassine Marrakchi, Arber Zela, Frank Hutter, and Thomas Brox.
\newblock {AutoDispNet}: Improving disparity estimation with automl.
\newblock In {\em Proceedings of the IEEE International Conference on Computer
  Vision}, pages 1812--1823, 2019.

\bibitem{sandler2018mobilenetv2}
Mark Sandler, Andrew Howard, Menglong Zhu, Andrey Zhmoginov, and Liang-Chieh
  Chen.
\newblock {MobileNetv2}: Inverted residuals and linear bottlenecks.
\newblock In {\em Proceedings of the IEEE Conference on Computer Vision and
  Pattern Recognition}, pages 4510--4520, 2018.

\bibitem{seki2017sgm}
Akihito Seki and Marc Pollefeys.
\newblock {SGM-Nets}: Semi-global matching with neural networks.
\newblock In {\em Proceedings of the IEEE Conference on Computer Vision and
  Pattern Recognition}, pages 231--240, 2017.

\bibitem{shi2020point}
Weijing Shi and Raj Rajkumar.
\newblock {Point-GNN}: Graph neural network for 3d object detection in a point
  cloud.
\newblock In {\em Proceedings of the IEEE Conference on Computer Vision and
  Pattern Recognition}, pages 1711--1719, 2020.

\bibitem{wang2019normalized}
He Wang, Srinath Sridhar, Jingwei Huang, Julien Valentin, Shuran Song, and
  Leonidas~J Guibas.
\newblock Normalized object coordinate space for category-level 6d object pose
  and size estimation.
\newblock In {\em Proceedings of the IEEE Conference on Computer Vision and
  Pattern Recognition}, pages 2642--2651, 2019.

\bibitem{wu2019semantic}
Zhenyao Wu, Xinyi Wu, Xiaoping Zhang, Song Wang, and Lili Ju.
\newblock Semantic stereo matching with pyramid cost volumes.
\newblock In {\em Proceedings of the IEEE International Conference on Computer
  Vision}, pages 7484--7493, 2019.

\bibitem{ye20193d}
Rongtian Ye, Fangyu Liu, and Liqiang Zhang.
\newblock 3d depthwise convolution: Reducing model parameters in 3d vision
  tasks.
\newblock In {\em Canadian Conference on Artificial Intelligence}, pages
  186--199, 2019.

\bibitem{yee2020fast}
Kyle Yee and Ayan Chakrabarti.
\newblock Fast deep stereo with 2d convolutional processing of cost signatures.
\newblock In {\em Proceedings of the IEEE Winter Conference on Applications of
  Computer Vision}, pages 183--191, 2020.

\bibitem{zabih1994non}
Ramin Zabih and John Woodfill.
\newblock Non-parametric local transforms for computing visual correspondence.
\newblock In {\em Proceedings of the European Conference on Computer Vision},
  pages 151--158, 1994.

\bibitem{zbontar2016stereo}
Jure Zbontar, Yann LeCun, et~al.
\newblock Stereo matching by training a convolutional neural network to compare
  image patches.
\newblock {\em Journal of Machine Learning Research}, 17(1):2287--2318, 2016.

\bibitem{zhang2019ga}
Feihu Zhang, Victor Prisacariu, Ruigang Yang, and Philip~HS Torr.
\newblock {GA-Net}: Guided aggregation net for end-to-end stereo matching.
\newblock In {\em Proceedings of the IEEE Conference on Computer Vision and
  Pattern Recognition}, pages 185--194, 2019.

\end{thebibliography}
}
\clearpage
\newpage
\appendix
\setcounter{figure}{0}    
\setcounter{table}{0}
\onecolumn
\hspace{3.49 cm}
\begin{center}
	\title{\Large \textbf{MobileStereoNet: Towards Lightweight Deep Networks for Stereo Matching\\ \vspace{1cm}
			(Appendix)} }	
	\maketitle
\end{center}
\hspace{2cm}

This appendix provides more details of our work. Here, we have the following sections: \ref{baseline}: Detailed baseline architectures, \ref{Qualitative}: More qualitative results, \ref{Incorporating}: Incorporating light blocks in other modules, \ref{Implementation}: Implementation details, and \ref{Analyzing}: Analyzing the complexity.
\section{Detailed baseline architectures}
\label{baseline}
\begin{figure*}[h]
	\begin{center}
		\includegraphics[width=0.98\linewidth]{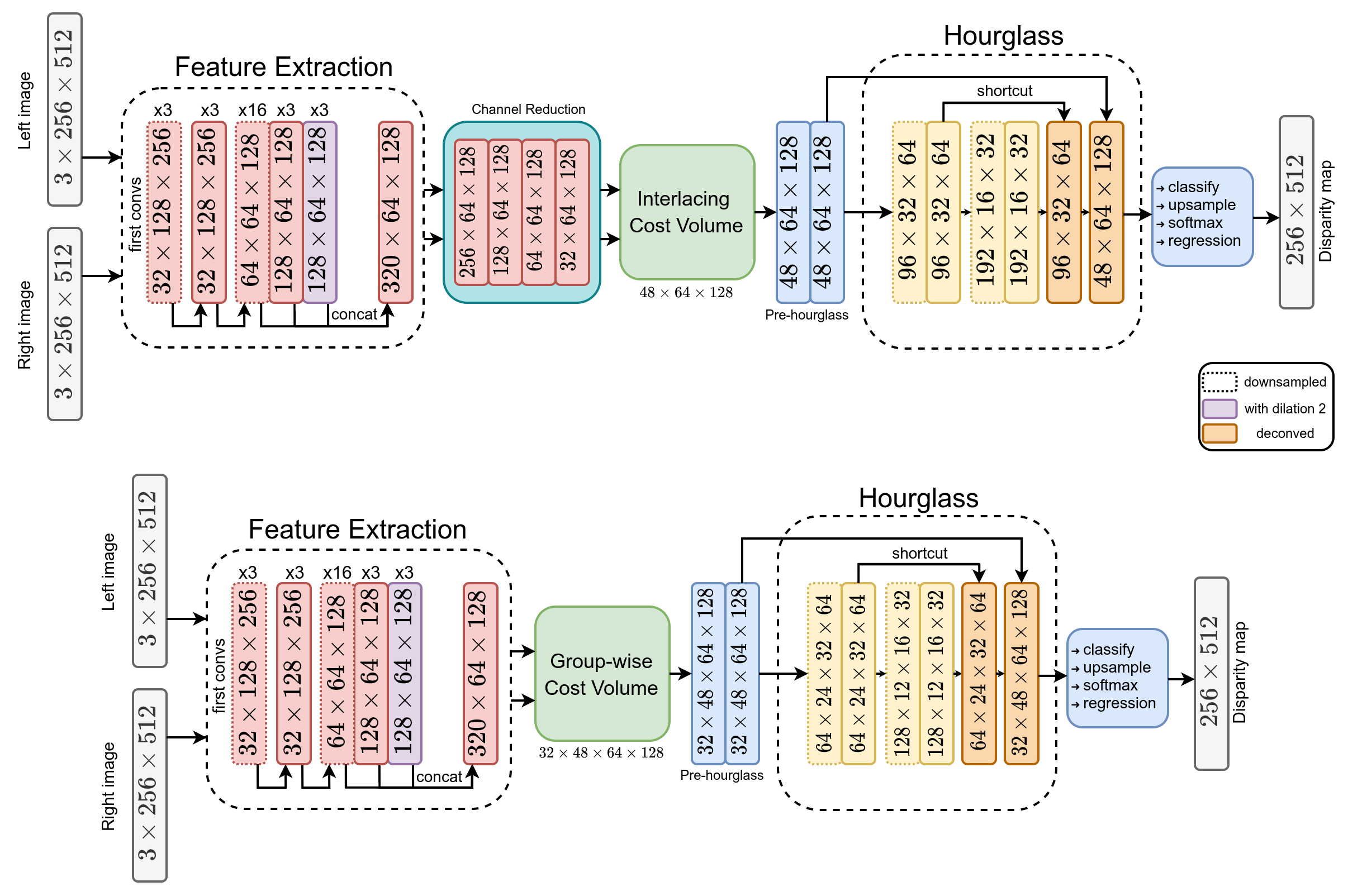}	
	\end{center}
	\caption{\emph{Top}: 2D baseline, \emph{Bottom}: 3D baseline. The numbers in the blocks indicate the output size of each particular layer/module.}
	\label{fig:mobilenets_supp}
\end{figure*}

The detailed architectures of the 2D and 3D baseline models are displayed in Fig. \ref{fig:mobilenets_supp}. The numbers in the blocks indicate the output size of each particular layer/module. The feature extraction step is the same for the two models. The architecture of hourglass and its intraconnections are also similar, except that in the 2D baseline, the convolutions are all in 2D type, while there are 3D convolutions in hourglass of the 3D baseline. These two models differ in the cost volume construction and the channel reduction module as well.
\section{More qualitative results}
\label{Qualitative}
\begin{figure*}[h!]
	\captionsetup[subfigure]{labelformat=empty}
	\centering
	\begin{subfigure}[c]{.33\linewidth}
		\includegraphics[width=1\linewidth]{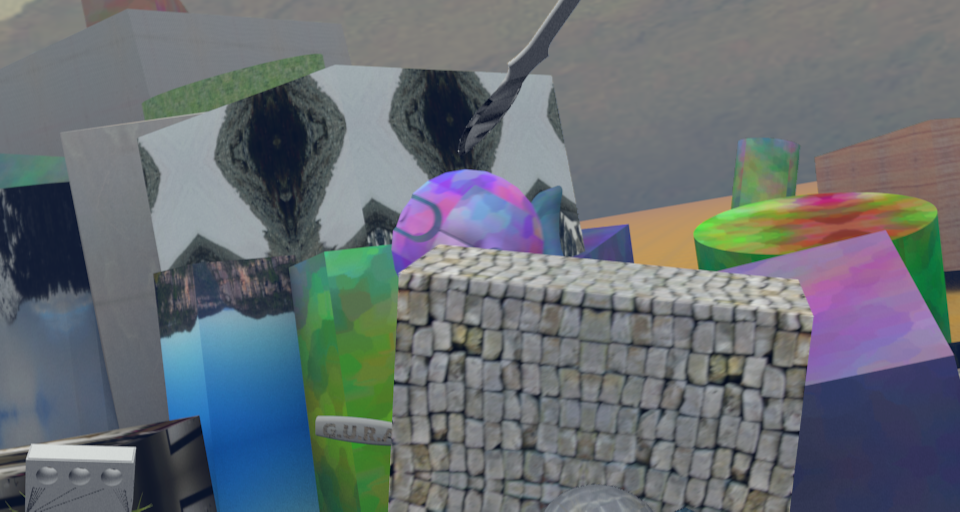}
		\vspace*{-4mm}
	\end{subfigure}
	\begin{subfigure}[c]{.33\linewidth}
		\includegraphics[width=1\linewidth]{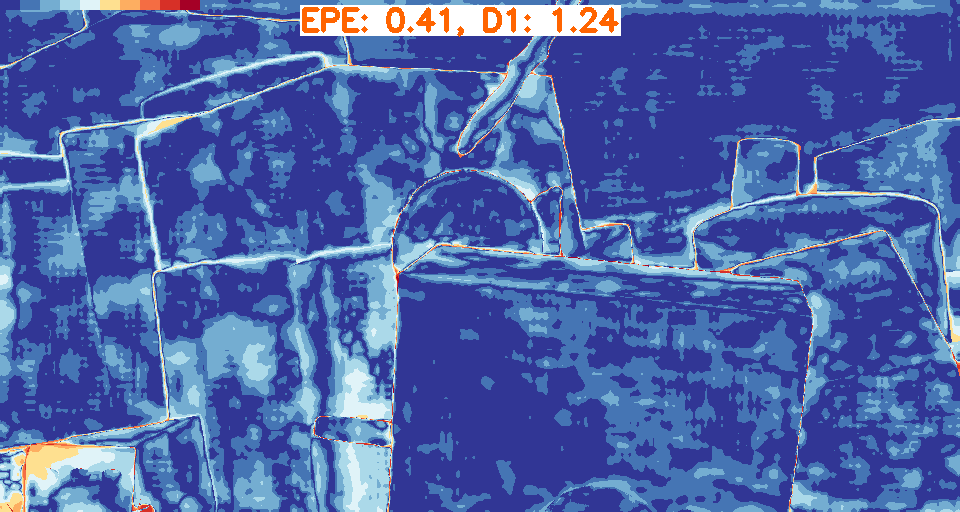}
		\vspace*{-4mm}
	\end{subfigure}
	\begin{subfigure}[c]{.33\linewidth}
		\includegraphics[width=1\linewidth]{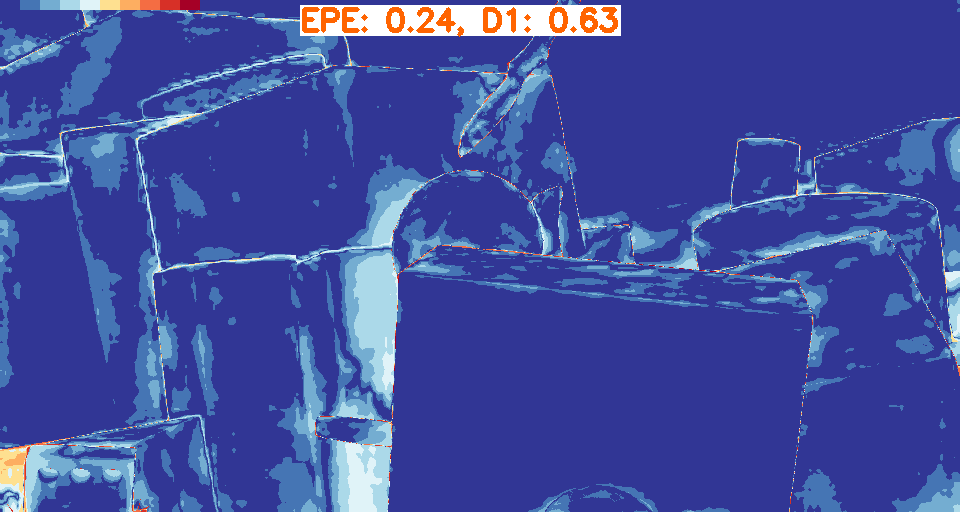}
		\vspace*{-4mm}
	\end{subfigure}
	\begin{subfigure}[c]{.33\linewidth}
		\includegraphics[width=1\linewidth]{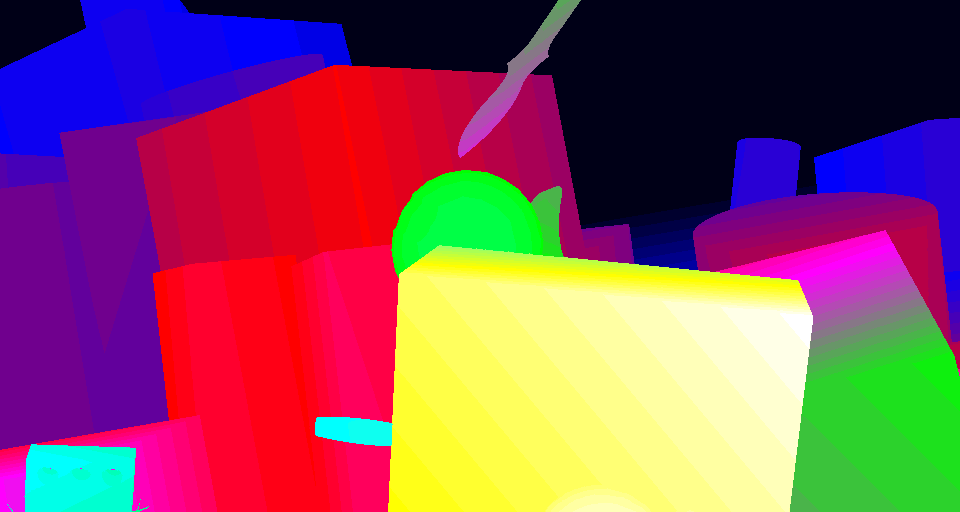}
		\vspace*{-4mm}
	\end{subfigure}
	\begin{subfigure}[c]{.33\linewidth}
		\includegraphics[width=1\linewidth]{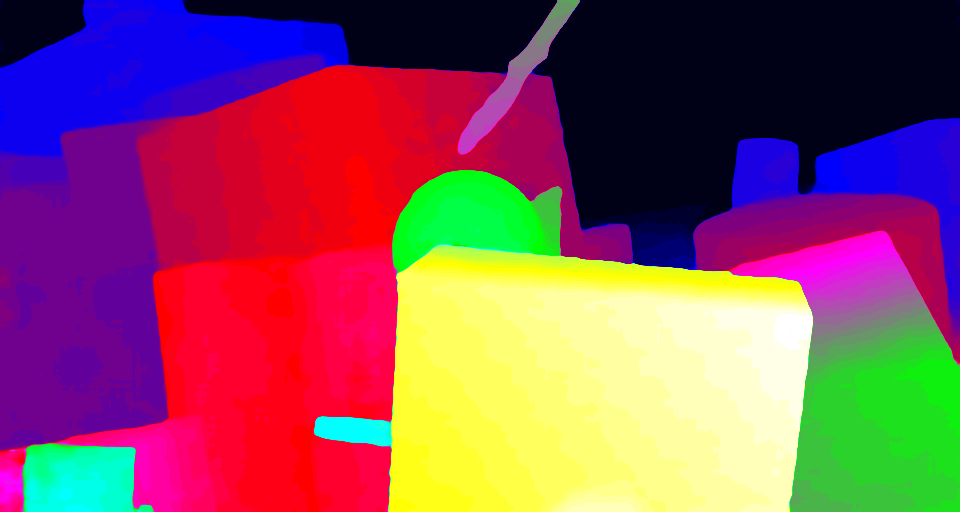}
		\vspace*{-4mm}
	\end{subfigure}
	\begin{subfigure}[c]{.33\linewidth}
		\includegraphics[width=1\linewidth]{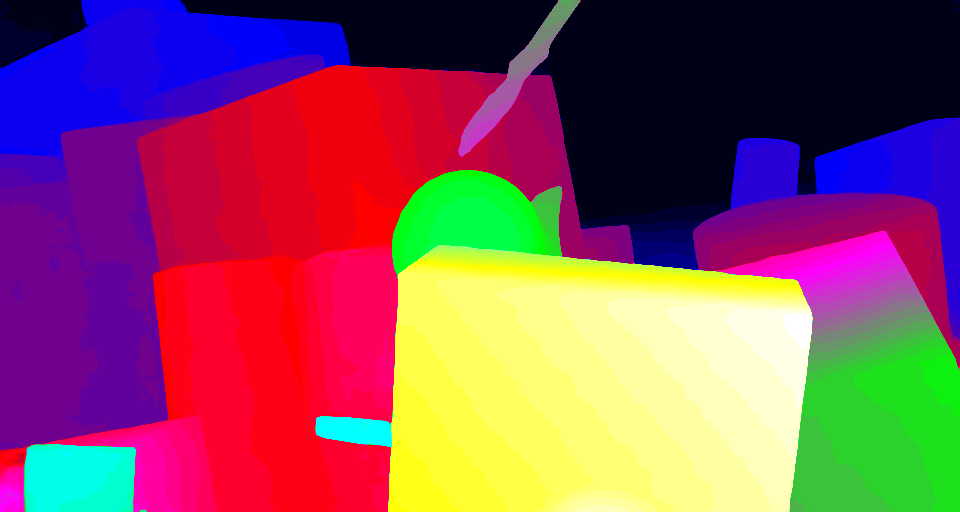}
		\vspace*{-4mm}
	\end{subfigure}
	\begin{subfigure}[c]{.33\linewidth}
		\includegraphics[width=1\linewidth]{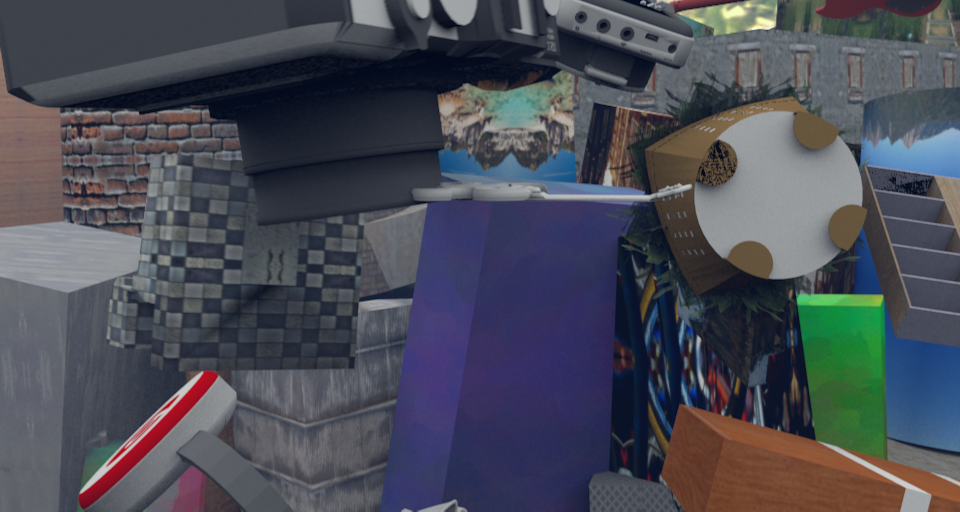}
		\vspace*{-4mm}
	\end{subfigure}
	\begin{subfigure}[c]{.33\linewidth}
		\includegraphics[width=1\linewidth]{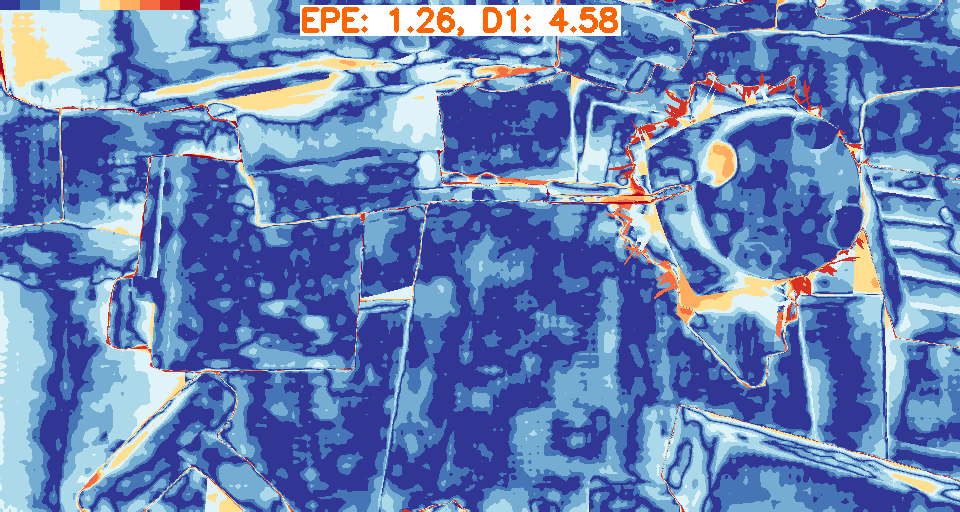}
		\vspace*{-4mm}
	\end{subfigure}
	\begin{subfigure}[c]{.33\linewidth}
		\includegraphics[width=1\linewidth]{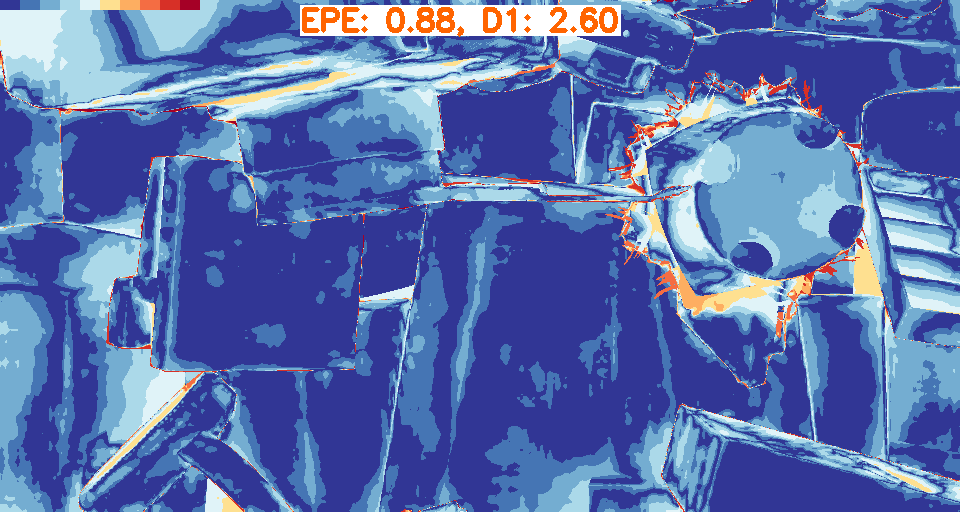}
		\vspace*{-4mm}
	\end{subfigure}
	\begin{subfigure}[c]{.33\linewidth}
		\includegraphics[width=1\linewidth]{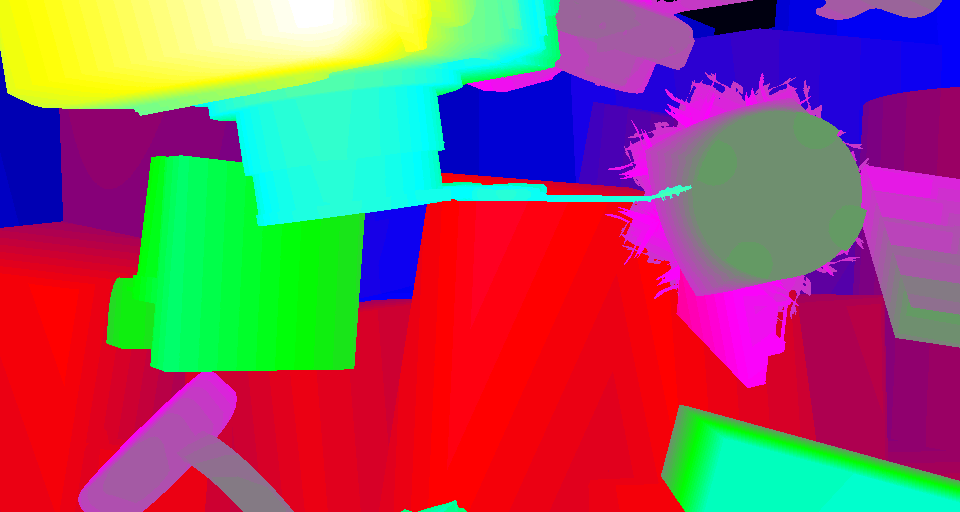}
		\vspace*{-4mm}
	\end{subfigure}
	\begin{subfigure}[c]{.33\linewidth}
		\includegraphics[width=1\linewidth]{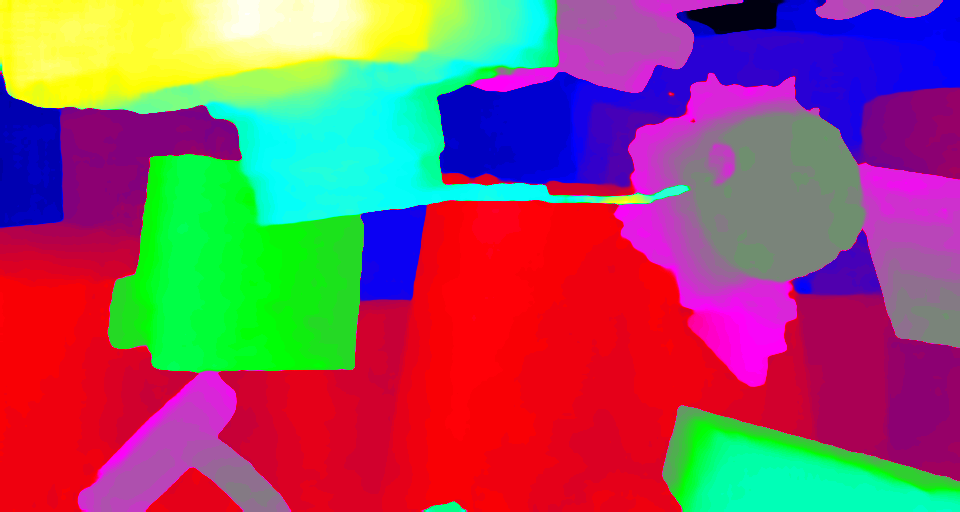}
		\vspace*{-4mm}
	\end{subfigure}
	\begin{subfigure}[c]{.33\linewidth}
		\includegraphics[width=1\linewidth]{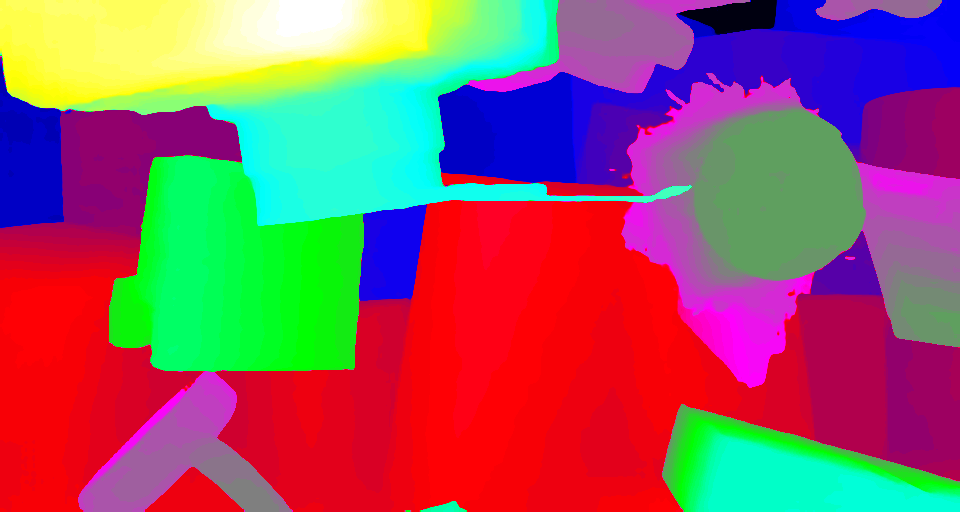}
		\vspace*{-4mm}
	\end{subfigure}
	\begin{subfigure}[c]{.33\linewidth}
		\includegraphics[width=1\linewidth]{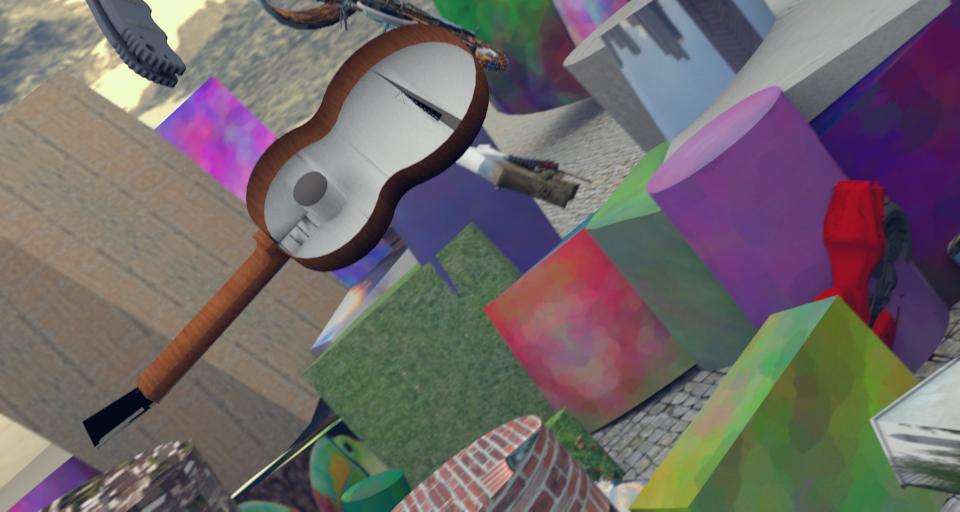}
		\vspace*{-4mm}
	\end{subfigure}
	\begin{subfigure}[c]{.33\linewidth}
		\includegraphics[width=1\linewidth]{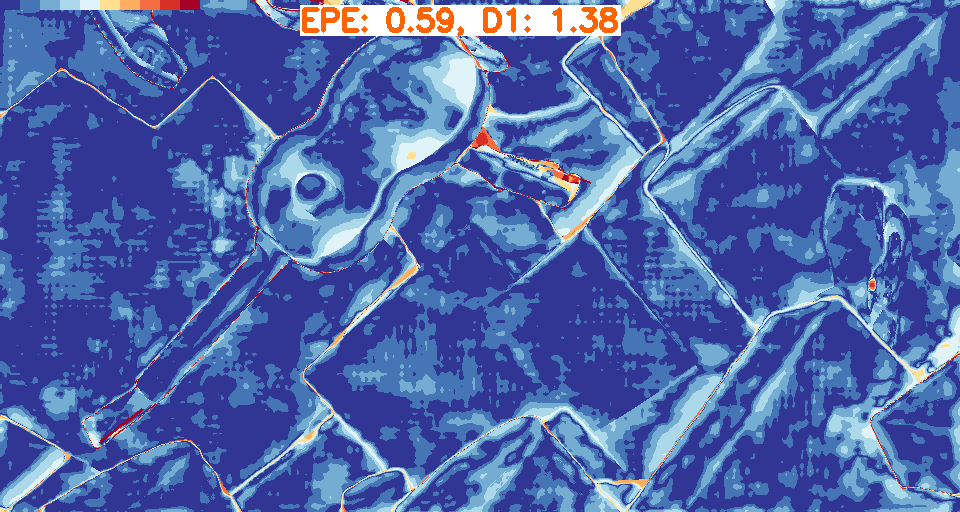}
		\vspace*{-4mm}
	\end{subfigure}
	\begin{subfigure}[c]{.33\linewidth}
		\includegraphics[width=1\linewidth]{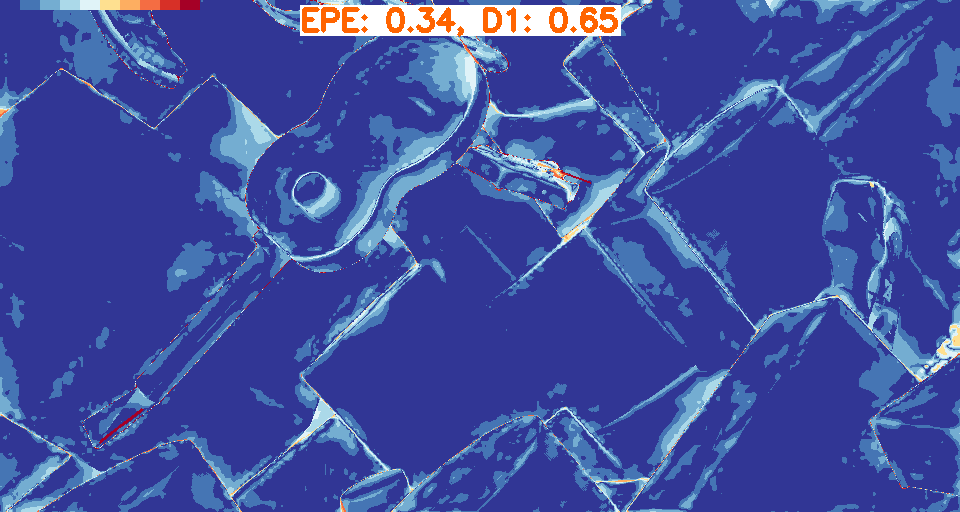}
		\vspace*{-4mm}
	\end{subfigure}
	\begin{subfigure}[c]{.33\linewidth}
		\includegraphics[width=1\linewidth]{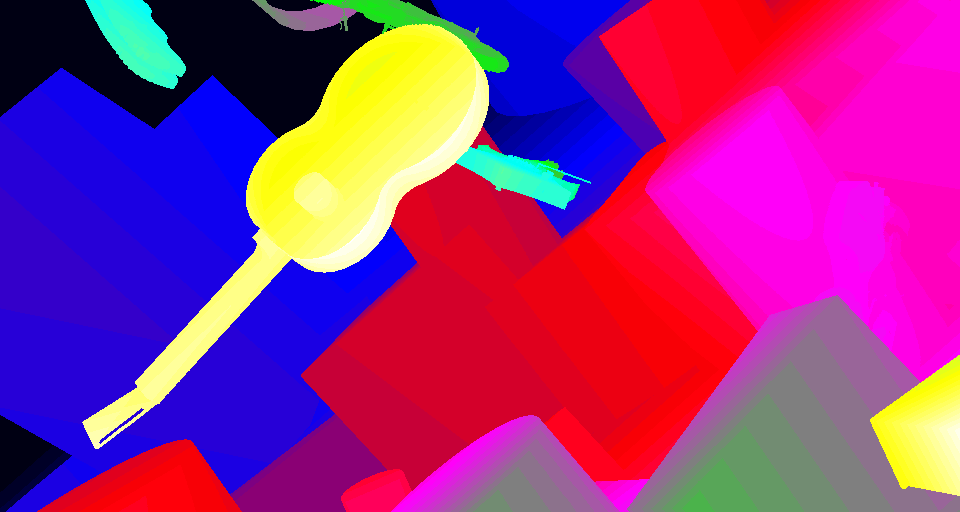}
		\vspace*{-0.6cm}
		\caption{\small{Left image/Disparity}}
	\end{subfigure}
	\begin{subfigure}[c]{.33\linewidth}
		\includegraphics[width=1\linewidth]{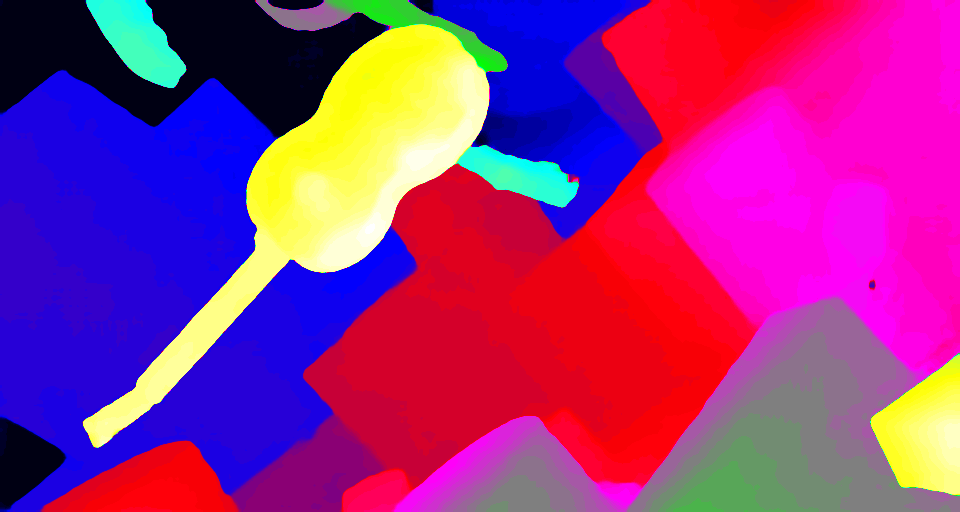}
		\vspace*{-0.6cm}
		\caption{\small{2D-MobileStereoNet}}
	\end{subfigure}
	\begin{subfigure}[c]{.33\linewidth}
		\includegraphics[width=1\linewidth]{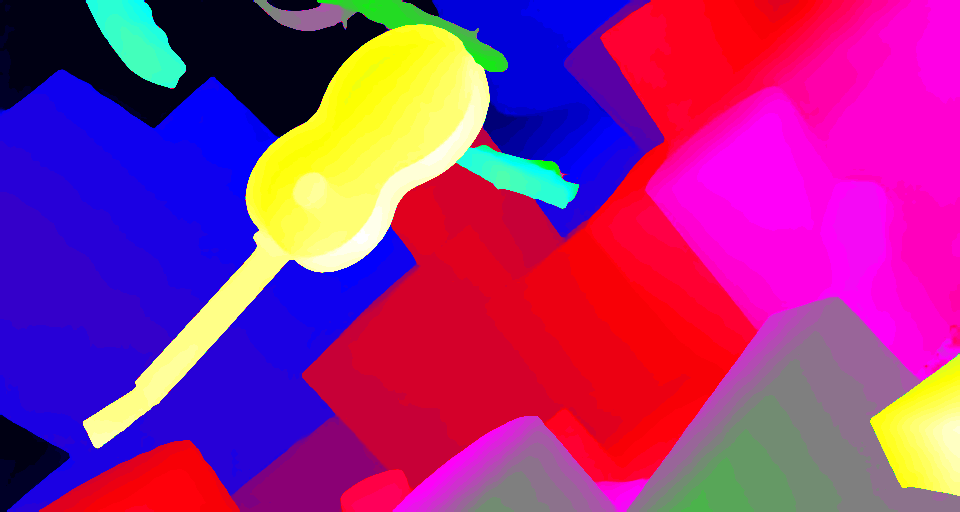}
		\vspace*{-0.6cm}
		\caption{\small{3D-MobileStereoNet}}
	\end{subfigure}
	\\
	\vspace{-0.3cm}
	\caption{Qualitative performance on SceneFlow: Every two rows correspond to a test sample. In the left-most column, the samples and the ground-truth disparity maps are illustrated. The following two columns show the disparity and error maps (embedded with error values) estimated by 2D-MobileStereoNet and 3D-MobileStereoNet. Warmer colors in error maps denote higher errors.}
	\label{fig:SceneFlow_supp}
	\vspace*{-0.3cm}
\end{figure*}

\begin{figure*}[h!]
	\captionsetup[subfigure]{labelformat=empty}
	\centering
	\small{Left image}\\
	\begin{subfigure}[c]{0.33\linewidth}
		\includegraphics[width=1\linewidth]{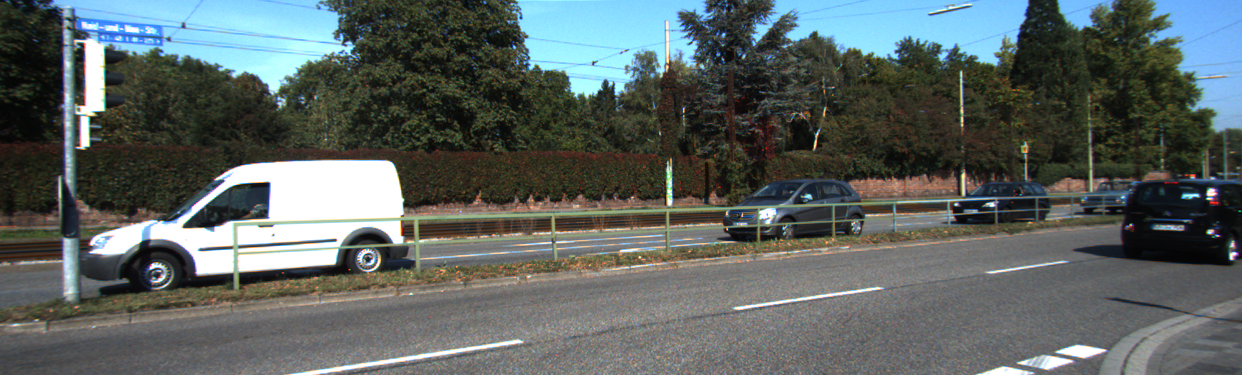}
	\end{subfigure}	
	\begin{subfigure}[c]{0.33\linewidth}
		\includegraphics[width=1\linewidth]{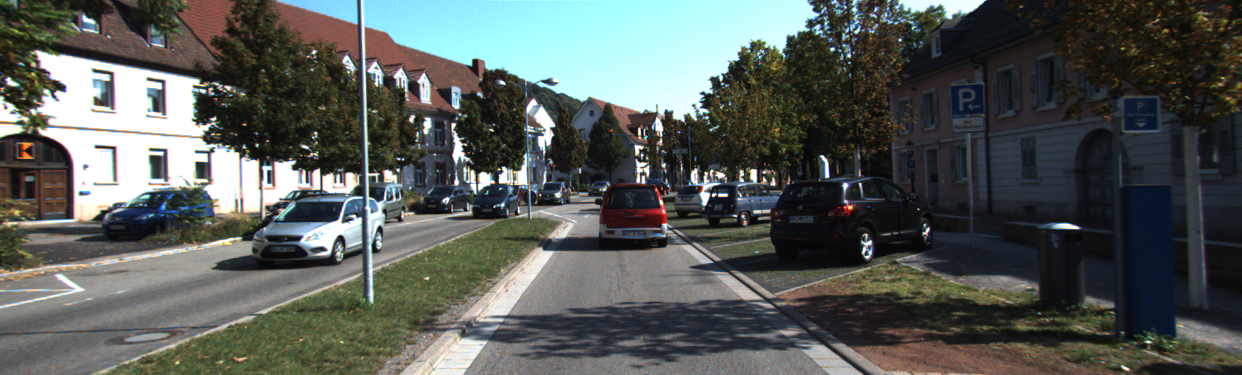}
	\end{subfigure}	
	\begin{subfigure}[c]{0.33\linewidth}
		\includegraphics[width=1\linewidth]{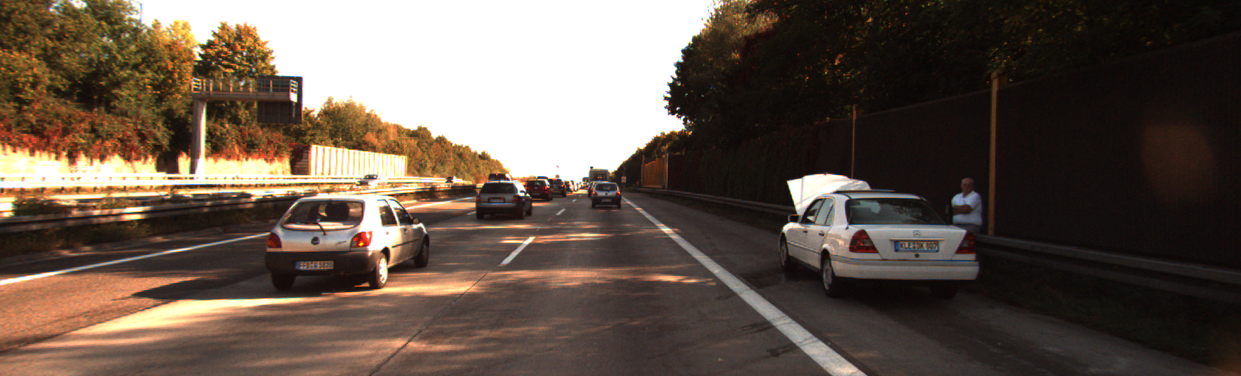}
	\end{subfigure}\\
	\small{Ground-truth}\\
	\begin{subfigure}[c]{0.33\linewidth}
		\includegraphics[width=1\linewidth]{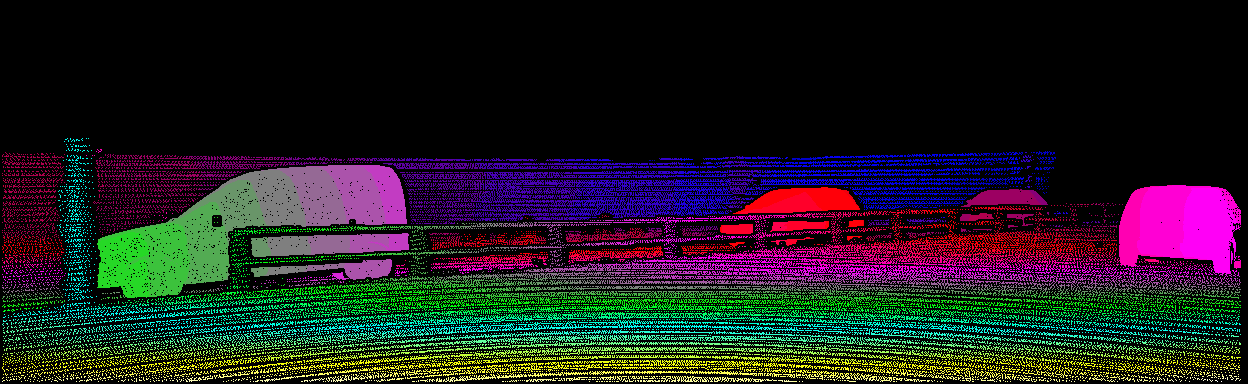}
	\end{subfigure}	
	\begin{subfigure}[c]{0.33\linewidth}
		\includegraphics[width=1\linewidth]{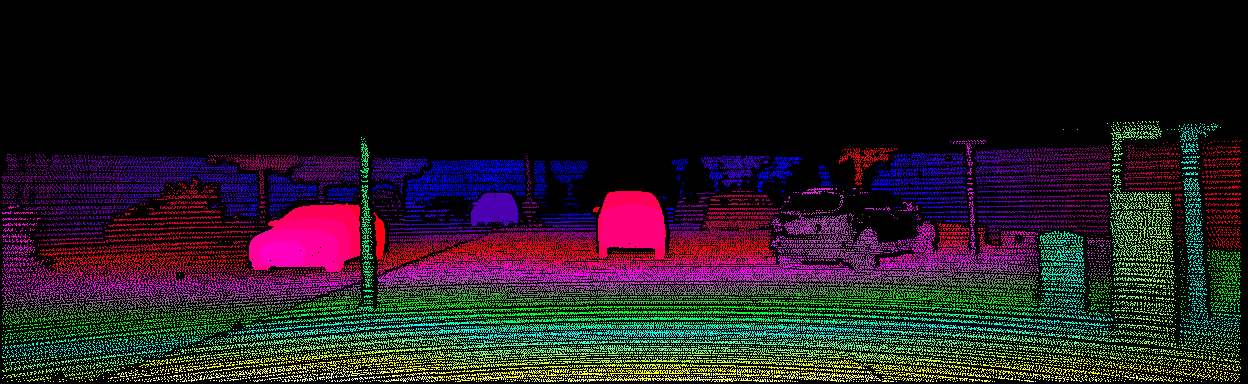}
	\end{subfigure}	
	\begin{subfigure}[c]{0.33\linewidth}
		\includegraphics[width=1\linewidth]{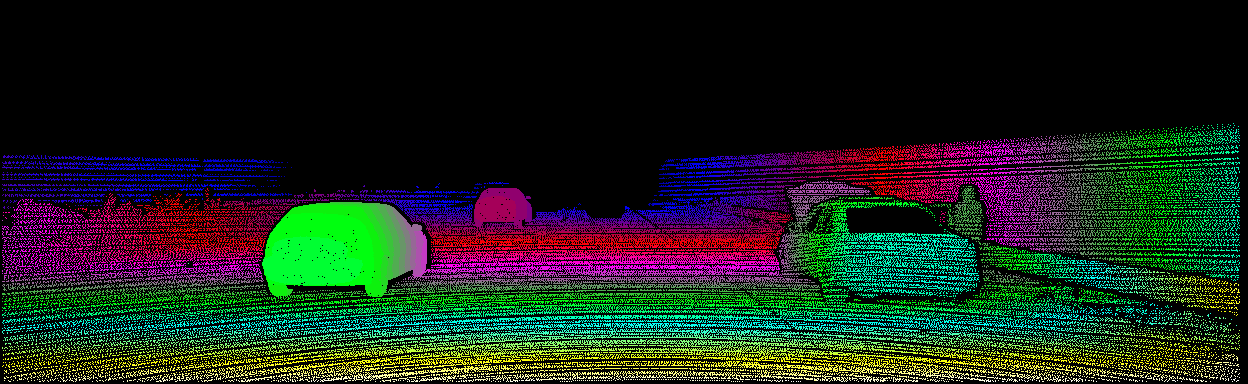}
	\end{subfigure}\\
	\small{PSMNet \cite{chang2018pyramid}}\\
	\begin{subfigure}[c]{0.33\linewidth}
		\includegraphics[width=1\linewidth]{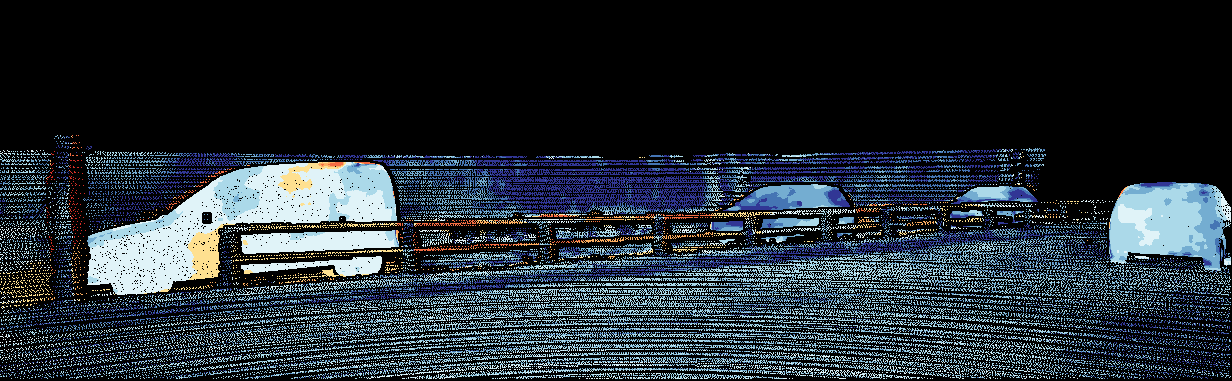}
	\end{subfigure}	
	\begin{subfigure}[c]{0.33\linewidth}
		\includegraphics[width=1\linewidth]{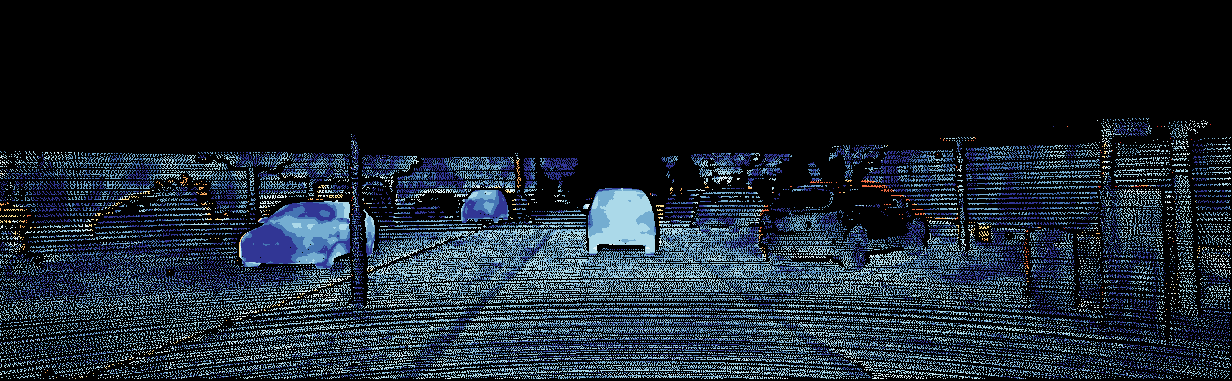}
	\end{subfigure}	
	\begin{subfigure}[c]{0.33\linewidth}
		\includegraphics[width=1\linewidth]{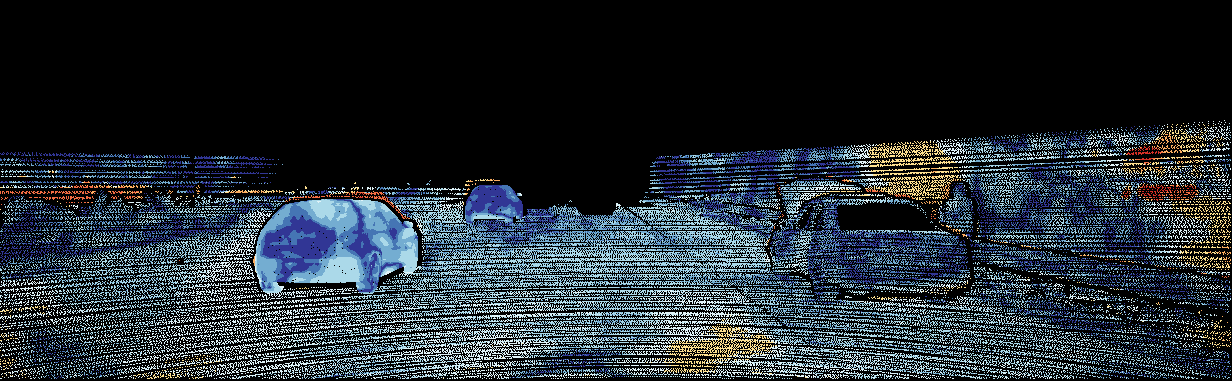}
	\end{subfigure}\\
	\small{GA-Net-11 \cite{zhang2019ga}}\\
	\begin{subfigure}[c]{0.33\linewidth}
		\includegraphics[width=1\linewidth]{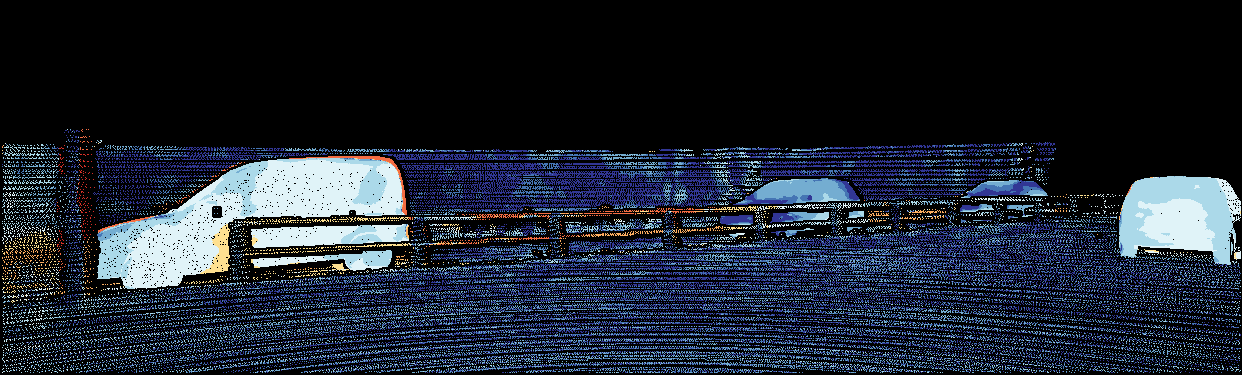}
	\end{subfigure}	
	\begin{subfigure}[c]{0.33\linewidth}
		\includegraphics[width=1\linewidth]{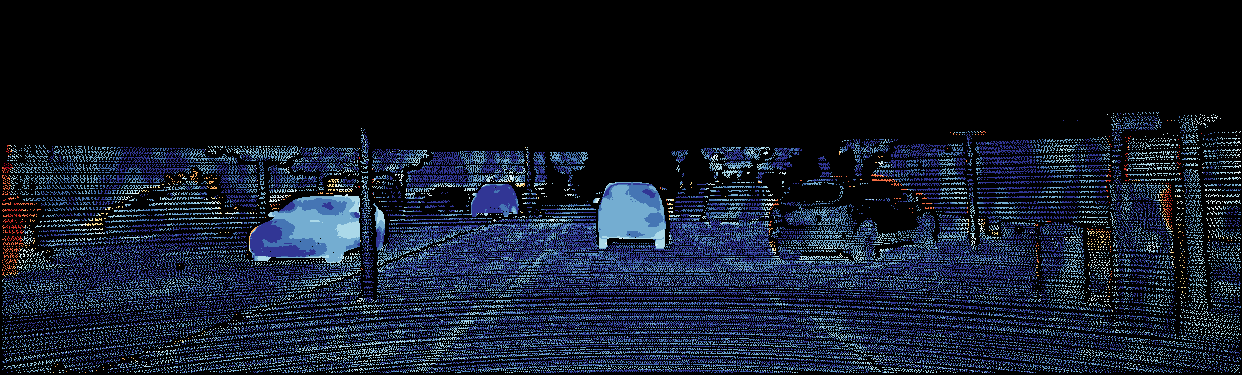}
	\end{subfigure}	
	\begin{subfigure}[c]{0.33\linewidth}
		\includegraphics[width=1\linewidth]{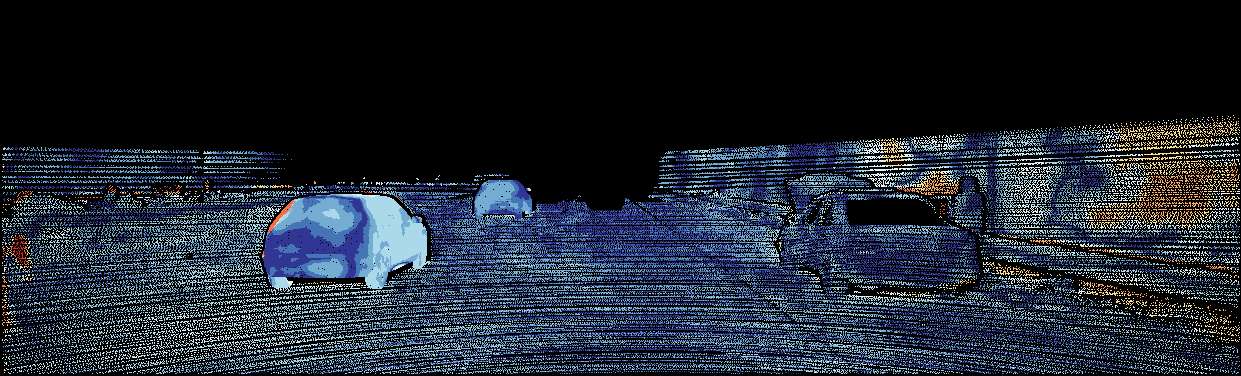}
	\end{subfigure}\\
	\small{GA-Net-deep \cite{zhang2019ga}}\\
	\begin{subfigure}[c]{0.33\linewidth}
		\includegraphics[width=1\linewidth]{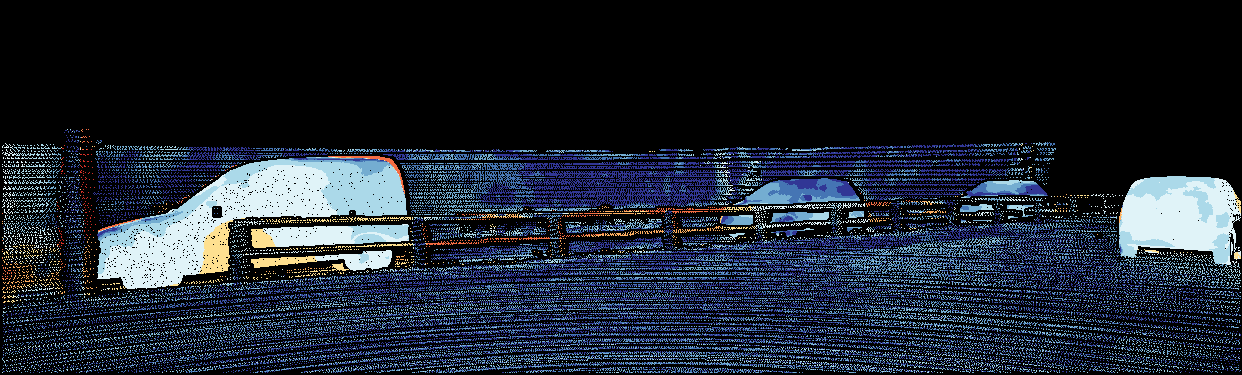}
	\end{subfigure}	
	\begin{subfigure}[c]{0.33\linewidth}
		\includegraphics[width=1\linewidth]{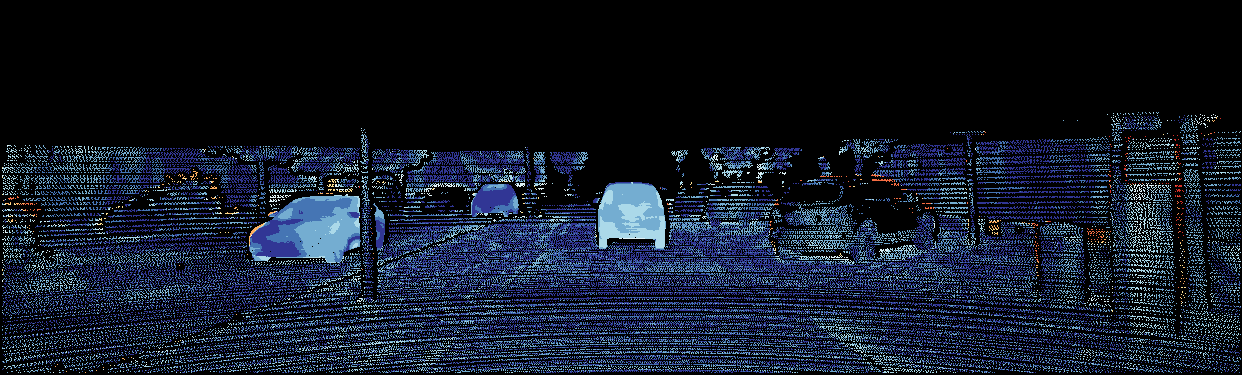}
	\end{subfigure}	
	\begin{subfigure}[c]{0.33\linewidth}
		\includegraphics[width=1\linewidth]{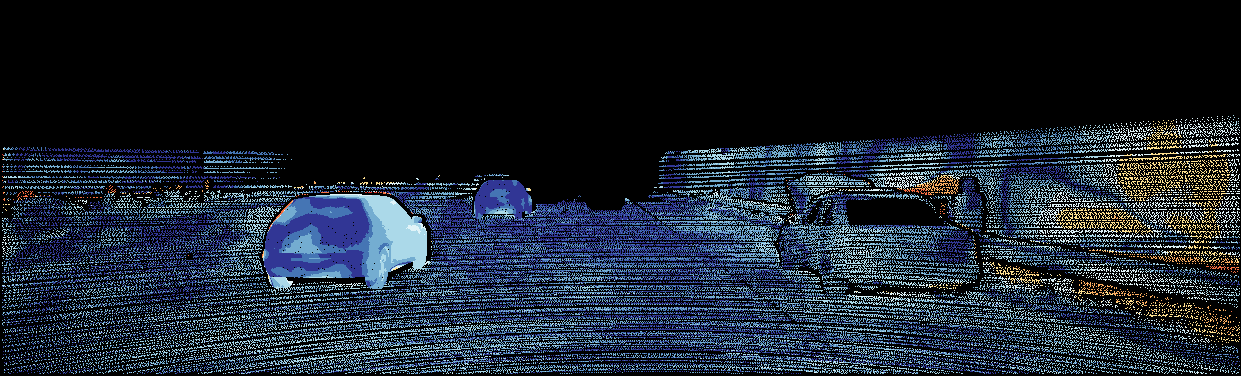}
	\end{subfigure}\\
	\small{GwcNet-g \cite{guo2019group}}\\
	\begin{subfigure}[c]{0.33\linewidth}
		\includegraphics[width=1\linewidth]{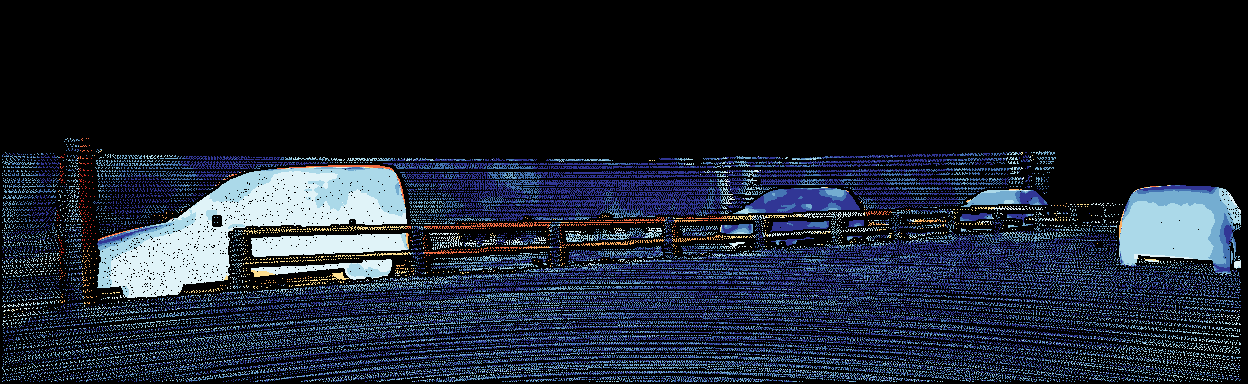}
	\end{subfigure}	
	\begin{subfigure}[c]{0.33\linewidth}
		\includegraphics[width=1\linewidth]{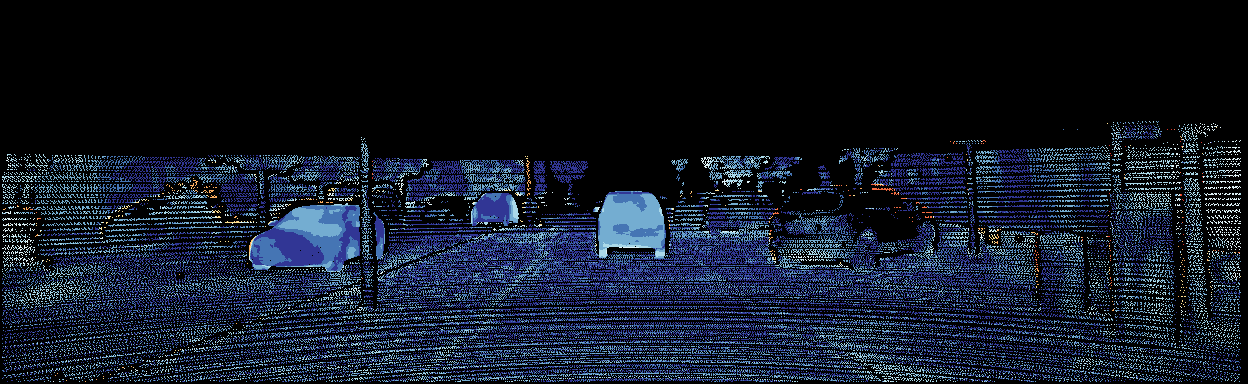}
	\end{subfigure}	
	\begin{subfigure}[c]{0.33\linewidth}
		\includegraphics[width=1\linewidth]{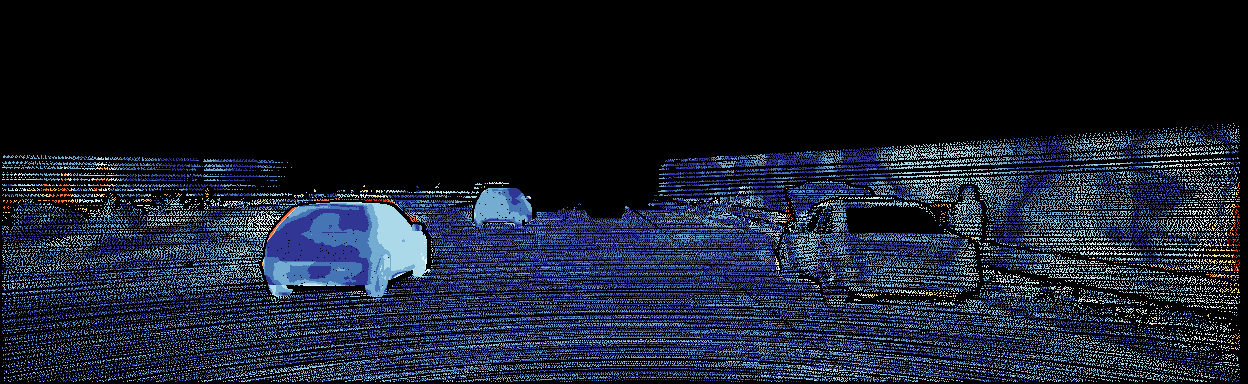}
	\end{subfigure}\\
	\small{2D-MobileStereoNet}\\
	\begin{subfigure}[c]{0.33\linewidth}
		\includegraphics[width=1\linewidth]{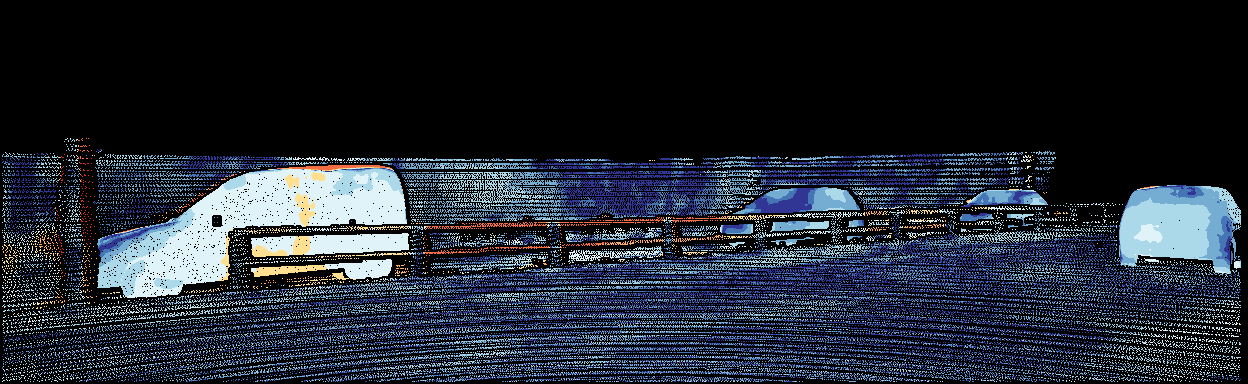}
	\end{subfigure}	
	\begin{subfigure}[c]{0.33\linewidth}
		\includegraphics[width=1\linewidth]{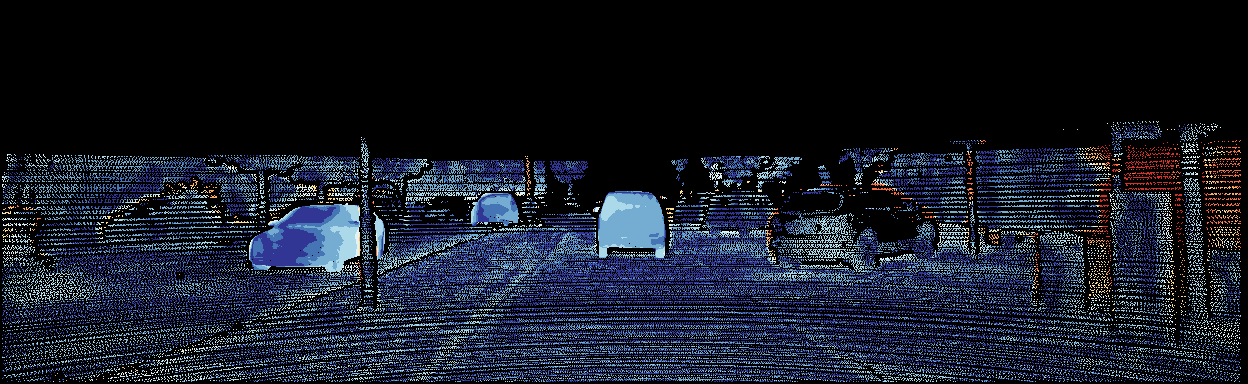}
	\end{subfigure}	
	\begin{subfigure}[c]{0.33\linewidth}
		\includegraphics[width=1\linewidth]{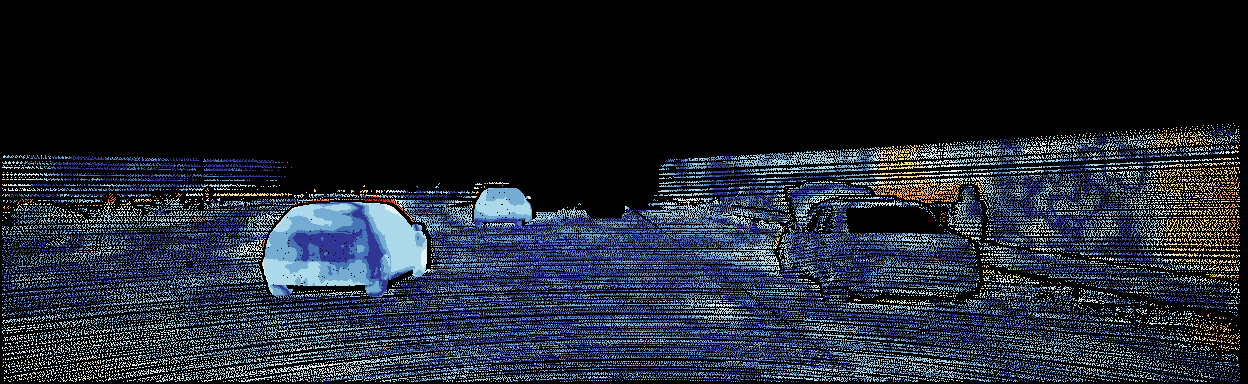}
	\end{subfigure}\\
	\small{3D-MobileStereoNet}\\
	\begin{subfigure}[c]{0.33\linewidth}
		\includegraphics[width=1\linewidth]{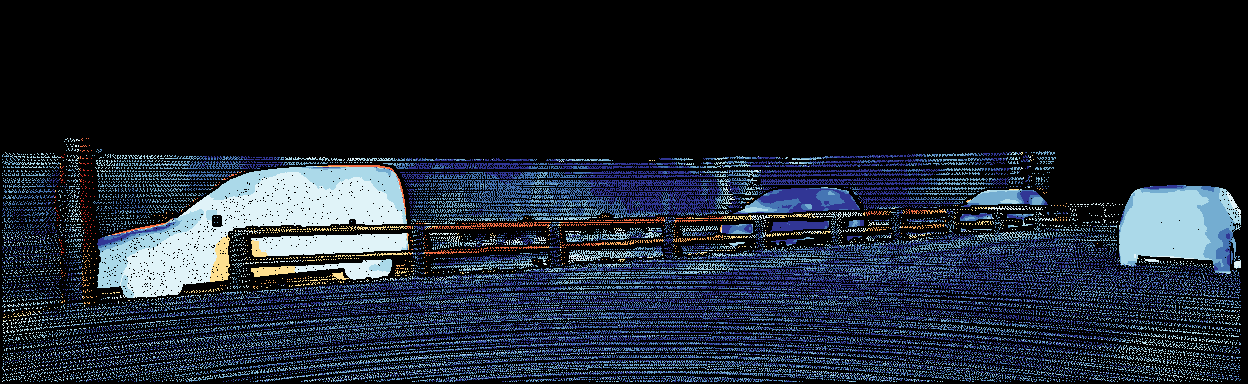}
	\end{subfigure}	
	\begin{subfigure}[c]{0.33\linewidth}
		\includegraphics[width=1\linewidth]{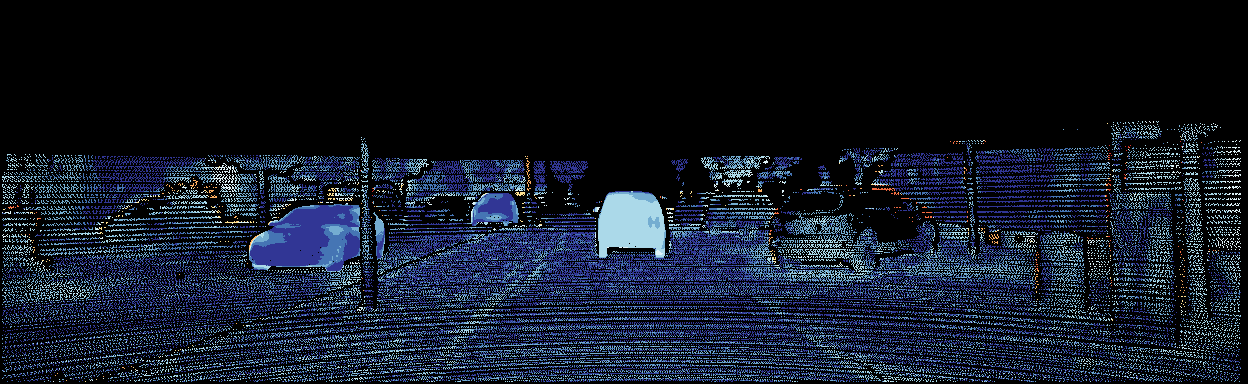}
	\end{subfigure}	
	\begin{subfigure}[c]{0.33\linewidth}
		\includegraphics[width=1\linewidth]{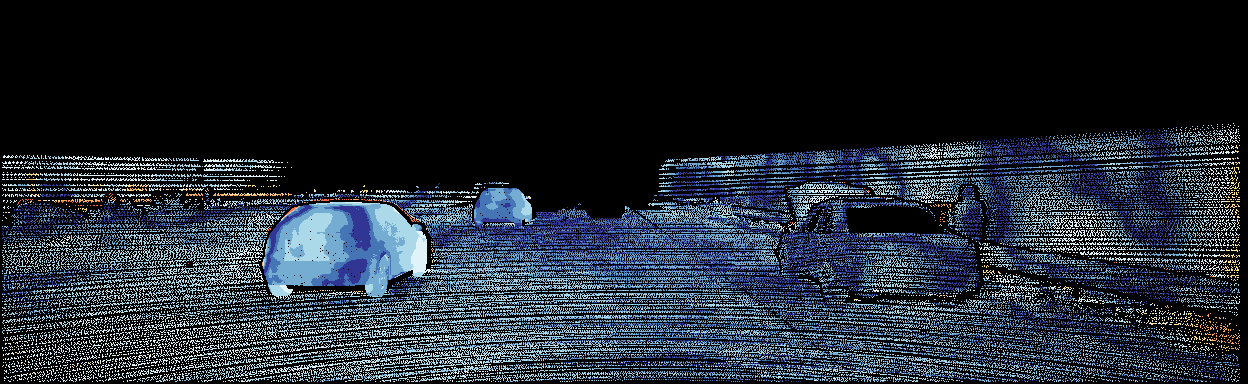}
	\end{subfigure}
	\caption{Qualitative performance on KITTI 2015 validation set: From top to bottom, the left image, the ground-truth disparity map and the estimated disparity maps by PSMNet \cite{chang2018pyramid}, GA-Net-11 \cite{zhang2019ga}, GA-Net-deep \cite{zhang2019ga}, GwcNet-g \cite{guo2019group}, 2D-MobileStereoNet and 3D-MobileStereoNet are illustrated. For a fair comparison, we trained all the models with a 160/40 split of KITTI 2015 training test. Warmer colors in error maps denote higher errors.}
	\label{fig:KittiVal}
\end{figure*}

Figure \ref{fig:SceneFlow_supp} depicts more qualitative results on SceneFlow dataset. We have also shown qualitative comparison on KITTI 2015 validation set in Fig. \ref{fig:KittiVal}.  

From Fig. \ref{fig:KittiVal}, once again, we can verify that 2D-MobileStereoNet shows close performance to 3D models with the least number of operations. Also, 3D-MobileStereoNet obtains competitive or better accuracy with the least number of parameters among other methods.
\section{Incorporating light blocks in other modules}
\label{Incorporating}
As mentioned in the paper, in order to further reduce the complexity, the first convolutions in the feature extraction and the pre-hourglass convolutions (\emph{cf.} Fig. \ref{fig:mobilenets_supp}) are replaced with MobileNet-V2 ($v_2$). The experimental results are reported in Tables \ref{tab:2DNet_comb_further} and \ref {tab:3DNet_comb_further}. Note that the first convolutions are of the 2D type for both 2D and 3D baselines; however, the pre-hourglass comes in 2D or 3D convolutions depending on the baseline. We can observe that in 2D-MobileStereoNet, when the two modules are replaced with MobileNet-V2 ($v_2$), the network obtains the least EPE. In 3D-MobileStereoNet, this combination yields slightly higher EPE. However, due to the nice reduction in the computation cost, we consider the same design choice for the 3D network. It is noteworthy that we have examined MobileNet-V1 ($v_1$) for these modules as well. However, as it deteriorates the performance, we ignore $v_1$ for these modules, albeit it shows much decrease in the cost.
\begin{table}[h]
	\begin{center}
		\vspace{-0.2cm}
		\footnotesize	
		\begin{tabular}{c;{1pt/3pt}c|S[table-format=1.2]S[table-format=2.2]S[table-format=1.2]}
			\hline
			{first-conv$_{2D}$}  & {pre-HG$_{2D}$}  & {EPE($px$) $\downarrow$} & {MACs($G$) $\downarrow$} & {Params($M$) $\downarrow$} \\ \hline
			conv.	      & conv.   & 1.50 & 30.33 & 1.21 \\\hdashline
			conv.	     & $v_2$   & 1.41 & 30.0 & 1.16 \\		
			$v_2$	  	 & conv.   & 1.54 & 29.75 & 1.20\\
			$v_2$	  & $v_2$   & 1.40 & 29.42 & 1.15 \\ \hline
		\end{tabular}
	\end{center}
	\vspace*{-0.5cm}
	\caption{Performance evaluation for the selected variant of 2D baseline (FE$_{2D}$:$v_1$, HG$_{2D}$:$v_2$) from Tab. 3a of the paper, when replacing other components with $v_2$ block ($t=2$).}
	\label{tab:2DNet_comb_further}
\end{table}
\vspace{-0.5cm}
\begin{table}[h]
	\begin{center}
		\footnotesize	
		\begin{tabular}{c;{1pt/3pt}c|S[table-format=1.2]S[table-format=2.2]S[table-format=1.2]}
			\hline
			{first-conv$_{2D}$}  & {pre-HG$_{3D}$}  & {EPE($px$) $\downarrow$} & {MACs($G$) $\downarrow$} & {Params($M$) $\downarrow$} \\ \hline
			conv.  	 & conv. & 0.99 & 105.01 & 0.98 \\ \hdashline
			conv.    & $v_2$ & 1.01 & 69.44  & 0.89 \\		
			$v_2$	 & conv. & 0.99 & 104.44 & 0.97 \\
			$v_2$	 & $v_2$ & 1.01 & 68.86  & 0.88 \\ \hline
		\end{tabular}
	\end{center}
	\vspace*{-0.5cm}
	\caption{Performance evaluation for the selected variant of 3D baseline (FE$_{2D}$:$v_1$, HG$_{3D}$:$v_2$) from Tab. 3b of the paper, when replacing other components with $v_2$ block ($t=2$).}
	\label{tab:3DNet_comb_further}
\end{table}
\section{Implementation details}
\label{Implementation}
We used PyTorch for implementation and conducting experiments. All the trainings are executed on 4 $\times$ NVIDIA GeForce GTX 1080 Ti. We adapt the Adam optimizer with $\beta_1=0.9$ and $\beta_2=0.999$. On the SceneFlow dataset, the networks are trained for 20 epochs, starting with a learning rate of 0.001. The learning rate is halved after epoch 10, 12, 14, and 16. The best model is selected based on the least EPE value. In the experiments on the KITTI 2015 validation set, we finetune the best SceneFlow model for 400 epochs, reducing the initial learning rate 0.001 by a factor of 10 after 200 epochs. To submit the results to the KITTI 2015 benchmark, we finetune starting from a SceneFlow checkpoint showing the best generalization performance from the SceneFlow to the KITTI 2015 images. For the 3D-MobileStereoNet, we used a batch size of 4, and for 2D-MobileStereoNet, the batch size is 8.
\section{Analyzing the complexity}
\label{Analyzing}
Table \ref{tab:2d/3d} shows the computation cost of the main modules, \ie feature extraction and encoder-decoder, in baselines (with standard convolutions) and in MobileStereoNets. Note that feature extraction is the same in 2D and 3D models. We see our design choice for feature extraction is significantly reducing the complexity both in operation (from 52.07 to 7.84 GigaMACs) and in parameters (from 7.84 to only 0.39 million). We also observe that the cost of the encoder-decoder modules, either in 2D or 3D, is reduced in lighter networks in both number of operations and parameters. Evidently, the major bottleneck for the 3D models is the encoder-decoder with 3D convolutions.
\begin{table*}[h]
	\begin{center}
		\vspace{-0.2cm}
		\footnotesize
		\begin{tabular}{lS[table-format=2.2];{1pt/3pt}S[table-format=2.2]|S[table-format=2.2];{1pt/3pt}S[table-format=2.2]}
			\cline{2-5}
			& \multicolumn{2}{c}{Baselines}       & \multicolumn{2}{c}{MobileStereoNets} \\ \cline{2-5} 
			\multicolumn{1}{l}{}                          & {MACs($G$)}   & {Params($M$)} 		& {MACs($G$)}     & {Params($M$)} \\ \hline
			\multicolumn{1}{l|}{Feature Extraction}       & 52.07         & 2.95                & 7.84       & 0.39    \\ 
			\multicolumn{1}{l|}{Encoder-decoder$_{2D}$ in 2D-MobileStereoNet}    & 4.38          & 2.61                & 3.92       & 1.64    \\ 
			\multicolumn{1}{l|}{Encoder-decoder$_{3D}$ in 3D-MobileStereoNet}    & 167.51        & 3.45                & 128.73     & 1.34    \\ \hline			
		\end{tabular}
	\end{center}
	\vspace*{-0.5cm}
	\caption{Analyzing the computation cost in terms of MACs and number of parameters for the main modules.}
	\label{tab:2d/3d}
\end{table*}

\end{document}